\pdfoutput=1
\documentclass{applemlr}
\usepackage{amsmath}
\usepackage{enumerate}
\usepackage{algorithm}
\usepackage{algpseudocode}
\usepackage{amsfonts}
\usepackage{amsthm}
\usepackage{cleveref}
\usepackage{diagbox}
\usepackage{colortbl}
\usepackage{amssymb}
\usepackage{xspace}
\usepackage{wrapfig}
\usepackage{adjustbox}
\usepackage{tabularx}
\usepackage{booktabs}
\usepackage{mathtools}
\usepackage{tikz}
\usepackage{enumitem}
\usepackage{silence}
\usepackage{dsfont}
\usepackage[table]{xcolor}
\usepackage[dvipsnames]{xcolor}
\usepackage{multirow}
\usepackage{makecell}
\usepackage{xfakebold}

\usepackage{amsmath,amsfonts,bm}

\def\ra{{\textnormal{a}}}

\DeclareMathAlphabet{\mathsfit}{\encodingdefault}{\sfdefault}{m}{sl}
\SetMathAlphabet{\mathsfit}{bold}{\encodingdefault}{\sfdefault}{bx}{n}

\usetikzlibrary{positioning, calc}
\usetikzlibrary{decorations.pathmorphing}

\makeatletter
\patchcmd{\wrong@fontshape}{\@gobbletwo}{}{}{}
\makeatother
\numberwithin{equation}{section}
\makeatletter
\AtBeginDocument{
  \urlstyle{sf}
}
\makeatother

\usepackage{graphicx}
\usepackage{tikz}
\usepackage{float}
\usepackage{wrapfig}
\usepackage{subcaption}
\usepackage{amsmath,amssymb,amsthm}
\usepackage{bm,xspace}
\usepackage{verbatim}
\usepackage{url}
\usepackage{booktabs}
\usepackage{etoolbox,siunitx}
\usepackage{calc}
\usepackage{pifont,hologo}
\usepackage{nicefrac}
\usepackage{enumitem}

\usepackage[dvipsnames,table]{xcolor}

\providecommand*{\Tab}{Tab\onedot}
\providecommand*{\Fig}{Fig\onedot}
\providecommand*{\Sec}{Sec\onedot}

\DeclareRobustCommand*\circled[1]{\tikz[baseline=(char.base)]{
    \node[shape=circle,draw,inner sep=0.8pt] (char) {#1};}}

\definecolor{tab_first}{RGB}{119, 221, 119}
\definecolor{tab_second}{RGB}{223, 255, 0}
\definecolor{tab_third}{RGB}{255, 253, 208}

\newcommand\cellfirst[1]{\cellcolor{tab_first}\textbf{#1}}
\newcommand\cellsecond[1]{\cellcolor{tab_second}#1}
\newcommand\cellthird[1]{\cellcolor{tab_third}#1}

\newcommand\grayed[1]{\textcolor{Gray}{#1}}
\newcommand\podiumbox[2]{{\setlength{\fboxsep}{1pt}\colorbox{#1}{\strut #2}}}
\DeclareRobustCommand\textfirst[1]{\podiumbox{tab_first}{\textbf{#1}}}
\DeclareRobustCommand\textsecond[1]{\podiumbox{tab_second}{#1}}
\DeclareRobustCommand\textthird[1]{\podiumbox{tab_third}{#1}}

\definecolor{abl_up_3}{RGB}{119, 221, 119}
\definecolor{abl_dn_3}{RGB}{221, 119, 119}

\newcommand\betterA[1]{\cellcolor{abl_up_3!12}#1}
\newcommand\betterB[1]{\cellcolor{abl_up_3!23}#1}
\newcommand\betterC[1]{\cellcolor{abl_up_3!35}#1}
\newcommand\worseA[1]{\cellcolor{abl_dn_3!12}#1}
\newcommand\worseB[1]{\cellcolor{abl_dn_3!23}#1}
\newcommand\worseC[1]{\cellcolor{abl_dn_3!35}#1}

\newcommand\fitwidth[1]{%
  \resizebox{\ifdim\width>\linewidth \linewidth\else \width\fi}{!}{#1}}

\newcommand\oursname{Depth Pro 2}
\newcommand\ours{\oursname\xspace}
\newcommand\ourss{\oursname-S\xspace}
\newcommand\oursb{\oursname-B\xspace}
\newcommand\oursl{\oursname-L\xspace}
\newcommand\oursh{\oursname-H\textsuperscript{+}\xspace}

\definecolor{grp_single}{HTML}{2E7FBF}
\definecolor{grp_multi}{HTML}{2E9E5B}
\definecolor{grp_ours}{HTML}{F08A3C}

\definecolor{qualbox}{HTML}{00FF66}
\newcommand\groupsingle[1]{\textcolor{grp_single}{#1}}
\newcommand\groupmulti[1]{\textcolor{grp_multi}{#1}}
\newcommand\groupours[1]{\textcolor{grp_ours}{#1}}

\newcommand\tDepthPro{Depth Pro}
\newcommand\tOursFoV{\oursname\xspace}
\newcommand\tAnyCalib{AnyCalib}
\newcommand\tGeoCalib{GeoCalib}
\newcommand\tDepthAnythingThree{DA3-NGL}
\newcommand\tDepthCrafter{DepthCrafter}
\newcommand\tChronoDepth{ChronoDepth}
\newcommand\tDAGE{DAGE}
\newcommand\tDVD{DVD}
\newcommand\tFlashDepth{FlashDepth-L}
\newcommand\tLotus{Lotus-2}
\newcommand\tMapAnything{MapAnything}
\newcommand\tMoGe{MoGe-2-L}
\newcommand\tMoGeThree{MoGe-3-G}
\newcommand\tPiThree{\(\pi^3\)}
\newcommand\tUniKThreeD{UniK3D-L}
\newcommand\tVGGT{VGGT-1B}
\newcommand\tVGGTOmega{VGGT-\(\Omega\)-1B}
\newcommand\tVideoDepthAnything{VDA-L}
\newcommand\tInfiniDepth{InfiniDepth}
\newcommand\tUniDepthTwo{UniDepthV2-L}
\newcommand\tVeloDepth{VeloDepth}
\newcommand\tFoundationGeo{FoundationGeo}
\newcommand\tLapis{Lapis\,\textit{(log)}}
\newcommand\tMetricAnything{MetricAnything}
\newcommand\tPixelPerfectDepth{Pixel-Perfect\,Depth\,\textit{(log)}}
\newcommand\tMarigoldVTwo{Marigold\,V2\,\textit{(log)}}

\newcommand\tOursS{\oursname-S\xspace}
\newcommand\tOursB{\oursname-B\xspace}
\newcommand\tOursL{\oursname-L\xspace}
\newcommand\tOursH{\oursname-H\textsuperscript{+}\xspace}

\newcommand\trefGT{Ground-truth depth}
\newcommand\trefBlur{\oursname-H\textsuperscript{+}, blurred ($\sigma$ = 2\% width)}

\renewcommand{\ra}[1]{\renewcommand{\arraystretch}{#1}}

\robustify\bfseries

\DeclareMathSymbol{@}{\mathord}{letters}{"3B}

\newcommand\timess{\mathbin{\!\times\!}}

\newcommand\mypara[1]{\textbf{#1}}

\makeatletter
\def\onedot{\futurelet\@let@token\@onedot}
\def\@onedot{\ifx\@let@token.\else.\null\fi\xspace}
\makeatother
\def\eg{\textit{e.g}\onedot}

\newcommand\velodepth{VeloDepth\xspace}
\newcommand\depthproname{Depth Pro}
\newcommand\depthpro{\depthproname\xspace}
\newcommand\vggtomeganame{VGGT-\(\Omega\)}
\newcommand\vggtomega{\vggtomeganame\xspace}
\newcommand\pithree{\(\pi^3\)\xspace}
\newcommand\dathreename{Depth Anything 3}
\newcommand\dathree{\dathreename\xspace}
\newcommand\mogetwoname{MoGe-2}
\newcommand\mogethreename{MoGe-3}
\newcommand\mogetwo{\mogetwoname\xspace}
\newcommand\vdaname{Video Depth Anything}
\newcommand\vda{\vdaname\xspace}
\newcommand\unikthreedname{UniK3D}
\newcommand\unikthreed{\unikthreedname\xspace}
\newcommand\unidepthtwoname{UniDepthV2}
\newcommand\metricanything{MetricAnything\xspace}
\newcommand\infinidepth{InfiniDepth\xspace}
\newcommand\depthcrafter{DepthCrafter\xspace}

\newcommand\lotustwoname{Lotus-2}
\newcommand\lotustwo{\lotustwoname\xspace}
\newcommand\marigoldvtwo{Marigold V2\xspace}
\newcommand\lapis{Lapis\xspace}
\newcommand\pixelperfectdepth{Pixel-Perfect Depth\xspace}

\newcommand\dathreeg{\dathreename-G\xspace}
\newcommand\dathreeng{\dathreename-NGL\xspace}
\newcommand\mogetwol{\mogetwoname-L\xspace}
\newcommand\mogethree{\mogethreename\xspace}
\newcommand\mogethreeg{\mogethreename-G\xspace}
\newcommand\mogethreel{\mogethreename-L\xspace}

\newcommand\unikthreedl{\unikthreedname-L\xspace}
\newcommand\unidepthtwol{\unidepthtwoname-L\xspace}

\newcommand\realbokeh{RealBokeh\xspace}
\newcommand\bokehme{BokehMe\xspace}

\newcommand\sintel{Sintel\xspace}
\newcommand\spring{Spring\xspace}
\newcommand\dydtof{DyDToF\xspace}

\newcommand\bonn{Bonn\xspace}
\newcommand\scannet{ScanNet\xspace}

\newcommand\kitti{KITTI\xspace}
\newcommand\eththreed{ETH3D\xspace}
\newcommand\diode{DIODE\xspace}

\newcommand\hdrtumdynamics{TUM-D}
\newcommand\hdrmiddleburyf{M.bury}
\newcommand\hdribims{iBims}
\newcommand\hdrbooster{Booster}
\newcommand\hdrnuscenes{nuSc}
\newcommand\hdrhammer{HAMMER}
\newcommand\hdrdiode{DIODE}

\newcommand\hdravgrank{\smash{\raisebox{\dimexpr\ht\csname @arstrutbox\endcsname
  +\dp\csname @arstrutbox\endcsname+\aboverulesep+\belowrulesep+\cmidrulewidth\relax}%
  {Avg.\ rank$\downarrow$}}}

\newcommand\dinovtwo{DINOv2\xspace}
\newcommand\dinovthree{DINOv3\xspace}

\newcommand\tartanair{TartanAir\xspace}
\newcommand\hypersim{Hypersim\xspace}
\newcommand\bedlam{BEDLAM\xspace}
\newcommand\matrixcity{MatrixCity\xspace}
\newcommand\eden{EDEN\xspace}
\newcommand\dynamicreplica{Dynamic Replica\xspace}
\newcommand\sceneflow{SceneFlow\xspace}
\newcommand\megasynth{MegaSynth\xspace}
\newcommand\pointodyssey{PointOdyssey\xspace}
\newcommand\urbansyn{UrbanSyn\xspace}

\newcommand\px[0]{\text{px}}

\usepackage{flafter}

\title{Depth Pro 2: Sharp and Temporally Consistent Metric Monocular Depth}

\author{Peiyun Hu}
\author{Marko Mihajlovic}
\author{Lyndon Duong}
\author{Lahav Lipson}
\author{Wei Dong}
\author{Ama\"{e}l Delaunoy}
\author{Stephan R. Richter}
\author{Vladlen Koltun}
\affiliation{Apple}

\let\mlrabstract\abstract
\RenewDocumentEnvironment{abstract}{+b}{\mlrabstract{#1\codeline}\global\let\abstractlist\abstractlist}{}
\begin{abstract}
We present Depth Pro 2, an approach to estimating dense metric depth for images and video at full resolution. By analyzing prior methods for monocular depth estimation, we find that they commonly trade off temporal consistency and boundary sharpness: multi-frame methods favor consistency, single-frame methods favor sharpness. To overcome this trade-off, we train a single, scalable network that processes both images and video and predicts depth maps at native resolution. Depth Pro 2 produces depth maps that are metrically accurate, sharp, and temporally consistent all at once, outperforming prior work in all of these dimensions. We release a family of models that define a new Pareto frontier in terms of metric accuracy, sharpness, temporal consistency, and runtime.

\end{abstract}

\newcommand{\codeurl}{\url{https://apple-aiml-research.github.io/ml-depth-pro-2/}}
\newcommand{\codeline}{\par{\centering\large\codeurl\par}\vskip-1.cm}
\newcommand{\website}{\codeurl%
}

\begin{document}

\maketitle

\input{tex/figures/qualgrid}

\example{teaser005}{005}{0.55}{0.70}{0.75}{0.80}
\example{teaser166}{166}{0.55}{0.70}{0.75}{0.80}
\example{teaser306}{306}{0.67}{0.55}{0.87}{0.65}

\begin{figure*}[h]
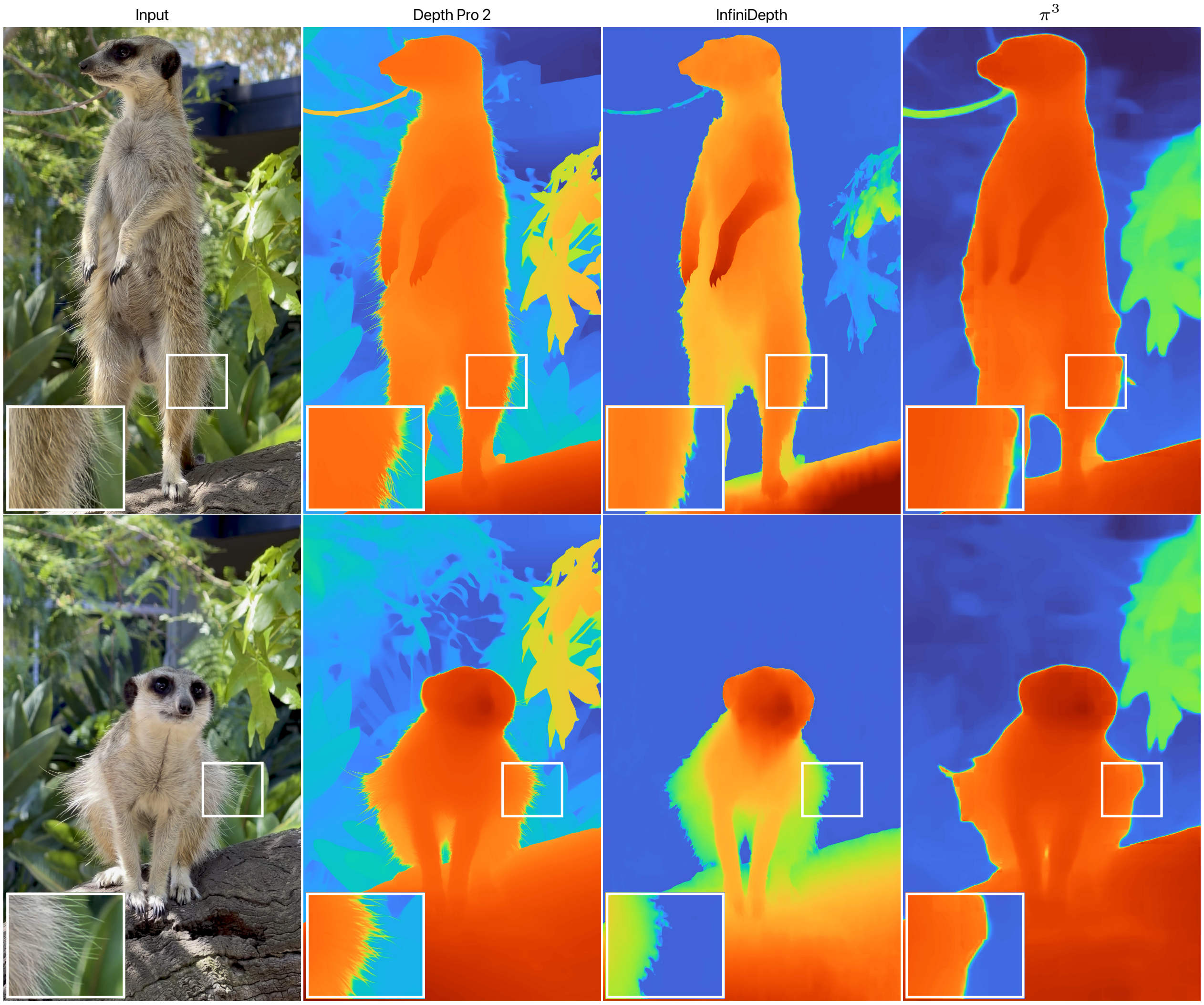

  \vspace{-8mm}
  \centering
  \renewcommand{\qualdir}{data/teaser/15194944}
  \methods{
    \method{source.jpeg}{Input}
    \method{depthpro2.jpeg}{\ours{}}
    \method{infini.jpeg}{\tInfiniDepth}
    \method{pi3.jpeg}{\pithree}
  }
  \qualfigure[rows=examples, zoom=bottom left, inset=0.4, inset margin=1pt,
              row gap=0.5pt, column gap=1pt, height=0.65\textheight,
              zoom border=true, qual lines,
              box color=white, box width=1pt, box outline=none, zoom border color=white,
              crop top=0.08]{
              teaser166,
              teaser306
              }
  \vspace{-6mm}
  \caption{%
  Left to right: two frames from a video~\citep{Pexels} with depth maps produced by Depth Pro 2 (our approach), \tInfiniDepth{}~\citep{Yu2026:CVPR}, and \pithree{}~\citep{Wang2026:ICLR}. InfiniDepth recovers fur but at inconsistent depth. \pithree{} is temporally consistent but blurs boundaries. Depth Pro 2 delivers fine details and consistent depth between frames. Note the insets.
  }
  \label{fig:teaser}
  \vspace{-3mm}
\end{figure*}

\section{Introduction}
\label{sec:intro}

Consider the meerkat (Figure~\ref{fig:teaser}). You may have recorded it on a recent holiday and are watching it now on your high-resolution screen. While you admire the sharp details of the meerkat, you find the background leaves distracting and regret not having recorded the video at a shallower depth of field. Applications that synthetically blur the background could help, but to reliably separate the meerkat from its background, they require a depth map for every frame~\citep{Bertalmio2004, Peng2022BokehMe}.

Picking a suitable method for estimating depth from monocular videos is surprisingly difficult. Methods designed for single images can provide high-resolution depth maps with sharp occlusion boundaries. \infinidepth{}~\citep{Yu2026:CVPR}, for instance, reasonably delineates the fur (third column of Figure~\ref{fig:teaser}). Yet its predictions are temporally inconsistent as it processes each frame separately and places the meerkat at different distances over time.
A method designed for video, such as \pithree{}~\citep{Wang2026:ICLR}, reliably places the meerkat at a consistent distance across frames (last column of Figure~\ref{fig:teaser}). But it operates at a lower resolution than the input video, which smooths over the fur.

\begin{wrapfigure}{R}[0pt]{0.55\linewidth}
\vspace{-6mm}
\centering
\includegraphics[width=\linewidth]{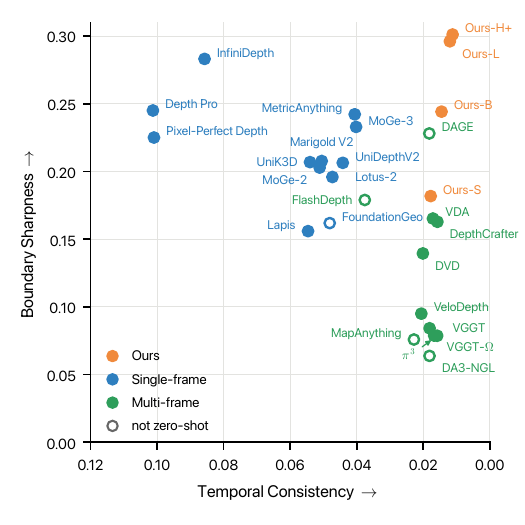}
\vspace{-8mm}
\caption{%
\textbf{Sharpness versus temporal consistency.}\\
\groupours{Depth Pro 2} outperforms prior work operating on a~\groupsingle{single frame} or~\groupmulti{multiple frames}
in terms of sharpness (Boundary F1 score~\citep{Bochkovskii2025:ICLR}) and temporal consistency (Jitter, Sec.~\ref{sec:appendix-jitter}). Results averaged over \sintel{} and \spring{}.
}
\label{fig:tradeoff}
\end{wrapfigure}

This trade-off between sharp boundaries and temporal consistency pervades monocular depth estimation. Figure~\ref{fig:tradeoff} presents the state of the art on boundary sharpness and temporal consistency, measured over the Sintel and Spring datasets~\citep{Butler2012:ECCV,Mehl2023:CVPR}. (We choose these datasets for their dense, high-resolution ground truth on video sequences.) We find that single-image methods cluster at the top left and multi-frame methods cluster at the bottom right. That is, depth maps from single-image methods feature the sharpest boundaries but are unstable over time, while depth maps from multi-frame methods are stable but overly smooth.
Multi-frame approaches are commonly optimized for stability through temporal attention over many frames~\citep{Wang2025:CVPR, Wang2026:ICLR}, video diffusion priors~\citep{Hu2025, Shao2025:CVPR, Yang2025:ICLR, Zhang2026:ICMLW}, and video training data~\citep{Wang2023:ICCV}. This biases the methods towards longer temporal context and costly diffusion, which require substantial resources (memory and runtime) and necessitate operating at reduced resolution, thus oversmoothing boundaries.

In this paper, we present \ours, a method that estimates sharp, metric, and temporally consistent depth from monocular images and video.
\ours outperforms both single-image and multi-frame methods in boundary sharpness and temporal consistency (see Figure~\ref{fig:tradeoff}), as well as in metric depth accuracy (see Section~\ref{sec:experiments}).
The key idea is an architecture that processes a flexible number of frames jointly at their native resolution. The joint processing stabilizes depth temporally. Operating at native resolution preserves sharp boundaries.
In contrast to training with complex multi-stage curricula~\citep{Chen2025:CVPR,Chou2025:ICCV,Lin2026:ICLR,Wang2023:ICCV}, we train \ours on clips of 1 to 32 frames, mixing image, video, and multi-view datasets. By training on single images alongside video, \ours learns to infer metric scale from individual frames. By training on videos with a temporal consistency loss, it learns to keep its estimates stable across frames.

\ours predicts inverse depth normalized by the focal length, and uses the focal length of input images to recover metric depth~\citep{Bochkovskii2025:ICLR, Yin2023:ICCV}. To reliably estimate metric depth when the focal length is not provided, we train a focal length estimator, following Depth Pro~\citep{Bochkovskii2025:ICLR}. In contrast to Depth Pro, however, we design our focal length estimator to be lightweight to facilitate streaming video processing.
Our focal length estimator runs at 118 fps at 1080p on a B200\ and is more accurate than prior estimators.

To support a broad set of applications that may vary in the relative importance of accuracy and latency, we release a family of models ranging from 24M to 872M parameters.
In the streaming setting, where frames arrive one at a time, our smallest model (\tOursS) runs at 53 fps at 1080p on a B200 (Figure~\ref{fig:runtime}), faster than the 30 fps at which most videos are recorded. Our largest model (\tOursH) runs at 8 fps on 1080p video and outperforms all prior work in terms of metric depth, boundary sharpness, and temporal consistency, on both images and video (Section~\ref{sec:experiments}).

\section{Related work}
\label{sec:related}

\mypara{Zero-shot metric depth.}
Early work trained on single-domain datasets with known camera intrinsics for domain-specific metric depth prediction~\citep{Eigen2014:NIPS,Eigen2015:ICCV, Liu2016:TPAMI}.
\citet{Ranftl2022,Ranftl2021} pioneered training on diverse cross-domain mixtures of datasets for generalist monocular depth estimation. While this substantially increased generality and enabled zero-shot deployment across domains, their formulation relaxed the estimated depth from metric to relative.
Subsequent work focused on recovering metric depth while retaining strong generalist performance~\citep{Yin2023:TPAMI, Yin2023:ICCV}. One line of work developed approaches for handling arbitrary camera models~\citep{Piccinelli2024:CVPR, Piccinelli2025:PAMI, Piccinelli2025:CVPR}.
Another predicted point clouds to supervise via depth-affine losses~\citep{Wang2025:CVPR2, Wang2025:NeurIPS, Kong2026:arxiv}. MetricAnything pretrained on large-scale diverse depth completion to forego reasoning about camera intrinsics~\citep{Ma2026:ECCV}.
Depth Pro demonstrated strong metric depth estimation through a separate field-of-view estimator, which can be trained without any depth supervision~\citep{Bochkovskii2025:ICLR}.
We build on the Depth Pro approach and train a lightweight field-of-view estimator that outperforms prior work while substantially reducing the estimator's size to enable accurate field-of-view and metric depth estimation at high frame rates.

\mypara{Sharp depth.}
A line of work processes images in tiles at multiple scales to obtain sharper depth predictions~\citep{Bochkovskii2025:ICLR,Li2024:CVPR,Li2024:ECCV,Li2026:ICLR,Miangoleh2021:CVPR}.
\citet{Kong2026:arxiv} lift point estimates into sparse voxels and apply sparse 3D convolutions to iteratively refine local geometry and sharpness.
\citet{Yu2026:CVPR} represent depth as a neural implicit field that can be queried at arbitrary resolution, but require sparse metric depth as additional input to predict dense metric depth.
\citet{He2026:arxiv, Ke2024:CVPR, Pavlovic2026:SIGGRAPHAsia} leverage latent diffusion priors and synthetic training data for predicting sharp depth boundaries. To avoid artifacts from decoding latents, \citet{Xu2025:NeurIPS}
denoise in pixel space through a cascade of diffusion transformers.
In contrast to prior work that combines multiple scales or iteratively refines predictions, \ours predicts sharp boundaries at native resolution via a single pass through a vision transformer.

\mypara{Stable depth.}
Single-image methods are inconsistent when applied to individual video frames separately. To produce consistent depth maps over video sequences, researchers have proposed specialized modules~\citep{Piccinelli2026:3DV,Chou2025:ICCV,Wang2025:CVPR3,Zhang2019}, plug-and-play stabilization~\citep{Wang2023:ICCV}, and spatiotemporal heads~\citep{Chen2025:CVPR}.
Another line of work fine-tunes video diffusion models for temporally consistent depth prediction~\citep{Hu2025,Shao2025:CVPR,Yang2025:ICLR}, or adapts them into deterministic single-step regressors~\citep{Zhang2026:ICMLW}.
Instead of relying on custom modules or computationally heavy video diffusion priors, we adapt a single-image vision transformer (ViT)~\citep{Dosovitskiy2021:ICLR} to handle a flexible number of frames and penalize inconsistency between predictions during training.

Another line of work takes inspiration from classic 3D reconstruction and optimizes depth jointly with camera poses~\citep{Wang2024:CVPR, Wang2025:CVPR, Wang2026:CVPR, Wang2026:ICLR, Lin2026:ICLR}.
To improve the consistency of reconstructed geometry, these methods are commonly optimized for handling large sets of views, which in turn limits their ability to process images at high resolution, a key requirement for producing sharp depth maps.

In contrast, \ours estimates metric depth from a flexible number of input frames and processes them at native resolution to produce significantly sharper occlusion boundaries.
To enable processing of multiple images, we simply adapt a subset of the self-attention layers of our transformer to attend over tokens from multiple views without introducing additional parameters.
Variants of this idea have been used in video transformers~\citep{bertasius2021space, arnab2021vivit}, multi-view recognition~\citep{Chen2021:BMVC, Lee2025:ICLR}, and generation~\citep{khachatryan2023text2video, Shi2024:ICLR}.
Most closely related, \dathree{}~\citep{Lin2026:ICLR} rearranges tokens so that pretrained DINOv2 attention operates across views for feed-forward reconstruction.
Unlike these approaches, we design our network to predict metric depth at native resolution.

\section{Method}
\label{sec:method}

\subsection{Architecture}
\label{sec:architecture}
\begin{figure}[t]
    \centering
    \includegraphics[width=1\linewidth,trim={142pt 244pt 170pt 146pt},clip]{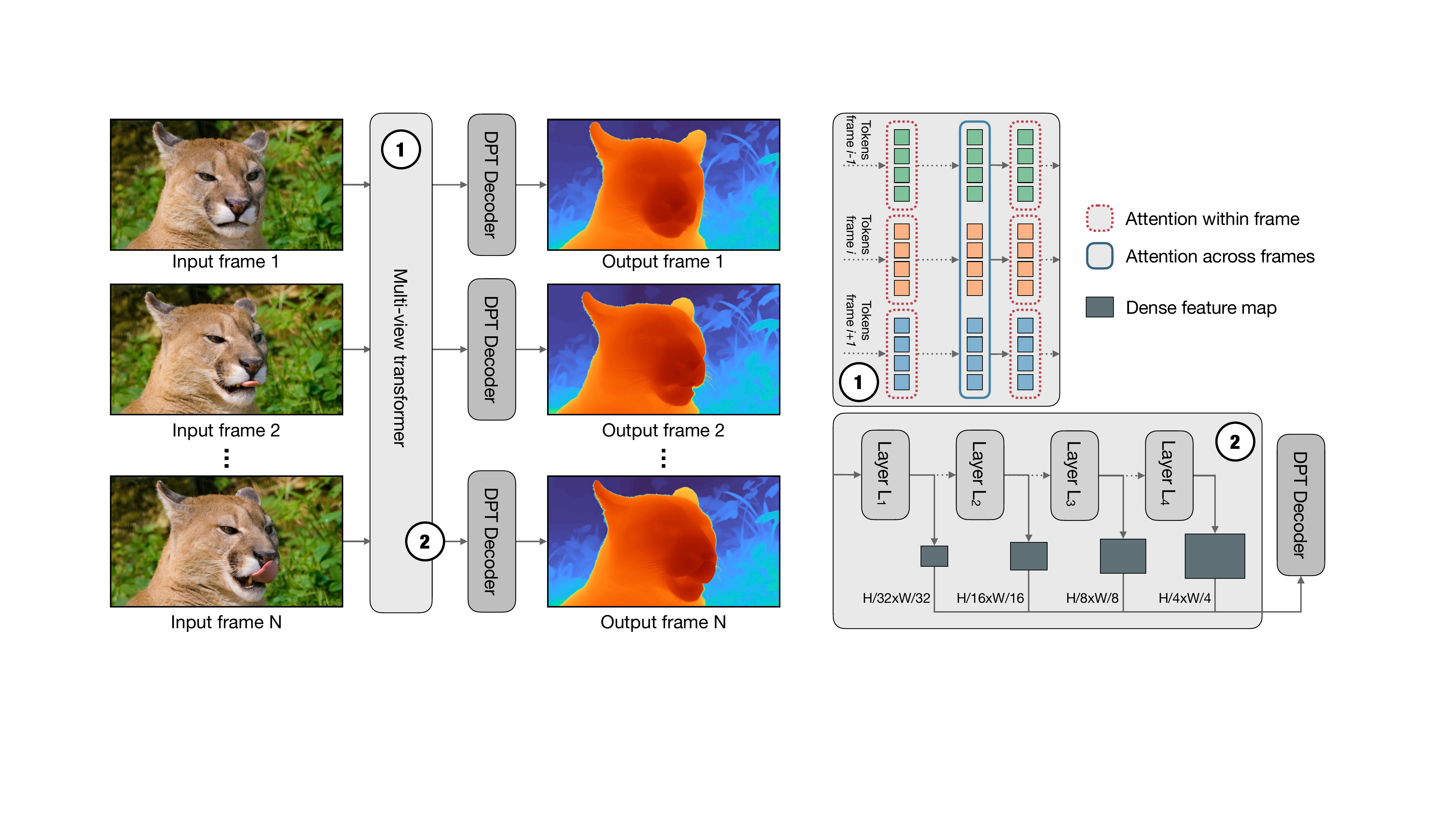}
    \caption{\textbf{\ours Architecture.}
    Our depth estimation network consists of a multi-view transformer that ingests one or multiple input frames and processes them jointly, and a DPT decoder, which predicts an inverse depth map per input frame. \circled{1} The transformer alternates attention within and across frames. \circled{2} Features for fusion by the DPT decoder are extracted at intermediate layers. Table~\ref{tab:model-family-params} details layer configurations for each model of the Depth Pro 2 family.
    }
    \label{fig:architecture}
    \vspace{-3mm}
\end{figure}

\mypara{Overview.}
The architecture of \ours is depicted in \Fig.~\ref{fig:architecture}. It consists of a multi-view transformer that ingests a single image or multiple frames of a video, processes them jointly, and decodes them to an inverse depth map per input image or video frame.
Following Depth Pro, our depth estimation network predicts inverse depth normalized by focal length, and we combine it with a known or estimated focal length to predict metric depth~\citep{Bochkovskii2025:ICLR}.
A separate field-of-view estimator processes the input image(s) and predicts the field-of-view per image, which we convert to the focal length. If the focal length is known a priori, e.g., through EXIF data, we use that instead of running the field-of-view estimator.

\mypara{Lightweight multi-frame depth estimation network.}
We design our depth estimation network to be lightweight and to flexibly handle a variable number of images, starting from a ViT variant pretrained for single-image processing~\citep{Dosovitskiy2021:ICLR, Simeoni2026:TMLR}.
Since temporal stability requires an architecture that aligns predictions across frames, our model shares information across jointly processed frames via cross-attention, alternating blocks of within-frame self-attention and cross-frame attention~\citep{Wang2025:CVPR,Lin2026:ICLR}. (See \Fig~\ref{fig:architecture}, \circled{1}.)
Since early blocks in multi-view transformers tend to encode low-level features and only later blocks approximate feature matching~\citep{Stary2026:NeurIPS}, we restrict the first $1/3$ of the $L$ transformer blocks to self-attention operating only over tokens from the same frame for higher efficiency. The remaining $2L/3$ blocks alternate between within- and cross-frame attention.
When encoding spatial positions of tokens processed by the transformer, we specifically omit any encoding of frame order, making our network equivariant to frame permutation.
This allows flexible processing of any number of frames and directly reusing the attention weights from single-image pretrained encoder checkpoints~\citep{Lin2026:ICLR} without adding transformer blocks that need to be trained from scratch for multi-view processing.

\mypara{Native resolution processing.}
Boundary sharpness correlates with the resolution at which frames are processed~\
\citep{Miangoleh2021:CVPR,Bochkovskii2025:ICLR,Yu2026:CVPR}. We therefore design \ours to operate at the full native resolution of the input image or video. To this end, we initialize our multi-view transformer from DINOv3 weights, as DINOv3 is trained specifically to produce dense, high-resolution features~\citep{Simeoni2026:TMLR}.
An input image is split into patches of size
16, and each patch is embedded as a token. We encode each token's spatial position through a 2D rotary embedding over their coordinates. Instead of normalizing coordinates per axis~\citep{Simeoni2026:TMLR}, which would discard any information about the image's aspect ratio, we divide both axes by the longer side. The position $(x,y)$ of a patch $(i, j)$ within a grid of size $h\times w$ is thus
\[
y' = 2(i+\tfrac{1}{2})/\max(h, w) - 1, \quad x' = 2(j+\tfrac{1}{2})/\max(h, w) - 1.
\]

\mypara{DPT decoder.}
A DPT decoder~\citep{Ranftl2021} decodes features extracted per frame into inverse depth maps. (See \Fig~\ref{fig:architecture}, \circled{2}.)
Following DPT, we extract tokens after several transformer blocks, assemble them into feature maps at different resolutions, and project them to a common channel dimension.
Depending on the size of the multi-view transformer, we adjust the positions at which we extract features, and scale the number of channel dimensions the decoder operates on. See Table~\ref{tab:model-family-params} for the exact specification.
To upsample feature maps within the decoder, we replace bilinear upsampling with transposed convolutions~\citep{Bochkovskii2025:ICLR}.

\mypara{Focal length estimation.}
Our multi-view transformer predicts inverse depth normalized by the focal   length~\citep{Bochkovskii2025:ICLR}. Converting this prediction to metric depth therefore requires the focal length. The focal length is often recorded in EXIF metadata; when it is not, we estimate it with a separate, lightweight vision transformer. This estimator predicts the field of view along the shorter image side, from which the focal length follows given the length of that side in pixels. To minimize the overhead, we build the estimator on DINOv3 ViT-S/16~\citep{Simeoni2026:TMLR} and fine-tune it end-to-end on Flickr photographs with EXIF focal lengths~\citep{Thomee2016:CACM}. The estimator has 23.5M parameters and, for a 1080p frame, adds about 8\%, 3\%, 1\%, and 0.5\% to the compute of \ourss, \oursb, \oursl, and \oursh, respectively.
For multi-view input, we assume a constant focal length and take the median of the per-frame estimates.
On the zero-shot benchmark of~\citet{Bochkovskii2025:ICLR}, our estimator attains a mean $\delta_{25}$ of 88.6\%, compared to 71.7\% for \depthpro{} (Table~\ref{tab:focal-photo}).
Appendix~\ref{sec:appendix-focal-setup} describes the training setup,   Appendix~\ref{sec:appendix-focal-design} analyzes design choices, and   Table~\ref{tab:focal-photo} compares the estimator to prior work.

\subsection{Training}
\label{sec:training}
\mypara{Training data.}
Depth Pro 2 is trained on a diverse set of 19 datasets with a strong focus on synthetic ground truth. See Table~\ref{tab:training-data} for the full list.
To preserve the sharpness of ground-truth depth maps, we avoid any interpolation of ground-truth depth during training, as averaging over depth discontinuities breeds flying pixels~\citep{Bochkovskii2025:ICLR}.
To expose our model to a broad set of image sizes and aspect ratios, we crop and sample images at varying strides.
During training, we use ground-truth camera intrinsics (focal length) and adjust accordingly depending on how we crop or subsample.
Controlled experiments in the supplement (Section~\ref{sec:appendix-controlled}) show that training without resampling improves boundary accuracy by more than 5 percentage points over Depth Pro's resizing.

To handle both images and video, Depth Pro 2 is trained on clips with varying length, of which all frames are processed jointly. Images are treated as clips of length 1; for video and multi-view datasets, a clip contains up to 32 frames. (We use \emph{frame} for both video frames and views.) For each clip, we draw the source dataset~$i$ with probability $p_i \propto n_i^{1/T}$, where $n_i$ is the size of dataset~$i$ and $T=2$~\citep{arivazhagan2019massively}. To avoid overfitting to small datasets, we cap their sampling weight.

\mypara{Batching.}
To efficiently train with varying resolutions and clip lengths, we batch by tokens. Each GPU receives a fixed budget of $73{,}728$ tokens, \eg, 18 frames at  $1024\times1024$.
When a clip leaves part of the budget unused, we fill the remainder with further clips of the same resolution.

\mypara{Training objective.}
Our loss is the unweighted sum of four terms:
\begin{equation}
  \mathcal{L} = \mathcal{L}_{\mathit{mae}} + \mathcal{L}_{\mathit{ssi}} + \mathcal{L}_\mathit{mge} + \mathcal{L}_\mathit{mvc}.
  \label{eq:loss}
\end{equation}
$\mathcal{L}_{\mathit{mae}}$ is a trimmed $\mathcal{L}_1$ loss on focal-normalized inverse depth. It anchors metric depth, and we apply it only to sources with metric ground truth. $\mathcal{L}_{ssi}$ is a trimmed $\mathcal{L}_1$ loss after normalizing prediction and ground truth independently to zero median and unit mean absolute deviation~\citep{Ranftl2022}. We apply it to all datasets. Both trimmed losses discard, per image, the fraction $\tau$ of pixels with the largest loss, with $\tau = 1\%$ on synthetic data and $\tau = 10\%$ on real data, since the noise level on real data is higher.  The multi-scale gradient   term $\mathcal{L}_{\mathit{mge}}$ matches first-order gradients of the normalized depth maps over a Gaussian pyramid of $M=3$ scales~\citep{Bochkovskii2025:ICLR}. These gradients are reliable only where depth boundaries are exact, so we apply $\mathcal{L}_{\mathit{mge}}$ only to   synthetic data. Next we describe the multi-view consistency term $\mathcal{L}_{\mathit{mvc}}$.

\mypara{Multi-view self-consistency.}
To stabilize predictions across frames, we introduce a multi-view consistency loss $\mathcal{L}_{\mathit{mvc}}$. Our formulation is inspired by \citet{Luo2020:TOG}, who warp points via their predicted depth and penalize deviations from estimated optical flow. For robustness, we replace optical flow with correspondences given by ground-truth camera poses.
Let $p$ be a pixel in frame $i$ with predicted depth $\hat{Z}_i(p)$, and let $\pi_i$ denote the projection from world space to the image plane of frame $i$.
We unproject $p$ using its predicted depth and reproject it into frame $j$ using the ground-truth poses, which yields the pixel location $p'$ and depth $z'$:
\begin{equation*}
    (p', z') = \pi_j\bigl(\pi_i^{-1}(p, \hat{Z}_i(p))\bigr).
  \end{equation*}
In a static scene, $z'$ should match the prediction $\hat{Z}_j(p')$ in frame $j$.
We penalize deviations in inverse depth over a random subset of
  pixels $\mathcal{P}$:
  \begin{equation}
    \mathcal{L}_{\mathit{mvc}} = \frac{1}{\lvert\mathcal{P}\rvert} \sum_{p \in \mathcal{P}} \bigl\lvert 1/z' - 1/\hat{Z}_j(p') \bigr\rvert_{1^\tau}.
    \label{eq:mvc}
  \end{equation}
Occlusions and moving objects violate the static-scene assumption.
To exclude them, we warp each pixel using its ground-truth depth and discard   pixels whose warped depth differs from the ground-truth depth at the target location by more than 2\%. This allows us to apply $\mathcal{L}_{\mathit{mvc}}$ to dynamic scenes as well.

\subsection{Inference}
\label{sec:long-video-inference}
Depth Pro 2 is trained on clips of up to 32 frames but at inference time processes videos of arbitrary length. It supports two modes. Buffered streaming accumulates frames into clips and processes each clip with full attention across its frames. Unbuffered streaming produces a depth map for each frame as soon as it arrives, for applications that require minimal latency.

\mypara{Buffered streaming.}
For buffered streaming, we divide the video into clips of $N$ frames, with consecutive clips overlapping by $K$ frames, and process each clip independently. The overlap keeps predictions consistent across clip boundaries. Within the overlap, we blend the two predictions with weights that change linearly with frame position.
Section~\ref{sec:appendix-long-video} reports controlled experiments on the choice of $N$ and $K$.

\mypara{Unbuffered streaming.}
Unbuffered streaming minimizes per-frame latency. In this mode, \ours processes frames sequentially with causal cross-attention: each frame attends only to itself and to preceding frames. The keys and values of past frames remain fixed as new frames arrive, so we compute them once and keep them in a first-in-first-out (FIFO) cache. Each frame attends to a window of $C$ frames: itself and the $C-1$ frames before it, the keys and values of which are cached. Only the cross-frame attention layers use this cache; the per-frame layers operate on the current frame alone.

\subsection{Model family}
\label{sec:model-family}

The Depth Pro 2 family comprises four models, each built on a different pretrained DINOv3 backbone. Table~\ref{tab:model-family-params} details the respective backbone and DPT configurations.
\begin{table}[h]
    \centering
    \footnotesize
    \setlength{\tabcolsep}{4pt}
    \caption{
    \textbf{Architecture specifications of the \ours{} model family.} The backbones are initialized from DINOv3 ViT-S/16, ViT-B/16, ViT-L/16, and ViT-H+/16, respectively. Parameter counts exclude the focal-length estimator. $L$ is the number of transformer blocks in the backbone and $D$ is their token channel dimension. `Cross-frame blocks' lists the indices of the backbone blocks that use cross-frame attention. `Reassembled blocks' lists the indices of blocks whose tokens are passed to the DPT decoder, which reassembles them into feature maps with $D_s$ channels at $1/s$ of the input resolution. $\hat D$ is the channel dimension of the decoder's fusion stages, to which all feature maps are projected. Block indices are zero-based. All variants use patch size \(16 \times 16\).
    }
    \label{tab:model-family-params}
    \fitwidth{
    \begin{tabular}{lr@{\hspace{16pt}}rcrc@{\hspace{16pt}}rccr}
        \toprule
        & & \multicolumn{4}{c}{Backbone} & \multicolumn{4}{c}{DPT decoder} \\
        \cmidrule(r){3-6}\cmidrule{7-10}
        Model & Total params & Params & $L$ & $D$ & Cross-frame blocks & Params & Reassembled blocks & $D_s$ ($s=4,8,16,32$) & $\hat D$ \\
        \midrule
        \tOursS & 24.3M & 21.6M & 12 & 384 & $5, 7, 9, 11$ & 2.7M & $2, 5, 8, 11$ & $48, 96, 192, 384$ & 64 \\
        \tOursB & 96.5M & 85.6M & 12 & 768 & $5, 7, 9, 11$ & 10.9M & $2, 5, 8, 11$ & $96, 192, 384, 768$ & 128 \\
        \tOursL & 334.0M & 303.1M & 24 & 1024 & $9, 11, \dots, 23$ & 30.9M & $5, 11, 17, 23$ & $256, 512, 1024, 1024$ & 256 \\
        \tOursH & 872.1M & 840.5M & 32 & 1280 & $11, 13, \dots, 31$ & 31.6M & $7, 15, 23, 31$ & $256, 512, 1024, 1024$ & 256 \\
        \bottomrule
    \end{tabular}
    }
\end{table}

\section{Experiments}
\label{sec:experiments}
\begin{table}[t]
 \centering\ra{1.05}\footnotesize\setlength{\tabcolsep}{1pt}
 \caption{
 \textbf{Zero-shot metric depth accuracy on images.}
 All methods receive individual frames. On video and multi-view datasets, each of the first 50 frames of a sequence is scored separately.
 We report $\delta_1$ (higher is better) under ground-truth and predicted intrinsics, rank methods per condition and dataset, and average the ranks over datasets (last column).
 Non-zero-shot results are \textcolor{gray}{gray}. $^{\ddagger}$MetricAnything uses GT intrinsics only.
 }
 \label{tab:paper-metric-image}
 \fitwidth{\begin{tabular}{lcc@{\hspace{5pt}}cc@{\hspace{5pt}}cc@{\hspace{5pt}}cc@{\hspace{5pt}}cc@{\hspace{5pt}}cc@{\hspace{5pt}}cc@{\hspace{5pt}}cc@{\hspace{5pt}}cc@{\hspace{5pt}}cc@{\hspace{5pt}}cc@{\hspace{6pt}}|@{\hspace{6pt}}cc}
  \toprule
  & \multicolumn{8}{c}{Real, indoor} & \multicolumn{4}{c}{Real, indoor+outdoor} & \multicolumn{4}{c}{Real, outdoor} & \multicolumn{6}{c}{Synthetic} & \\
  \cmidrule(lr){2-9}\cmidrule(lr){10-13}\cmidrule(lr){14-17}\cmidrule(lr){18-23}
  Method & \multicolumn{2}{c}{\hdribims} & \multicolumn{2}{c}{\hdrmiddleburyf} & \multicolumn{2}{c}{\hdrbooster} & \multicolumn{2}{c}{\hdrhammer} & \multicolumn{2}{c}{\eththreed} & \multicolumn{2}{c}{\hdrdiode} & \multicolumn{2}{c}{\kitti} & \multicolumn{2}{c}{\hdrnuscenes} & \multicolumn{2}{c}{\sintel} & \multicolumn{2}{c}{\spring} & \multicolumn{2}{c}{\dydtof} & \multicolumn{2}{c}{\hdravgrank} \\
  \cmidrule(lr){2-3}\cmidrule(lr){4-5}\cmidrule(lr){6-7}\cmidrule(lr){8-9}\cmidrule(lr){10-11}\cmidrule(lr){12-13}\cmidrule(lr){14-15}\cmidrule(lr){16-17}\cmidrule(lr){18-19}\cmidrule(lr){20-21}\cmidrule(lr){22-23}\cmidrule(l){24-25}
   & GT & pred & GT & pred & GT & pred & GT & pred & GT & pred & GT & pred & GT & pred & GT & pred & GT & pred & GT & pred & GT & pred & GT & pred \\
  \midrule
  \tMapAnything~\citep{Keetha2026} & 74.8 & 58.0 & 34.6 & 42.3 & 41.8 & 50.1 & 13.0 & 29.9 & 54.1 & 51.5 & 28.2 & 18.2 & 91.6 & 90.6 & 57.3 & 34.4 & 20.4 & 18.4 & \grayed{96.7} & \grayed{96.7} & 7.4 & 7.1 & 10.2 & 8.6 \\
  \tDepthPro~\citep{Bochkovskii2025:ICLR} & 88.1 & 82.9 & \cellsecond{72.2} & \cellsecond{75.1} & 56.7 & 59.0 & 71.7 & 55.2 & 46.3 & 41.0 & 45.4 & 40.6 & 84.8 & 82.7 & 56.5 & 60.5 & \cellsecond{36.5} & 30.1 & 12.5 & 16.7 & 67.2 & 27.4 & 7.6 & 6.5 \\
  \tMetricAnything$^{\ddagger}$~\citep{Ma2026:ECCV} & 86.3 & -- & 51.9 & -- & 60.6 & -- & 38.1 & -- & 74.4 & -- & 60.4 & -- & 5.7 & -- & 87.6 & -- & 25.6 & -- & \cellthird{22.5} & -- & 73.3 & -- & 7.6 & -- \\
  \tFoundationGeo~\citep{Liu2026:ECCV-foundationgeo} & 93.1 & 70.0 & 57.3 & 12.9 & 50.0 & 56.2 & 68.7 & 72.4 & 83.2 & 77.3 & 67.3 & 55.1 & 96.3 & \cellthird{96.7} & 87.2 & 79.7 & 11.7 & 21.0 & \grayed{33.4} & \grayed{27.8} & 68.1 & 1.3 & 6.9 & 6.8 \\
  \tMoGeThree~\citep{Kong2026:arxiv} & 92.0 & 72.8 & 58.4 & 41.2 & 56.1 & 43.8 & 67.1 & 67.6 & 90.3 & 80.3 & 65.7 & 50.9 & 94.2 & 56.0 & 89.3 & \cellthird{88.1} & 13.2 & 24.6 & 22.1 & \cellsecond{30.1} & 75.0 & 13.1 & 6.2 & 6.4 \\
  \tMoGe~\citep{Wang2025:NeurIPS} & 89.0 & 77.8 & 57.1 & 46.2 & 57.7 & 43.9 & \cellthird{75.3} & 66.9 & \cellsecond{92.1} & \cellsecond{90.9} & 67.2 & 51.3 & 91.5 & 42.7 & 87.0 & 85.6 & 22.2 & \cellthird{36.1} & \cellfirst{27.2} & 17.7 & 71.0 & 8.1 & 5.6 & 6.2 \\
  \tDepthAnythingThree~\citep{Lin2026:ICLR} & \cellfirst{96.1} & 86.4 & \cellfirst{73.9} & 48.8 & 71.6 & \cellthird{72.0} & 73.9 & \cellsecond{78.4} & 83.7 & 80.2 & \cellthird{68.4} & 59.4 & 79.3 & 50.6 & 83.6 & 49.6 & 21.9 & 19.0 & \grayed{17.2} & \grayed{8.5} & 59.1 & 23.4 & 5.6 & 5.7 \\
  \tDAGE~\citep{Ngo2026:CVPR} & 89.9 & 84.2 & \cellthird{69.6} & \cellthird{69.2} & \cellfirst{76.4} & \cellfirst{74.8} & \cellsecond{76.1} & \cellthird{74.8} & 90.4 & 84.0 & 63.5 & \cellthird{60.8} & 95.8 & 89.1 & 74.8 & 83.2 & 22.2 & 15.7 & \grayed{91.3} & \grayed{93.1} & \cellfirst{82.3} & \cellthird{47.8} & 4.7 & 4.3 \\
  \tUniDepthTwo~\citep{Piccinelli2025:PAMI} & \cellthird{94.5} & \cellfirst{94.6} & 55.7 & 55.3 & \cellsecond{73.6} & 66.8 & 48.0 & 41.3 & 87.5 & 73.9 & 62.6 & 54.3 & \cellthird{97.4} & \cellfirst{97.6} & \cellsecond{92.2} & \cellsecond{91.9} & \cellthird{36.4} & \cellsecond{37.0} & \cellsecond{25.4} & \cellthird{22.8} & \cellthird{80.1} & 41.1 & \cellthird{4.5} & \cellthird{4.0} \\
  \tUniKThreeD~\citep{Piccinelli2025:CVPR} & \cellsecond{95.3} & \cellsecond{94.4} & 59.3 & 51.6 & \cellthird{73.5} & \cellsecond{73.8} & 58.1 & 61.1 & \cellthird{92.0} & \cellthird{84.9} & \cellsecond{72.2} & \cellsecond{67.6} & \cellsecond{98.0} & \cellsecond{97.0} & \cellfirst{92.4} & \cellfirst{92.1} & 31.7 & 31.4 & 19.2 & 19.8 & 76.4 & \cellsecond{61.1} & \cellsecond{3.6} & \cellsecond{3.1} \\
  \midrule
  \tOursH & 94.4 & \cellthird{86.6} & \cellfirst{73.9} & \cellfirst{79.1} & 65.7 & 43.1 & \cellfirst{89.4} & \cellfirst{86.3} & \cellfirst{97.8} & \cellfirst{91.8} & \cellfirst{74.2} & \cellfirst{69.2} & \cellfirst{98.2} & 87.9 & \cellthird{90.9} & 87.7 & \cellfirst{60.2} & \cellfirst{43.4} & 20.5 & \cellfirst{37.0} & \cellsecond{80.5} & \cellfirst{78.2} & \cellfirst{2.4} & \cellfirst{2.7} \\
  \bottomrule
\end{tabular}
\vspace{-3mm}
}
 \vspace{10pt}
 \centering\ra{1.05}\footnotesize\setlength{\tabcolsep}{5pt}
 \caption{
 \textbf{Zero-shot metric depth accuracy on video.}
 All methods receive clips of up to 50 frames. Single-image methods process frames independently.
 We report $\delta_1$ and average rank, as in Table~\ref{tab:paper-metric-image}. $^{\ddagger}$MetricAnything uses GT intrinsics only.
 }
 \label{tab:paper-metric-video}
 \fitwidth{\begin{tabular}{lcc@{\hspace{5pt}}cc@{\hspace{5pt}}cc@{\hspace{5pt}}cc@{\hspace{6pt}}|@{\hspace{6pt}}cc}
  \toprule
  & \multicolumn{2}{c}{Real, outdoor} & \multicolumn{6}{c}{Synthetic} & \\
  \cmidrule(lr){2-3}\cmidrule(lr){4-9}
  Method & \multicolumn{2}{c}{\kitti} & \multicolumn{2}{c}{\sintel} & \multicolumn{2}{c}{\spring} & \multicolumn{2}{c}{\dydtof} & \multicolumn{2}{c}{\hdravgrank} \\
  \cmidrule(lr){2-3}\cmidrule(lr){4-5}\cmidrule(lr){6-7}\cmidrule(lr){8-9}\cmidrule(l){10-11}
   & GT & pred & GT & pred & GT & pred & GT & pred & GT & pred \\
  \midrule
  \tMapAnything~\citep{Keetha2026} & 94.1 & 94.7 & 21.2 & 18.6 & \grayed{97.0} & \grayed{97.0} & 8.4 & 7.5 & 10.0 & 8.7 \\
  \tFoundationGeo~\citep{Liu2026:ECCV-foundationgeo} & 96.3 & 96.7 & 11.7 & 21.0 & \grayed{33.4} & \grayed{27.8} & 68.1 & 1.3 & 9.0 & 8.0 \\
  \tDepthAnythingThree~\citep{Lin2026:ICLR} & 95.7 & 25.8 & 30.3 & 20.3 & \grayed{16.9} & \grayed{11.1} & 77.4 & 32.1 & 5.0 & 9.0 \\
  \tVeloDepth~\citep{Piccinelli2026:3DV} & 84.2 & 74.0 & 27.5 & 30.1 & 22.0 & \cellthird{25.0} & 69.3 & 57.8 & 8.2 & 5.0 \\
  \tDepthPro~\citep{Bochkovskii2025:ICLR} & 84.8 & 82.7 & \cellsecond{36.5} & 30.1 & 12.5 & 16.7 & 67.2 & 27.4 & 7.8 & 6.8 \\
  \tMetricAnything$^{\ddagger}$~\citep{Ma2026:ECCV} & 5.7 & -- & 25.6 & -- & 22.5 & -- & 73.3 & -- & 7.8 & -- \\
  \tMoGeThree~\citep{Kong2026:arxiv} & 94.2 & 56.0 & 13.2 & 24.6 & 22.1 & \cellsecond{30.1} & 75.0 & 13.1 & 7.2 & 6.5 \\
  \tMoGe~\citep{Wang2025:NeurIPS} & 91.5 & 42.7 & 22.2 & \cellthird{36.1} & \cellsecond{27.2} & 17.7 & 71.0 & 8.1 & 7.0 & 7.0 \\
  \tDAGE~\citep{Ngo2026:CVPR} & 96.7 & 84.6 & 28.5 & 22.3 & \grayed{96.8} & \grayed{97.4} & \cellsecond{82.2} & \cellsecond{65.0} & \cellthird{4.0} & 5.3 \\
  \tUniKThreeD~\citep{Piccinelli2025:CVPR} & \cellsecond{98.0} & \cellthird{97.0} & 31.7 & 31.4 & 19.2 & 19.8 & 76.4 & \cellthird{61.1} & 4.5 & \cellthird{3.8} \\
  \tUniDepthTwo~\citep{Piccinelli2025:PAMI} & \cellthird{97.4} & \cellfirst{97.6} & \cellthird{36.4} & \cellsecond{37.0} & \cellthird{25.4} & 22.8 & \cellthird{80.1} & 41.1 & \cellsecond{3.0} & \cellsecond{3.0} \\
  \midrule
  \tOursH & \cellfirst{98.1} & \cellsecond{97.2} & \cellfirst{51.2} & \cellfirst{41.4} & \cellfirst{51.7} & \cellfirst{45.0} & \cellfirst{86.2} & \cellfirst{94.8} & \cellfirst{1.0} & \cellfirst{1.2} \\
  \bottomrule
\end{tabular}
}
\end{table}

\begin{table}[p]
 \centering\ra{1.00}\footnotesize\setlength{\tabcolsep}{4pt}
 \renewcommand{\thesubtable}{(\alph{subtable})}
 \captionsetup[subtable]{labelformat=simple,hypcap=true}
 \caption{
 \textbf{Zero-shot boundary sharpness.} We report the boundary $F1$ scores (higher is better) from \citet{Bochkovskii2025:ICLR} along with average ranks across datasets. Table~\ref{tab:paper-sharpness-single} methods receive a single frame (DepthCrafter and VDA are excluded here as they do not support single-image input, as well as VeloDepth, which reduces to UniK3D on a single frame). Table~\ref{tab:paper-sharpness-video} methods receive 50-frame clips. Non-zero-shot results are \textcolor{gray}{gray}.
 }
 \label{tab:paper-sharpness}
 \begin{subtable}{\linewidth}
 \centering
 \caption{Single image}
 \label{tab:paper-sharpness-single}
 \fitwidth{\begin{tabular}{lccccc@{\hspace{6pt}}|@{\hspace{6pt}}c}
  \toprule
  & \multicolumn{2}{c}{Real, indoor} & \multicolumn{3}{c}{Synthetic} & Avg. \\
  \cmidrule(lr){2-3}\cmidrule(lr){4-6}
  Method & \hdribims & \hdrmiddleburyf & \sintel & \spring & \dydtof & rank$\downarrow$ \\
  \midrule
  \tPiThree~\citep{Wang2026:ICLR} & 6.4 & 0.2 & 10.8 & 4.0 & 7.1 & 19.8 \\
  \tMapAnything~\citep{Keetha2026} & 9.2 & 0.4 & 9.9 & \grayed{5.0} & 7.9 & 19.8 \\
  \tVGGTOmega~\citep{Wang2026:CVPR} & 3.7 & 0.1 & 10.8 & 4.7 & 8.1 & 19.4 \\
  \tDepthAnythingThree~\citep{Lin2026:ICLR} & 8.0 & 3.6 & 7.9 & \grayed{3.4} & 7.6 & 18.2 \\
  \tVGGT~\citep{Wang2025:CVPR} & 7.1 & 0.5 & 12.3 & 4.6 & 8.7 & 18.0 \\
  \tChronoDepth~\citep{Shao2025:CVPR} & 9.7 & 3.6 & 16.2 & 3.9 & 14.3 & 15.8 \\
  \tFoundationGeo~\citep{Liu2026:ECCV-foundationgeo} & 13.6 & 0.8 & 23.9 & \grayed{8.5} & 15.7 & 15.2 \\
  \tDVD~\citep{Zhang2026:ICMLW} & 15.9 & 1.3 & 19.5 & 7.0 & 18.3 & 14.0 \\
  \tLapis~\citep{Liu2026:ECCV} & 13.2 & 8.9 & 22.7 & 8.5 & 11.1 & 12.4 \\
  \tDAGE~\citep{Ngo2026:CVPR} & 19.0 & 1.6 & 31.9 & \grayed{12.6} & 22.2 & 11.5 \\
  \tFlashDepth~\citep{Chou2025:ICCV} & 13.7 & 10.8 & 22.2 & \grayed{13.5} & 18.5 & 11.5 \\
  \tUniDepthTwo~\citep{Piccinelli2025:PAMI} & 15.9 & 2.9 & 34.3 & 7.0 & 23.9 & 11.2 \\
  \tUniKThreeD~\citep{Piccinelli2025:CVPR} & 17.6 & 3.0 & 33.5 & 7.9 & 24.6 & 9.6 \\
  \tLotus~\citep{He2026:arxiv} & 17.6 & 5.7 & 28.5 & 10.7 & 24.4 & 9.0 \\
  \tMoGe~\citep{Wang2025:NeurIPS} & \cellsecond{20.6} & 2.9 & 32.2 & 8.4 & 23.0 & 9.0 \\
  \tMarigoldVTwo~\citep{Pavlovic2026:SIGGRAPHAsia} & 15.0 & \cellthird{16.2} & 27.6 & \cellthird{13.9} & 22.6 & 8.4 \\
  \tPixelPerfectDepth~\citep{Xu2025:NeurIPS} & 20.4 & 6.1 & 34.9 & 10.1 & 27.7 & 5.8 \\
  \tDepthPro~\citep{Bochkovskii2025:ICLR} & 19.5 & 7.9 & 36.3 & 12.7 & 24.8 & 5.4 \\
  \tMoGeThree~\citep{Kong2026:arxiv} & \cellfirst{22.2} & 6.2 & \cellthird{37.1} & 9.5 & 27.5 & 5.0 \\
  \tMetricAnything~\citep{Ma2026:ECCV} & 19.2 & 9.6 & 34.4 & \cellsecond{14.0} & \cellthird{29.4} & \cellthird{4.6} \\
  \tInfiniDepth~\citep{Yu2026:CVPR} & 20.4 & \cellsecond{17.6} & \cellfirst{44.8} & 11.8 & \cellsecond{30.5} & \cellsecond{3.0} \\
  \midrule
  \tOursH & \cellthird{20.5} & \cellfirst{21.9} & \cellsecond{40.2} & \cellfirst{16.3} & \cellfirst{31.6} & \cellfirst{1.6} \\
  \bottomrule
\end{tabular}
}
 \end{subtable}

 \vspace{8pt}
 \begin{subtable}{\linewidth}
 \centering
 \caption{Video}
 \label{tab:paper-sharpness-video}
 \fitwidth{\begin{tabular}{lccc@{\hspace{6pt}}|@{\hspace{6pt}}c}
  \toprule
  & \multicolumn{3}{c}{Synthetic} & Avg. \\
  \cmidrule(lr){2-4}
  Method & \sintel & \spring & \dydtof & rank$\downarrow$ \\
  \midrule
  \tDepthAnythingThree~\citep{Lin2026:ICLR} & 9.0 & \grayed{3.8} & 8.0 & 24.0 \\
  \tMapAnything~\citep{Keetha2026} & 10.1 & \grayed{5.1} & 8.4 & 22.5 \\
  \tPiThree~\citep{Wang2026:ICLR} & 11.6 & 4.2 & 6.8 & 21.7 \\
  \tVGGTOmega~\citep{Wang2026:CVPR} & 10.8 & 4.9 & 7.5 & 21.0 \\
  \tVGGT~\citep{Wang2025:CVPR} & 12.2 & 4.7 & 8.1 & 20.0 \\
  \tVeloDepth~\citep{Piccinelli2026:3DV} & 15.0 & 4.0 & 9.9 & 19.7 \\
  \tChronoDepth~\citep{Shao2025:CVPR} & 15.6 & 3.8 & 12.6 & 19.0 \\
  \tFoundationGeo~\citep{Liu2026:ECCV-foundationgeo} & 23.9 & \grayed{8.5} & 15.7 & 16.0 \\
  \tFlashDepth~\citep{Chou2025:ICCV} & 22.2 & \grayed{13.5} & 18.5 & 15.5 \\
  \tDVD~\citep{Zhang2026:ICMLW} & 19.9 & 8.0 & 18.2 & 15.0 \\
  \tLapis~\citep{Liu2026:ECCV} & 22.7 & 8.5 & 11.1 & 15.0 \\
  \tDepthCrafter~\citep{Hu2025} & 24.8 & 7.7 & 15.9 & 14.3 \\
  \tVideoDepthAnything~\citep{Chen2025:CVPR} & 23.9 & 9.1 & 18.6 & 12.0 \\
  \tDAGE~\citep{Ngo2026:CVPR} & 32.9 & \grayed{12.7} & 21.5 & 10.5 \\
  \tMoGe~\citep{Wang2025:NeurIPS} & 32.2 & 8.4 & 23.0 & 10.3 \\
  \tUniDepthTwo~\citep{Piccinelli2025:PAMI} & 34.3 & 7.0 & 23.9 & 10.3 \\
  \tUniKThreeD~\citep{Piccinelli2025:CVPR} & 33.5 & 7.9 & 24.6 & 9.3 \\
  \tMarigoldVTwo~\citep{Pavlovic2026:SIGGRAPHAsia} & 27.6 & \cellthird{13.9} & 22.6 & 8.7 \\
  \tLotus~\citep{He2026:arxiv} & 28.5 & 10.7 & 24.4 & 8.3 \\
  \tMoGeThree~\citep{Kong2026:arxiv} & \cellthird{37.1} & 9.5 & 27.5 & 5.3 \\
  \tPixelPerfectDepth~\citep{Xu2025:NeurIPS} & 34.9 & 10.1 & 27.7 & 5.3 \\
  \tDepthPro~\citep{Bochkovskii2025:ICLR} & 36.3 & 12.7 & 24.8 & 4.7 \\
  \tMetricAnything~\citep{Ma2026:ECCV} & 34.4 & \cellsecond{14.0} & \cellthird{29.4} & \cellthird{3.7} \\
  \tInfiniDepth~\citep{Yu2026:CVPR} & \cellfirst{44.8} & 11.8 & \cellsecond{30.5} & \cellsecond{2.7} \\
  \midrule
  \tOursH & \cellsecond{43.4} & \cellfirst{16.8} & \cellfirst{34.4} & \cellfirst{1.3} \\
  \bottomrule
\end{tabular}
}
 \end{subtable}
\end{table}

In addition to the experiments and results reported in this section, the supplement provides model and training details (Appendix~\ref{sec:appendix-method}), further quantitative and qualitative comparisons to prior work (Appendix~\ref{sec:appendix-core-results}), an evaluation of the \ours{} model family (Appendix~\ref{sec:appendix-family-datasets}),
evaluations on downstream applications (Appendix~\ref{sec:appendix-applications}), controlled experiments including an evaluation on long video sequences (Appendix~\ref{sec:appendix-controlled}), and specification of the evaluation protocols (Appendix~\ref{sec:appendix-evaluation-protocols}).

We evaluate \ours in zero-shot settings on metric depth accuracy, boundary sharpness, and temporal consistency, comparing against a comprehensive set of baselines.
When comparing to prior work in Tables~\ref{tab:paper-metric-image}--\ref{tab:paper-temporal}, we use \oursh{}.

Figure~\ref{fig:qualitative-sharpness} provides a qualitative comparison to prior work.
For further qualitative assessment, we encourage the reader to watch the videos at \website{}.

\input{tex/figures/qualgrid}
\example{branches}{pexels-colin-g-667753876-36673310}{0.6200}{0.0700}{0.70}{0.1710}
\example{cow}{pexels-laura-paredis-1047081-13156607}{0.7500}{0.1700}{0.8847}{0.2598}
\example{picasjoe}{pexels-picasjoe-11361666}{0.0500}{0.6300}{0.1847}{0.7194}
\example{neodc}{pexels-neodc-16375786}{0.4600}{0.2000}{0.5947}{0.2898}
\example{spokes}{IMG_3307}{0.5}{0.1399}{0.6}{0.2409}
\example{cables}{pexels-jackson-tee-1056084-14827001}{0.3051}{0.0627}{0.4399}{0.1526}
\example{cholla}{IMG_5740}{0.4927}{0.3145}{0.6274}{0.4155}
\example{peacock}{pexels-navlakha-37596901}{0.2}{0.25}{0.35}{0.4}
\example{balloon}{DIS-TE3_2_Aircraft_2_Balloon_2487168092_641505883f_o}{0.4800}{0.1400}{0.5800}{0.2500}
\example{cart}{DIS-TR_17_Non-motor_Vehicle_9_ShoppingCart_8075259294_f23e243849_o}{0.35}{0.25}{0.5}{0.4}
\example{mower}{DIS-VD_17_Non-motor_Vehicle_7_Mower_16992510572_8a6ff27398_o}{0.55}{0.2}{0.65}{0.3}
\example{fence}{fence}{0.45}{0.7}{0.6}{0.85}
\example{monkey}{m_27bd7abe}{0.48}{0.2}{0.62}{0.38}
\example{bonsai}{DIS-TE4_18_Plant_1_Bonsai_3539420884_ca8973e2c0_o}{0.55}{0.15}{0.65}{0.3}

\methods{
  \method{input.jpg}{Input}
  \method{ours.jpeg}{\ours{}}
  \method{depthpro.jpeg}{\depthpro}
  \method{infinidepth.jpeg}{\tInfiniDepth}
  \method{moge3.jpeg}{\mogethree}
  \method{lotus2.jpeg}{\lotustwo}
}

\qualset{zoom gap=0pt}

\qualset{
  border input only/.style={zoom border=false, zoom border methods={input.jpg}},
  border all zooms/.style={zoom border=true, zoom border methods={}},
  qual accent/.style={zoom gap=0pt, border all zooms, qual lines},
  sharpness style/.style={rows=examples, zoom=below, gap=0pt, qual accent}}

\begin{figure*}[tp]
  \centering
  \qualfigure[rows=methods, zoom=bottom right, inset=0.35, qual accent,
                zoom border=true, zoom border methods={}, box width=1pt, box color=white]{peacock, cart, monkey}
  \caption{\textbf{Qualitative results.} Top to bottom: input images, followed by results from Depth~Pro~2 and representative baselines: \depthpro{}~\citep{Bochkovskii2025:ICLR}, \infinidepth{}~\citep{Yu2026:CVPR}, \mogethree{}~\citep{Kong2026:arxiv}, and \lotustwo{}~\citep{He2026:arxiv}.
  Images from \citet{Pexels} (left), DIS-5K~\citep{Qin2022:ECCV} (middle), and AM-2K~\citep{Li2022:IJCV} (right).
  }
  \label{fig:qualitative-sharpness}
\end{figure*}

\mypara{Zero-shot metric depth.}
We evaluate \ours for zero-shot metric depth accuracy against single-image and multi-image baselines that predict dense \emph{metric} depth.
We report results on iBims~\citep{Koch2018}, Middlebury~\citep{Scharstein2014}, Booster~\citep{ramirez2022open}, HAMMER~\citep{Jung2023:CVPR}, ETH3D~\citep{Schoeps2017:CVPR}, KITTI~\citep{Geiger2012:CVPR}, NuScenes~\citep{Caesar2020:CVPR}, DIODE~\citep{Vasiljevic2019:arxiv}, Sintel~\citep{Butler2012:ECCV}, Spring~\citep{Mehl2023:CVPR}, and DyDToF~\citep{Sun2023:CVPR}.
Ten datasets are adopted from the evaluation sets of Depth Pro~\citep{Bochkovskii2025:ICLR} and \mogethree~\citep{Kong2026:arxiv}. We add \dydtof{}, a synthetic video dataset with dense metric ground truth and sharp boundaries. These datasets have been excluded from the training data for most methods and are therefore ideal for zero-shot benchmarking. Baselines that do not conform to the zero-shot setting are marked in gray. We do not include datasets captured with Kinect sensors due to noisy, biased, and incomplete ground truth.

For fair comparison of single-image and video methods, we evaluate them in dedicated single-image and video settings.
In the single-image setting (Table~\ref{tab:paper-metric-image}), all methods are run and evaluated per frame.
In the video setting (Table~\ref{tab:paper-metric-video}), we use the datasets that provide video frames (\kitti{}, \sintel{}, \spring{}, and \dydtof{}). Every method receives the same clips of 50 frames. Single-image methods process each frame independently.
All of our subsequent zero-shot evaluations follow this split as well.
\eththreed{} is a multi-view stereo benchmark with wide-baseline views of a static scene rather than video, so it appears in the single-image setting only.
We evaluate all models under two conditions: first with ground-truth camera intrinsics and second with predicted camera intrinsics, if the model is capable of estimating focal length.
In both the single-image (Table~\ref{tab:paper-metric-image}) and the video setting (Table~\ref{tab:paper-metric-video}), Depth Pro 2 achieves the lowest average rank both with and without ground-truth camera intrinsics.
Appendix~\ref{sec:appendix-absrel} additionally reports the absolute relative error.

\mypara{Zero-shot boundary sharpness.}
Following \citet{Bochkovskii2025:ICLR}, we evaluate zero-shot boundary sharpness using the $F1$ score in Table~\ref{tab:paper-sharpness}. We report scores on datasets from Table~\ref{tab:paper-metric-image} that have high-quality \emph{dense} depth annotations.
Depth Pro 2 has the lowest average rank across datasets.

\begin{table}[t]
 \centering\ra{1.05}\footnotesize\setlength{\tabcolsep}{4pt}
  \caption{\textbf{Zero-shot temporal consistency along point tracks.}
  We apply one affine alignment per video and pool jitter over unoccluded ground-truth point tracks within each dataset. Appendices~\ref{sec:appendix-jitter} and~\ref{sec:appendix-alignment}.
  Scaled by 100 for readability. Lower is better.}
 \label{tab:paper-temporal}
 \fitwidth{\begin{tabular}{lc@{\hspace{9pt}}c@{\hspace{6pt}}|@{\hspace{6pt}}c}
  \toprule
  Method & \multicolumn{1}{c}{\sintel{}{}} & \multicolumn{1}{c}{\spring{}{}} & \multicolumn{1}{c}{Avg.\ rank$\downarrow$} \\
  \midrule
  \tPixelPerfectDepth~\citep{Xu2025:NeurIPS} & 10.29 & 9.90 & 22.0 \\
  \tDepthPro~\citep{Bochkovskii2025:ICLR} & 11.26 & 8.99 & 21.5 \\
  \tInfiniDepth~\citep{Yu2026:CVPR} & 7.66 & 9.48 & 21.0 \\
  \tFoundationGeo~\citep{Liu2026:ECCV-foundationgeo} & 6.79 & \grayed{2.83} & 20.0 \\
  \tLapis~\citep{Liu2026:ECCV} & 6.80 & 4.13 & 18.5 \\
  \tMoGe~\citep{Wang2025:NeurIPS} & 6.48 & 3.75 & 17.0 \\
  \tUniKThreeD~\citep{Piccinelli2025:CVPR} & 5.84 & 4.97 & 17.0 \\
  \tMarigoldVTwo~\citep{Pavlovic2026:SIGGRAPHAsia} & 7.08 & 3.03 & 16.5 \\
  \tFlashDepth~\citep{Chou2025:ICCV} & 5.33 & \grayed{2.19} & 16.0 \\
  \tLotus~\citep{He2026:arxiv} & 6.41 & 3.06 & 15.0 \\
  \tUniDepthTwo~\citep{Piccinelli2025:PAMI} & 5.31 & 3.53 & 14.5 \\
  \tChronoDepth~\citep{Shao2025:CVPR} & 4.07 & 3.13 & 12.5 \\
  \tMetricAnything~\citep{Ma2026:ECCV} & 5.25 & 2.87 & 11.5 \\
  \tMoGeThree~\citep{Kong2026:arxiv} & 5.13 & 2.91 & 11.5 \\
  \tMapAnything~\citep{Keetha2026} & 3.44 & \grayed{1.12} & 11.0 \\
  \tDAGE~\citep{Ngo2026:CVPR} & 2.64 & \grayed{1.00} & 9.0 \\
  \tDVD~\citep{Zhang2026:ICMLW} & 2.52 & 1.51 & 7.0 \\
  \tVGGT~\citep{Wang2025:CVPR} & 2.61 & 1.02 & 6.5 \\
  \tVeloDepth~\citep{Piccinelli2026:3DV} & 3.40 & \cellfirst{0.72} & 5.5 \\
  \tDepthAnythingThree~\citep{Lin2026:ICLR} & 2.47 & \grayed{1.16} & 5.0 \\
  \tVideoDepthAnything~\citep{Chen2025:CVPR} & 2.55 & \cellthird{0.88} & 5.0 \\
  \tDepthCrafter~\citep{Hu2025} & \cellsecond{2.09} & 1.07 & \cellthird{4.5} \\
  \tPiThree~\citep{Wang2026:ICLR} & \cellthird{2.13} & 1.04 & \cellthird{4.5} \\
  \tVGGTOmega~\citep{Wang2026:CVPR} & 2.36 & 0.99 & \cellsecond{4.0} \\
  \midrule
  \tOursH & \cellfirst{1.39} & \cellsecond{0.86} & \cellfirst{1.5} \\
  \bottomrule
\end{tabular}
}
\end{table}

\mypara{Zero-shot temporal consistency.}
We evaluate the temporal consistency of \ours over video sequences.
Table~\ref{tab:paper-temporal} reports jitter on video datasets for which dense high-quality optical flow and occlusion masks are available.
We refer to the appendices for the definition of the metric (\Sec~\ref{sec:appendix-jitter}), further results using the Temporal Alignment Error (\Tab~\ref{tab:paper-temporal-tae}), and a discussion of its limitations (\Sec~\ref{sec:appendix-tae}).

The results in Tables~\ref{tab:paper-sharpness} and~\ref{tab:paper-temporal} confirm our earlier findings from Figure~\ref{fig:tradeoff}. Multi-frame approaches are temporally consistent but elide boundary detail, while the opposite holds for single-frame approaches. Depth Pro 2 overcomes this trade-off and sets a new state of the art for both boundary sharpness and temporal consistency.

\mypara{Zero-shot focal-length.}
Table~\ref{tab:focal-photo} compares our focal length estimator against \depthpro{}, two dedicated camera-calibration methods~\citep{tiradogarin2025anycalib,veicht2024geocalib}, and three depth models that predict their own camera intrinsics~\citep{Piccinelli2025:PAMI,Piccinelli2025:CVPR,Lin2026:ICLR}.
\ours ranks first on all six datasets, with a mean $\delta_{25}$ of 88.6\% against 71.7\% for \depthpro.

\begin{table}[!ht]
  \centering\ra{1.05}\footnotesize
  \caption{
  {\bf Zero-shot focal length estimation.}
  We evaluate focal length estimation accuracy on 6 datasets~\citep{Bochkovskii2025:ICLR}. $\delta_{25}$ and $\delta_{50}$ are the percentages of images whose relative focal-length error is below $25\%$ and $50\%$; higher is better. Mean and rank use $\delta_{25}$ over all six datasets. Rank is the mean per-dataset rank for $\delta_{25}$, where lower is better.
  }
  \label{tab:focal-photo}
  \fitwidth{\begin{tabular}{lrrrrrrrrrrrr|rr}
  \toprule
  & \multicolumn{2}{c}{DDDP} & \multicolumn{2}{c}{FiveK} & \multicolumn{2}{c}{PPR10K} & \multicolumn{2}{c}{RAISE} & \multicolumn{2}{c}{SPAQ} & \multicolumn{2}{c}{ZOOM} & Mean & Rank \\
  \cmidrule(lr){2-3}\cmidrule(lr){4-5}\cmidrule(lr){6-7}\cmidrule(lr){8-9}\cmidrule(lr){10-11}\cmidrule(lr){12-13}
  Method & $\delta_{25}$ & $\delta_{50}$ & $\delta_{25}$ & $\delta_{50}$ & $\delta_{25}$ & $\delta_{50}$ & $\delta_{25}$ & $\delta_{50}$ & $\delta_{25}$ & $\delta_{50}$ & $\delta_{25}$ & $\delta_{50}$ & $\delta_{25}\uparrow$ & $\downarrow$ \\
  \midrule
  \tGeoCalib~\citep{veicht2024geocalib} & 33.4 & 74.6 & 47.8 & 74.4 & 40.4 & 78.4 & 58.2 & 83.2 & 52.8 & 72.6 & 40.2 & 67.2 & 45.5 & 6.17 \\
  \tUniDepthTwo~\citep{Piccinelli2025:PAMI} & 14.3 & 48.1 & 48.4 & 77.0 & 52.2 & 86.8 & 76.0 & 89.6 & 65.1 & 82.5 & 37.1 & 67.0 & 48.8 & 5.83 \\
  \tDepthAnythingThree~\citep{Lin2026:ICLR} & 27.3 & 66.3 & \cellthird{56.5} & 82.0 & 52.8 & 84.8 & 67.6 & 89.6 & 70.4 & 86.2 & 44.5 & 69.0 & 53.2 & 4.67 \\
  \tAnyCalib~\citep{tiradogarin2025anycalib} & \cellthird{36.5} & \cellthird{86.4} & 54.2 & 78.8 & 30.0 & 72.6 & \cellthird{78.0} & 91.2 & 67.8 & 81.6 & \cellthird{58.6} & \cellthird{77.9} & 54.2 & 4.17 \\
  \tUniKThreeD~\citep{Piccinelli2025:CVPR} & 30.1 & 77.0 & 53.0 & \cellthird{83.0} & \cellsecond{67.8} & \cellsecond{92.2} & 75.8 & \cellthird{93.8} & \cellsecond{75.2} & \cellsecond{87.6} & 45.1 & 77.6 & \cellthird{57.8} & \cellthird{3.83} \\
  \tDepthPro~\citep{Bochkovskii2025:ICLR} & \cellsecond{65.0} & \cellsecond{87.0} & \cellsecond{76.0} & \cellsecond{93.4} & \cellthird{64.6} & \cellthird{87.4} & \cellsecond{83.2} & \cellsecond{97.8} & \cellthird{71.2} & \cellthird{87.0} & \cellsecond{70.4} & \cellsecond{91.1} & \cellsecond{71.7} & \cellsecond{2.33} \\
  \midrule
  \tOursFoV & \cellfirst{95.0} & \cellfirst{99.5} & \cellfirst{89.0} & \cellfirst{97.4} & \cellfirst{80.2} & \cellfirst{97.8} & \cellfirst{97.2} & \cellfirst{99.6} & \cellfirst{86.0} & \cellfirst{94.0} & \cellfirst{84.5} & \cellfirst{96.3} & \cellfirst{88.6} & \cellfirst{1.00} \\
  \bottomrule
\end{tabular}
}
\end{table}

\mypara{Applications.}
To assess how \ours{} compares to baselines on downstream applications, we conduct additional experiments, reported in Appendix~\ref{sec:appendix-applications}. Specifically, we provide quantitative (Table~\ref{tab:bokeh-realbokeh}) and qualitative results (Figure~\ref{fig:bokeh-rendering}) on depth-guided bokeh rendering, and qualitative results on novel view synthesis from a single image (Figure~\ref{fig:nvs-comparison}).

\FloatBarrier\section{Conclusion}
We presented a family of models that estimate sharp and temporally consistent metric depth.
By analyzing prior approaches to monocular depth estimation, we observed a dichotomy between image-based methods that produce sharp depth maps and video-based methods that are temporally stable but elide detail. \ours overcomes this trade-off and provides unprecedented sharpness and temporal stability.

We release a family of models, each of which processes both single images and video sequences at full resolution, and can process video in a streaming fashion. The smallest model, \tOursS, processes streaming 1080p video at 53 fps on a single GPU. The largest model, \tOursH, processes 1080p video at 8 fps and sets the new state of the art on metric depth accuracy, boundary sharpness, and temporal stability all at once.

\clearpage

\bibliography{main}

\clearpage

\appendix
\section*{Supplement}
This supplement is organized as follows.
Appendix~\ref{sec:appendix-method} provides implementation and training details.
Appendix~\ref{sec:appendix-core-results} provides additional qualitative and quantitative results.
Appendix~\ref{sec:appendix-family-datasets} provides a quantitative comparison between different models in the \ours family.
Appendix~\ref{sec:appendix-applications} evaluates predicted depth in two downstream applications: bokeh rendering and novel-view synthesis.
Appendix~\ref{sec:appendix-controlled} presents controlled experiments and examines design choices and parameter settings.
Appendix~\ref{sec:appendix-evaluation-protocols} specifies the evaluation protocol used throughout the paper, including the held-out validation sets and how we measure runtime.

\section{Implementation details}
\label{sec:appendix-method}

\subsection{Training}
\label{sec:appendix-training-impl}

\mypara{Training optimization.}
We train all our models for 100K steps on 128 B200 GPUs. We use AdamW at a peak learning rate of $5 \times 10^{-5}$, cosine annealed after a $5\%$ warmup.

Section~\ref{sec:training} describes our native-resolution training strategy. Here, we provide additional implementation details on token-budget batching and the composition of the training data.

\mypara{Token-budget batching.}
Each GPU has a maximum budget of 73,728 patch tokens during training. Half of the steps use the full native frame. Otherwise, we draw an aspect ratio relative to the native aspect and a short-side limit, then crop to 60--100\% of the largest feasible window. Where native resolution permits, we choose spatial stride 2 with probability 0.25, taking every second pixel in both the image and depth. We do not interpolate ground truth to keep fidelity. Each clip uses as many frames as fit in the token budget, up to a separate cap of 32 frames. Remaining capacity is filled with more clips from the same source. The short side of the input video is 608/1024/2048 pixels at 10th/50th/90th percentiles. The sampled aspect ratios are 1.00/1.07/1.50 at the same percentiles.

\mypara{Loss application and normalization.}
We apply $\mathcal{L}_{\mathit{ssi}}$ to all data sources. For metric synthetic and real sensor sources, we apply $\mathcal{L}_{\mathit{mae}}$. We in addition apply $\mathcal{L}_{\mathit{mge}}$ on dense synthetic depth,
and $\mathcal{L}_{\mathit{mvc}}$ on multi-frame sources with camera poses.

For $\mathcal{L}_{\mathit{ssi}}$ and $\mathcal{L}_{\mathit{mae}}$, we first average over valid pixels within each frame, then average over frames per GPU. For $\mathcal{L}_{\mathit{mge}}$, we pool valid pixels across frames separately at each pyramid level and weight the levels equally. For $\mathcal{L}_{\mathit{mvc}}$, we average over sampled valid pixel pairs from frame pairs. Gradients are averaged equally across GPUs.

\mypara{Training datasets.}
Table~\ref{tab:training-data} lists each dataset used for training, whether it contains real or synthetic images, the number of scenes, total frames, and the maximum resolution of frames within the dataset.
In addition to publicly available datasets, we include internal real-world  and synthetic data.
\emph{Internal (Real)} consists of handheld smartphone videos with time-of-flight depth labels and camera poses.
\emph{Internal (Synthetic)} features humans, pets, and objects in rendered indoor and outdoor environments under varied lighting in static and dynamic scenes, rendered from multiple viewpoints and with moving cameras. Images are supplemented with dense metric depth, intrinsics, and extrinsics.

\begin{table}[h]
    \centering\footnotesize
    \caption{\textbf{Training data mixture.} Please refer to the text for a description of our internal datasets.}
    \label{tab:training-data}
    \fitwidth{%
    \begin{tabular}{lcrrw{r}{4em}@{$\times$}w{l}{4em}}
        \toprule
        Dataset & Real/Synthetic & \# Scenes & \# Frames & \multicolumn{2}{c}{Maximum resolution} \\
        \midrule
         Internal (Synthetic) & Synthetic & 964,332 & 24,150,620 & 4096&3072 \\
         Internal (Real) & Real & 2,001 & 3,046,117 & 1920&1440 \\
         \bedlam{}~\citep{Black2023} & Synthetic & 9,454 & 1,452,165 & 1280&720 \\
         \dynamicreplica{}~\citep{Karaev2023:CVPR} & Synthetic & 477 & 155,100 & 1280&720 \\
         \eden{}~\citep{Le2021:WACV} & Synthetic & 694 & 348,062 & 640&480 \\
         \hypersim{}~\citep{Roberts2021:ICCV} & Synthetic & 695 & 68,324 & 1024&768 \\
         IRS~\citep{Wang2021:ICME} & Synthetic & 92,976 & 92,976 & 960&540 \\
         \matrixcity{}~\citep{Li2023:ICCV} & Synthetic & 1,934 & 279,097 & 1000&1000 \\
         \megasynth{}~\citep{Jiang2025} & Synthetic & 64,200 & 3,081,600 & 512&512 \\
         \pointodyssey{}~\citep{Zheng2023} & Synthetic & 938 & 93,798 & 960&540 \\
         \sceneflow{} (FlyingThings3D)~\citep{Mayer2016:CVPR} & Synthetic & 407 & 4,067 & 960&540 \\
         \sceneflow{} (Monkaa)~\citep{Mayer2016:CVPR} & Synthetic & 82 & 8,182 & 960&540 \\
         \sceneflow{} (Driving)~\citep{Mayer2016:CVPR} & Synthetic & 44 & 4,400 & 960&540 \\
         \tartanair{} V2~\citep{Wang2020:IROS} & Synthetic & 1,046 & 10,551,809 & 640&640 \\
         \tartanair{}~\citep{Wang2020:IROS} & Synthetic & 2,727 & 272,697 & 640&480 \\
         TartanGround~\citep{Patel2025:IROS} & Synthetic & 816 & 10,604,597 & 640&640 \\
         \urbansyn{}~\citep{Gmez2025} & Synthetic & 7,515 & 7,515 & 2048&1024 \\
         Virtual KITTI 2~\citep{Cabon2020:arxiv} & Synthetic & 96 & 41,164 & 1242&375 \\
         Waymo Open Dataset~\citep{Sun2020:CVPR} & Real & 958 & 922,894 & 1920&1280 \\
        \bottomrule
    \end{tabular}%
    }
\end{table}

\subsection{Field-of-view estimation}
\label{sec:appendix-focal-setup}

\mypara{Architecture.}
We initialize our field-of-view estimator with a pretrained DINOv3 ViT-S/16, and finetune end-to-end on ImageNet-normalized inputs.
A small attention head (1.9M parameters, two transformer blocks of width 248) reads the tokens of blocks 2, 5, 8, and 11, and predicts the horizontal and vertical fields of view, each bounded to $[1^\circ, 179^\circ]$.

\mypara{Training.}
We train on a large corpus of $\sim$92M images from Flickr~\citep{Thomee2016:CACM}, discarding those with invalid or missing EXIF data.

\mypara{Augmentation and input size.}
We resize every image to 512 pixels along its shorter side, retaining original aspect ratios.
Augmentations comprise horizontal flips ($p=0.5$) and standard photometric changes (color removal or jitter, blur, noise, JPEG/WebP recompression).
Since we train at native aspect ratio, image size may vary between batches.
Each GPU therefore fills its batch up to a budget of $147{,}456$ tokens ($16\times 16$-pixel patches; e.g., 144 photos at $512 \times 512$ or 96 at $512\times 768$).

\mypara{Objective and optimization.}
Our objective is an $L_1$ loss on both angles in units of degrees.
Optimization uses AdamW with a peak learning rate of $10^{-4}$, weight decay $0.01$, and a cosine schedule with a 5\% warmup, in bf16 mixed precision. Each step packs about $3{,}050$ photos on 32 B200 GPUs; training runs for 120k steps.

\mypara{Inference.}
For inference we resize images to have a shorter side of 512 pixels. Smaller images keep their size, with the shorter side rounded down to a multiple of 32.
Frames wider than 16:9 are center-cropped to 16:9 along the long side.
This crop changes neither $f_\px$ nor the short-side field of view.
Only the short-side field of view is kept and converted to $f_\px$, assuming square pixels.

\section{Additional results}
\label{sec:appendix-core-results}

\subsection{Metric depth under absolute relative error}
\label{sec:appendix-absrel}

Tables~\ref{tab:paper-metric-image-absrel} and~\ref{tab:paper-metric-video-absrel} repeat the metric depth comparisons of Tables~\ref{tab:paper-metric-image} and~\ref{tab:paper-metric-video} under the absolute relative error (AbsRel).

\begin{table}[htb]
  \centering\ra{1.05}\footnotesize\setlength{\tabcolsep}{1pt}
  \caption{\textbf{Zero-shot metric depth error, single image.} We report the absolute relative error (AbsRel, lower is better) of the predictions in Table~\ref{tab:paper-metric-image}, under ground-truth and predicted intrinsics. Rows keep the order of Table~\ref{tab:paper-metric-image}. Ranks and \textcolor{gray}{gray} non-zero-shot results follow its conventions. Unlike $\delta_1$, which is the fraction of pixels within a factor of 1.25 of the ground truth, AbsRel averages the size of the relative error, so a few large errors weigh heavily. Most of our \diode{} error comes from a few outdoor pixels around bare branches, which the laser ground truth marks as valid at the depth of the branch although the image shows sky. We predict sky there, which the evaluation places at its maximum depth. AbsRel has no bound on such overestimates, while $\delta_1$ counts each of these pixels as one miss. $^{\ddagger}$MetricAnything uses GT intrinsics only.}
  \label{tab:paper-metric-image-absrel}
  \fitwidth{\begin{tabular}{lcc@{\hspace{5pt}}cc@{\hspace{5pt}}cc@{\hspace{5pt}}cc@{\hspace{5pt}}cc@{\hspace{5pt}}cc@{\hspace{5pt}}cc@{\hspace{5pt}}cc@{\hspace{5pt}}cc@{\hspace{5pt}}cc@{\hspace{5pt}}cc@{\hspace{6pt}}|@{\hspace{6pt}}cc}
  \toprule
  & \multicolumn{8}{c}{Real, indoor} & \multicolumn{4}{c}{Real, indoor+outdoor} & \multicolumn{4}{c}{Real, outdoor} & \multicolumn{6}{c}{Synthetic} & \\
  \cmidrule(lr){2-9}\cmidrule(lr){10-13}\cmidrule(lr){14-17}\cmidrule(lr){18-23}
  Method & \multicolumn{2}{c}{\hdribims} & \multicolumn{2}{c}{\hdrmiddleburyf} & \multicolumn{2}{c}{\hdrbooster} & \multicolumn{2}{c}{\hdrhammer} & \multicolumn{2}{c}{\eththreed} & \multicolumn{2}{c}{\hdrdiode} & \multicolumn{2}{c}{\kitti} & \multicolumn{2}{c}{\hdrnuscenes} & \multicolumn{2}{c}{\sintel} & \multicolumn{2}{c}{\spring} & \multicolumn{2}{c}{\dydtof} & \multicolumn{2}{c}{\hdravgrank} \\
  \cmidrule(lr){2-3}\cmidrule(lr){4-5}\cmidrule(lr){6-7}\cmidrule(lr){8-9}\cmidrule(lr){10-11}\cmidrule(lr){12-13}\cmidrule(lr){14-15}\cmidrule(lr){16-17}\cmidrule(lr){18-19}\cmidrule(lr){20-21}\cmidrule(lr){22-23}\cmidrule(l){24-25}
   & GT & pred & GT & pred & GT & pred & GT & pred & GT & pred & GT & pred & GT & pred & GT & pred & GT & pred & GT & pred & GT & pred & GT & pred \\
  \midrule
  \tMapAnything~\citep{Keetha2026} & 0.157 & 0.200 & 0.328 & 0.305 & 0.541 & 0.417 & 0.838 & 0.652 & 0.233 & 0.250 & 0.431 & 0.460 & 0.107 & 0.102 & 0.211 & 0.280 & 1.313 & 1.220 & \grayed{0.073} & \grayed{0.072} & 0.473 & 0.482 & 9.7 & 8.6 \\
  \tDepthPro~\citep{Bochkovskii2025:ICLR} & 0.120 & 0.165 & \cellfirst{0.197} & \cellsecond{0.167} & 0.281 & 0.289 & 0.275 & 0.489 & 0.291 & 0.320 & 0.506 & 0.521 & 0.124 & 0.142 & 0.215 & 0.206 & \cellsecond{0.539} & 0.704 & \cellsecond{0.543} & 0.610 & 0.208 & 0.379 & 6.5 & 7.0 \\
  \tMetricAnything$^{\ddagger}$~\citep{Ma2026:ECCV} & 0.133 & -- & 0.314 & -- & 0.262 & -- & \cellthird{0.220} & -- & 0.165 & -- & 0.544 & -- & 0.333 & -- & 0.163 & -- & 0.853 & -- & 0.843 & -- & 0.168 & -- & 7.5 & -- \\
  \tFoundationGeo~\citep{Liu2026:ECCV-foundationgeo} & 0.095 & 0.161 & 0.284 & 0.408 & 0.322 & 0.206 & 0.285 & \cellthird{0.178} & 0.139 & 0.149 & \cellfirst{0.324} & \cellfirst{0.305} & \cellthird{0.075} & \cellsecond{0.071} & 0.135 & 0.170 & 1.767 & 0.788 & \grayed{0.489} & \grayed{0.504} & 0.279 & 0.493 & 6.4 & 6.0 \\
  \tMoGeThree~\citep{Kong2026:arxiv} & 0.096 & 0.151 & 0.293 & 0.269 & 0.366 & 0.284 & 0.371 & 0.297 & \cellthird{0.116} & 0.137 & 0.402 & 0.360 & 0.102 & 0.193 & 0.119 & \cellthird{0.116} & 1.556 & 0.712 & 1.234 & \cellsecond{0.481} & 0.159 & 0.379 & 6.5 & 5.7 \\
  \tMoGe~\citep{Wang2025:NeurIPS} & 0.108 & 0.138 & 0.321 & 0.244 & 0.315 & 0.242 & 0.291 & 0.270 & \cellsecond{0.109} & \cellfirst{0.101} & \cellthird{0.360} & \cellthird{0.335} & 0.121 & 0.210 & 0.135 & 0.126 & 0.811 & \cellfirst{0.448} & 1.019 & \cellthird{0.564} & 0.198 & 0.421 & 6.3 & \cellthird{4.7} \\
  \tDepthAnythingThree~\citep{Lin2026:ICLR} & \cellfirst{0.075} & \cellthird{0.102} & 0.230 & 0.226 & \cellthird{0.207} & \cellthird{0.201} & \cellsecond{0.208} & \cellsecond{0.155} & 0.128 & 0.142 & 0.426 & 0.424 & 0.145 & 0.217 & 0.152 & 0.253 & 0.613 & \cellsecond{0.543} & \grayed{0.534} & \grayed{0.565} & 0.256 & 0.425 & 5.2 & 5.3 \\
  \tDAGE~\citep{Ngo2026:CVPR} & 0.102 & 0.116 & 0.230 & \cellsecond{0.167} & 0.261 & 0.263 & 0.309 & 0.296 & 0.117 & \cellthird{0.134} & \cellsecond{0.335} & \cellsecond{0.308} & 0.080 & 0.118 & 0.172 & 0.125 & 0.947 & 1.069 & \grayed{0.116} & \grayed{0.101} & \cellthird{0.156} & 0.264 & 5.6 & 4.9 \\
  \tUniDepthTwo~\citep{Piccinelli2025:PAMI} & \cellsecond{0.076} & \cellsecond{0.085} & \cellthird{0.211} & 0.227 & \cellfirst{0.185} & \cellsecond{0.200} & 0.375 & 0.430 & 0.146 & 0.189 & 0.801 & 0.783 & 0.076 & \cellfirst{0.067} & \cellsecond{0.086} & \cellsecond{0.110} & 0.687 & 0.645 & \cellthird{0.663} & 0.616 & \cellsecond{0.147} & \cellthird{0.253} & \cellthird{4.5} & \cellthird{4.7} \\
  \tUniKThreeD~\citep{Piccinelli2025:CVPR} & \cellthird{0.077} & \cellfirst{0.084} & 0.216 & \cellthird{0.211} & \cellsecond{0.195} & \cellfirst{0.194} & 0.318 & 0.293 & 0.118 & 0.142 & 0.497 & 0.485 & \cellfirst{0.057} & \cellthird{0.091} & \cellfirst{0.083} & \cellfirst{0.108} & \cellthird{0.578} & \cellthird{0.550} & 0.678 & 0.599 & 0.181 & \cellsecond{0.202} & \cellsecond{4.0} & \cellfirst{3.3} \\
  \midrule
  \tOursH & 0.086 & 0.110 & \cellsecond{0.210} & \cellfirst{0.163} & 0.230 & 0.675 & \cellfirst{0.127} & \cellfirst{0.147} & \cellfirst{0.093} & \cellsecond{0.116} & 2.418 & 2.403 & \cellsecond{0.060} & 0.106 & \cellthird{0.115} & 0.124 & \cellfirst{0.338} & 0.570 & \cellfirst{0.397} & \cellfirst{0.442} & \cellfirst{0.136} & \cellfirst{0.152} & \cellfirst{2.8} & \cellsecond{3.9} \\
  \bottomrule
\end{tabular}
}
\end{table}
\begin{table}[htb]
  \centering\ra{1.05}\footnotesize\setlength{\tabcolsep}{5pt}
  \caption{\textbf{Zero-shot metric depth error, video.} All methods receive clips up to 50 frames, as in Table~\ref{tab:paper-metric-video}. We report AbsRel and average rank, as in Table~\ref{tab:paper-metric-image-absrel}. $^{\ddagger}$MetricAnything uses GT intrinsics only.}
  \label{tab:paper-metric-video-absrel}
  \fitwidth{\begin{tabular}{lcc@{\hspace{5pt}}cc@{\hspace{5pt}}cc@{\hspace{5pt}}cc@{\hspace{6pt}}|@{\hspace{6pt}}cc}
  \toprule
  & \multicolumn{2}{c}{Real, outdoor} & \multicolumn{6}{c}{Synthetic} & \\
  \cmidrule(lr){2-3}\cmidrule(lr){4-9}
  Method & \multicolumn{2}{c}{\kitti} & \multicolumn{2}{c}{\sintel} & \multicolumn{2}{c}{\spring} & \multicolumn{2}{c}{\dydtof} & \multicolumn{2}{c}{\hdravgrank} \\
  \cmidrule(lr){2-3}\cmidrule(lr){4-5}\cmidrule(lr){6-7}\cmidrule(lr){8-9}\cmidrule(l){10-11}
   & GT & pred & GT & pred & GT & pred & GT & pred & GT & pred \\
  \midrule
  \tMapAnything~\citep{Keetha2026} & 0.091 & 0.087 & 1.176 & 1.114 & \grayed{0.065} & \grayed{0.065} & 0.448 & 0.453 & 9.3 & 8.3 \\
  \tFoundationGeo~\citep{Liu2026:ECCV-foundationgeo} & \cellsecond{0.075} & \cellsecond{0.071} & 1.767 & 0.788 & \grayed{0.489} & \grayed{0.504} & 0.279 & 0.493 & 8.3 & 7.3 \\
  \tDepthAnythingThree~\citep{Lin2026:ICLR} & 0.098 & 0.251 & \cellsecond{0.525} & \cellsecond{0.463} & \grayed{0.531} & \grayed{0.552} & 0.179 & 0.410 & 5.0 & 7.0 \\
  \tVeloDepth~\citep{Piccinelli2026:3DV} & 0.138 & 0.203 & 0.620 & 0.602 & 0.674 & 0.587 & 0.226 & 0.241 & 7.5 & 5.2 \\
  \tDepthPro~\citep{Bochkovskii2025:ICLR} & 0.124 & 0.142 & \cellthird{0.539} & 0.704 & \cellsecond{0.543} & 0.610 & 0.208 & 0.379 & 6.0 & 6.5 \\
  \tMetricAnything$^{\ddagger}$~\citep{Ma2026:ECCV} & 0.333 & -- & 0.853 & -- & 0.843 & -- & 0.168 & -- & 8.0 & -- \\
  \tMoGeThree~\citep{Kong2026:arxiv} & 0.102 & 0.193 & 1.556 & 0.712 & 1.234 & \cellsecond{0.481} & \cellthird{0.159} & 0.379 & 7.8 & 6.2 \\
  \tMoGe~\citep{Wang2025:NeurIPS} & 0.121 & 0.210 & 0.811 & \cellfirst{0.448} & 1.019 & \cellthird{0.564} & 0.198 & 0.421 & 7.8 & 5.8 \\
  \tDAGE~\citep{Ngo2026:CVPR} & 0.083 & 0.132 & 0.820 & 0.908 & \grayed{0.092} & \grayed{0.085} & \cellfirst{0.111} & \cellsecond{0.196} & 4.7 & 6.0 \\
  \tUniKThreeD~\citep{Piccinelli2025:CVPR} & \cellfirst{0.057} & 0.091 & 0.578 & \cellthird{0.550} & 0.678 & 0.599 & 0.181 & \cellthird{0.202} & \cellthird{4.2} & \cellsecond{4.0} \\
  \tUniDepthTwo~\citep{Piccinelli2025:PAMI} & \cellthird{0.076} & \cellfirst{0.067} & 0.687 & 0.645 & \cellthird{0.663} & 0.616 & \cellsecond{0.147} & 0.253 & \cellsecond{3.8} & \cellthird{4.8} \\
  \midrule
  \tOursH & 0.090 & \cellthird{0.077} & \cellfirst{0.341} & 0.620 & \cellfirst{0.259} & \cellfirst{0.405} & \cellfirst{0.111} & \cellfirst{0.125} & \cellfirst{2.0} & \cellfirst{2.5} \\
  \bottomrule
\end{tabular}
}
\end{table}

\subsection{Relative-depth accuracy}
\label{sec:appendix-relative}

In addition to the comparisons on metric depth, we compare \oursh{} to prior work on zero-shot relative depth estimation. Predictions are aligned to the ground truth per frame (see Section~\ref{sec:appendix-alignment}).
Tables~\ref{tab:paper-relative-single} and~\ref{tab:paper-relative-video} report $\delta_1$ from single images and videos, respectively. \oursh obtains the best average rank when evaluated on single images, and the second-best average rank when evaluated on videos.

\begin{table}[h]
  \centering\ra{1.05}\footnotesize\setlength{\tabcolsep}{2pt}
  \caption{\textbf{Zero-shot relative depth accuracy, single image.} We report $\delta_1$ score (higher is better) under per-frame affine alignment, applied to every method including metric-depth methods and ours, as described in Section~\ref{sec:appendix-alignment}. We evaluate on 14 datasets, including datasets that appeared separately in Tables~\ref{tab:paper-metric-image}, \ref{tab:paper-sharpness}, and~\ref{tab:paper-temporal-tae}.
  Every method receives one frame at a time, including video methods.
  We exclude three video methods: \depthcrafter{}~\citep{Hu2025} and \vda{} cannot run on a single frame, and \velodepth{}~\citep{Piccinelli2026:3DV} reduces to \unikthreed{}~\citep{Piccinelli2025:CVPR}.}
  \label{tab:paper-relative-single}
  \fitwidth{\begin{tabular}{lcccccccccccccc@{\hspace{6pt}}|@{\hspace{6pt}}c}
  \toprule
  & \multicolumn{7}{c}{Real, indoor} & \multicolumn{2}{c}{Real, indoor+outdoor} & \multicolumn{2}{c}{Real, outdoor} & \multicolumn{3}{c}{Synthetic} & \\
  \cmidrule(lr){2-8}\cmidrule(lr){9-10}\cmidrule(lr){11-12}\cmidrule(lr){13-15}
  Method & \hdribims & \hdrmiddleburyf & \hdrbooster & \hdrhammer & \scannet & \bonn & \hdrtumdynamics & \eththreed & \diode & \kitti & \hdrnuscenes & \sintel & \spring & \dydtof & Avg.\ rank$\downarrow$ \\
  \midrule
  \tChronoDepth~\citep{Shao2025:CVPR} & 83.0 & 89.6 & 86.1 & 82.8 & 87.4 & 83.6 & 77.8 & 86.4 & 69.2 & 77.4 & 68.4 & 66.0 & 58.0 & 83.3 & 21.0 \\
  \tDVD~\citep{Zhang2026:ICMLW} & 96.8 & 99.4 & 98.4 & 96.8 & 95.9 & 96.9 & 87.1 & 97.7 & 79.7 & 92.7 & 84.1 & 74.7 & 77.0 & 93.3 & 17.9 \\
  \tFlashDepth~\citep{Chou2025:ICCV} & 97.7 & 99.2 & 96.9 & 97.1 & 96.3 & 97.7 & 93.1 & 96.4 & 79.2 & 93.7 & 83.5 & 78.2 & \grayed{89.3} & 93.9 & 16.7 \\
  \tMapAnything~\citep{Keetha2026} & 96.9 & 98.7 & 97.9 & 97.8 & 97.7 & 97.5 & 90.6 & 98.3 & 79.9 & 95.5 & 88.1 & 75.3 & \grayed{97.0} & 85.6 & 16.3 \\
  \tLapis~\citep{Liu2026:ECCV} & 96.3 & 99.0 & 98.4 & 98.7 & 96.2 & 97.2 & 84.0 & 97.8 & 81.3 & 96.4 & 84.8 & 79.0 & 82.6 & 82.1 & 16.2 \\
  \tVGGT~\citep{Wang2025:CVPR} & 97.1 & 98.8 & 97.3 & 98.0 & \grayed{98.6} & 97.8 & 91.3 & 98.2 & 80.9 & 96.1 & 86.4 & 76.8 & \cellsecond{90.6} & 91.5 & 15.5 \\
  \tDepthAnythingThree~\citep{Lin2026:ICLR} & 97.5 & 99.0 & 99.1 & 98.8 & 96.5 & 97.7 & 90.8 & 98.3 & 80.4 & 95.2 & 87.0 & 76.7 & \grayed{83.5} & 90.3 & 15.1 \\
  \tVGGTOmega~\citep{Wang2026:CVPR} & 96.9 & 98.7 & 97.5 & 98.6 & \grayed{98.0} & 91.3 & 93.3 & 98.9 & 81.2 & 96.3 & 87.3 & 80.3 & \cellfirst{94.9} & 95.1 & 13.5 \\
  \tPixelPerfectDepth~\citep{Xu2025:NeurIPS} & 97.9 & 99.4 & 99.0 & \cellsecond{99.4} & 97.6 & 95.6 & 89.1 & 98.3 & \cellsecond{84.6} & 96.2 & 87.6 & 80.4 & 85.6 & 93.9 & 10.9 \\
  \tDepthPro~\citep{Bochkovskii2025:ICLR} & 98.0 & 99.6 & \cellsecond{99.5} & 99.2 & \grayed{98.2} & 97.9 & 92.1 & 97.7 & 81.9 & 97.4 & 87.1 & 83.2 & 75.8 & 92.8 & 10.8 \\
  \tPiThree~\citep{Wang2026:ICLR} & 97.7 & 99.1 & 97.9 & 99.1 & \grayed{98.8} & \cellfirst{98.6} & 93.3 & 98.8 & 81.2 & 96.9 & 88.7 & 80.6 & 88.0 & 93.2 & 10.8 \\
  \tLotus~\citep{He2026:arxiv} & 98.1 & \cellthird{99.7} & 98.9 & \cellthird{99.3} & 97.8 & 98.0 & 92.0 & 98.4 & 80.6 & 95.1 & 86.2 & 79.8 & 82.0 & 96.0 & 10.8 \\
  \tInfiniDepth~\citep{Yu2026:CVPR} & 98.1 & \cellthird{99.7} & 98.4 & 98.0 & 97.6 & 97.9 & \cellthird{93.8} & 98.8 & 80.9 & 96.9 & 87.6 & 82.2 & 77.2 & \cellthird{96.7} & 9.9 \\
  \tDAGE~\citep{Ngo2026:CVPR} & 98.2 & 99.5 & 98.5 & 98.5 & \grayed{98.8} & 98.0 & 92.5 & 99.1 & 82.3 & 97.8 & 89.2 & 80.1 & \grayed{97.4} & 93.0 & 9.8 \\
  \tMetricAnything~\citep{Ma2026:ECCV} & \cellsecond{98.5} & \cellthird{99.7} & 98.9 & 99.1 & \grayed{99.2} & 97.3 & 93.7 & 98.7 & 81.9 & 97.3 & 89.8 & 78.1 & 82.4 & 95.5 & 9.6 \\
  \tMarigoldVTwo~\citep{Pavlovic2026:SIGGRAPHAsia} & \cellfirst{98.6} & 99.3 & 94.3 & \cellfirst{99.5} & 98.3 & 97.8 & 93.7 & 98.4 & \cellfirst{85.3} & 97.6 & 85.5 & 83.3 & 84.2 & \cellsecond{96.9} & 8.1 \\
  \tMoGe~\citep{Wang2025:NeurIPS} & \cellthird{98.3} & \cellthird{99.7} & \cellsecond{99.5} & 99.1 & \cellthird{98.5} & 97.9 & 86.5 & 99.1 & 82.0 & 98.0 & 89.4 & 81.4 & 85.7 & 95.2 & 7.3 \\
  \tUniDepthTwo~\citep{Piccinelli2025:PAMI} & 98.2 & \cellthird{99.7} & 98.9 & \cellsecond{99.4} & \grayed{98.5} & 97.8 & 93.5 & \cellthird{99.4} & 81.6 & 97.9 & \cellsecond{92.5} & \cellthird{83.8} & 82.7 & 96.1 & 6.4 \\
  \tFoundationGeo~\citep{Liu2026:ECCV-foundationgeo} & 98.1 & 99.6 & \cellthird{99.3} & 99.2 & \cellsecond{98.6} & \cellsecond{98.5} & \cellfirst{96.1} & 99.1 & 82.8 & \cellsecond{98.3} & 89.2 & 82.6 & \grayed{96.7} & 94.0 & 5.7 \\
  \tUniKThreeD~\citep{Piccinelli2025:CVPR} & \cellthird{98.3} & \cellthird{99.7} & 99.0 & \cellsecond{99.4} & \grayed{98.5} & 97.8 & \cellsecond{93.9} & 99.3 & 81.9 & \cellthird{98.2} & \cellfirst{92.7} & 83.6 & 84.2 & 96.4 & \cellthird{5.5} \\
  \tMoGeThree~\citep{Kong2026:arxiv} & \cellfirst{98.6} & \cellsecond{99.8} & 99.0 & \cellfirst{99.5} & 97.4 & \cellthird{98.1} & 88.5 & \cellsecond{99.5} & 82.2 & \cellthird{98.2} & \cellthird{90.2} & \cellsecond{85.4} & \cellthird{90.2} & \cellfirst{97.4} & \cellsecond{4.5} \\
  \midrule
  \tOursH & \cellfirst{98.6} & \cellfirst{99.9} & \cellfirst{99.7} & 99.2 & \cellfirst{98.8} & 98.0 & 93.5 & \cellfirst{99.6} & \cellthird{83.0} & \cellfirst{98.4} & 90.1 & \cellfirst{86.0} & 89.8 & 95.2 & \cellfirst{3.4} \\
  \bottomrule
\end{tabular}
}
\end{table}
\begin{table}[htb]
  \centering\ra{1.05}\footnotesize\setlength{\tabcolsep}{2pt}
  \caption{\textbf{Zero-shot relative depth accuracy, video.} The same $\delta_1$ score on video datasets. Every method receives the same input clips of up to 50 frames and processes them with its own inference procedure, as described in Section~\ref{sec:appendix-baselines}. Video methods can report different numbers here compared to Table~\ref{tab:paper-relative-single} due to different input setup.}
  \label{tab:paper-relative-video}
  \fitwidth{\begin{tabular}{lccccccc@{\hspace{6pt}}|@{\hspace{6pt}}c}
  \toprule
  & \multicolumn{3}{c}{Real, indoor} & \multicolumn{1}{c}{Real, outdoor} & \multicolumn{3}{c}{Synthetic} & \\
  \cmidrule(lr){2-4}\cmidrule(lr){5-5}\cmidrule(lr){6-8}
  Method & \scannet & \bonn & \hdrtumdynamics & \kitti & \sintel & \spring & \dydtof & Avg.\ rank$\downarrow$ \\
  \midrule
  \tChronoDepth~\citep{Shao2025:CVPR} & 87.9 & 83.3 & 74.8 & 73.6 & 67.5 & 56.1 & 80.3 & 23.0 \\
  \tVeloDepth~\citep{Piccinelli2026:3DV} & \grayed{95.3} & 94.2 & 93.6 & 91.9 & 77.6 & 81.4 & 90.8 & 19.7 \\
  \tLapis~\citep{Liu2026:ECCV} & 96.2 & 97.2 & 84.0 & 96.4 & 79.0 & 82.6 & 82.1 & 19.0 \\
  \tFlashDepth~\citep{Chou2025:ICCV} & 96.3 & 97.7 & 93.1 & 93.7 & 78.2 & \grayed{89.3} & 93.9 & 17.3 \\
  \tMapAnything~\citep{Keetha2026} & 98.4 & 97.6 & 91.9 & 97.0 & 77.7 & \grayed{97.2} & 88.2 & 17.2 \\
  \tDepthCrafter~\citep{Hu2025} & 90.3 & \cellthird{98.3} & 84.3 & 91.1 & 79.6 & 82.7 & 92.3 & 16.1 \\
  \tPixelPerfectDepth~\citep{Xu2025:NeurIPS} & 97.5 & 95.1 & 89.1 & 96.2 & 80.4 & 85.6 & 93.9 & 15.7 \\
  \tMetricAnything~\citep{Ma2026:ECCV} & \grayed{99.2} & 97.3 & 93.7 & 97.3 & 78.1 & 82.4 & 95.5 & 15.2 \\
  \tDVD~\citep{Zhang2026:ICMLW} & 98.1 & 98.0 & 88.1 & 97.3 & 78.1 & 79.7 & 95.5 & 14.7 \\
  \tVideoDepthAnything~\citep{Chen2025:CVPR} & 98.0 & 97.7 & 93.9 & 95.0 & 79.4 & 83.9 & 91.6 & 14.4 \\
  \tDepthPro~\citep{Bochkovskii2025:ICLR} & \grayed{98.2} & 97.9 & 92.1 & 97.4 & 83.2 & 75.8 & 92.8 & 13.8 \\
  \tLotus~\citep{He2026:arxiv} & 97.8 & 98.0 & 92.0 & 95.1 & 79.8 & 82.0 & 96.0 & 13.6 \\
  \tVGGT~\citep{Wang2025:CVPR} & \grayed{98.7} & 97.8 & 91.9 & 97.3 & 80.3 & \cellsecond{91.0} & 97.0 & 11.8 \\
  \tInfiniDepth~\citep{Yu2026:CVPR} & 97.6 & 97.9 & 93.8 & 96.9 & 82.2 & 77.2 & 96.7 & 11.7 \\
  \tMoGe~\citep{Wang2025:NeurIPS} & \cellthird{98.5} & 97.9 & 86.5 & 98.0 & 81.4 & 85.7 & 95.2 & 11.3 \\
  \tUniDepthTwo~\citep{Piccinelli2025:PAMI} & \grayed{98.5} & 97.8 & 93.5 & 97.9 & 83.8 & 82.7 & 96.1 & 10.0 \\
  \tMarigoldVTwo~\citep{Pavlovic2026:SIGGRAPHAsia} & 98.3 & 97.8 & 93.7 & 97.6 & 83.3 & 84.2 & 96.9 & 8.4 \\
  \tUniKThreeD~\citep{Piccinelli2025:CVPR} & \grayed{98.5} & 97.8 & 93.9 & 98.2 & 83.6 & 84.2 & 96.4 & 8.3 \\
  \tDepthAnythingThree~\citep{Lin2026:ICLR} & 98.3 & 98.1 & 93.4 & 97.6 & 82.3 & \grayed{87.9} & 97.1 & 8.2 \\
  \tMoGeThree~\citep{Kong2026:arxiv} & 97.4 & 98.1 & 88.5 & 98.2 & \cellsecond{85.4} & 90.2 & 97.4 & 8.1 \\
  \tVGGTOmega~\citep{Wang2026:CVPR} & \grayed{98.7} & 91.7 & 94.3 & 98.2 & 83.7 & \cellfirst{96.5} & \cellsecond{98.1} & 6.8 \\
  \tDAGE~\citep{Ngo2026:CVPR} & \grayed{99.3} & \cellsecond{98.4} & 93.2 & 98.2 & 82.2 & \grayed{97.7} & \cellfirst{98.6} & 6.4 \\
  \tFoundationGeo~\citep{Liu2026:ECCV-foundationgeo} & \cellsecond{98.6} & \cellfirst{98.5} & \cellfirst{96.1} & \cellthird{98.3} & 82.6 & \grayed{96.7} & 94.0 & \cellthird{5.3} \\
  \tPiThree~\citep{Wang2026:ICLR} & \grayed{99.2} & 98.2 & \cellthird{95.2} & \cellsecond{98.5} & \cellthird{85.0} & \cellthird{90.7} & \cellthird{97.9} & \cellfirst{3.2} \\
  \midrule
  \tOursH & \cellfirst{99.2} & 98.1 & \cellsecond{95.4} & \cellfirst{98.7} & \cellfirst{88.8} & \cellthird{90.7} & 95.8 & \cellsecond{3.6} \\
  \bottomrule
\end{tabular}
}
\end{table}

\clearpage

\subsection{Temporal alignment error}
\label{sec:appendix-tae}

In addition to the results on jitter (Table~\ref{tab:paper-temporal}), we evaluate our method using the Temporal Alignment Error (TAE), and report results in Table~\ref{tab:paper-temporal-tae}. This evaluation uses the Spring~\citep{Mehl2023:CVPR}, Sintel~\citep{Butler2012:ECCV}, ScanNet~\citep{Dai2017:CVPR}, Bonn~\citep{Palazzolo2019:IROS}, and KITTI~\citep{Geiger2012:CVPR} datasets.

TAE differs from our jitter metric (Section~\ref{sec:appendix-jitter}) in the computation of correspondences. While jitter relies on track-based correspondences that account for dynamic objects, TAE reprojects predicted depth between consecutive frames using the ground-truth camera intrinsics and poses, and measures the resulting absolute relative error~\citep{Yang2025:ICLR}. TAE assumes a static scene and does not account for independently moving objects.

TAE is ambiguous at occlusions, which can result in non-zero TAE for ground-truth depth and predictions that attain lower errors than the ground truth, demonstrated in Table~\ref{tab:paper-temporal-tae}.
Further non-zero TAE for ground-truth depth can stem from imprecise depth annotations (\eg., from a Kinect sensor), meaning methods may be penalized for better depth coverage. (We define jitter relative to the ground truth, which circumvents this problem.)
On \sintel{} and \spring{}, the depth annotations are sufficiently dense and high-quality that we compute a mask to exclude dynamic content and occlusions, whereas the depth of \scannet{}, \bonn{}, and \kitti{} is sparse and/or low-quality.

\begin{table}[htb]
  \centering
  \ra{1.05}
  \footnotesize
  \setlength{\tabcolsep}{4pt}
  \caption{\textbf{Temporal alignment error (TAE)} after a single affine alignment per video, scaled by $10^2$. Lower is better. TAE measures how well a prediction can be explained by the camera motion annotations. On Sintel and Spring, the depth annotations are accurate and \emph{dense}, enabling us to compute TAE only over unoccluded, static content. For datasets with sparse and/or low-quality depth annotations, we argue that TAE is an unreliable metric since it favors blurrier predictions, and assigns non-zero error to ground-truth depth. The ground-truth and blurred-depth reference rows are excluded from the average rank, as are any non-zero-shot results (which appear grayed-out). ScanNet uses 170-frame clips, on which our model runs in the buffered streaming setting with clips of $N=64$ frames that overlap by $K=32$ (Appendix~\ref{sec:appendix-baselines}).
  }
  \label{tab:paper-temporal-tae}
  \fitwidth{\begin{tabular}{lc@{\hspace{9pt}}c@{\hspace{9pt}}c@{\hspace{9pt}}c@{\hspace{9pt}}c@{\hspace{6pt}}|@{\hspace{6pt}}c}
  \toprule
  Method & \multicolumn{1}{c}{\sintel{}{}} & \multicolumn{1}{c}{\spring{}{}} & \multicolumn{1}{c}{\scannet{}{}} & \multicolumn{1}{c}{\bonn{}{}} & \multicolumn{1}{c}{\kitti{}{}} & \multicolumn{1}{c}{Avg.\ rank$\downarrow$} \\
  \midrule
  \trefGT & 0.13 & 0.06 & 0.44 & 1.97 & 0.82 & -- \\
  \trefBlur & 8.80 & 0.92 & 0.42 & 2.70 & 2.56 & -- \\
  \midrule
  \tDepthPro~\citep{Bochkovskii2025:ICLR} & 14.85 & 10.14 & \grayed{2.09} & 5.09 & 6.51 & 22.8 \\
  \tPixelPerfectDepth~\citep{Xu2025:NeurIPS} & 11.64 & 10.22 & 2.61 & 4.78 & 9.77 & 20.6 \\
  \tInfiniDepth~\citep{Yu2026:CVPR} & 9.83 & 11.11 & 2.59 & 4.84 & 6.06 & 19.6 \\
  \tLapis~\citep{Liu2026:ECCV} & 8.22 & 4.15 & 2.93 & 4.65 & 5.94 & 17.8 \\
  \tFoundationGeo~\citep{Liu2026:ECCV-foundationgeo} & 8.30 & \grayed{2.97} & 2.04 & 4.53 & 5.26 & 16.0 \\
  \tFlashDepth~\citep{Chou2025:ICCV} & 10.84 & \grayed{2.30} & 1.54 & 3.89 & 5.85 & 15.8 \\
  \tMarigoldVTwo~\citep{Pavlovic2026:SIGGRAPHAsia} & 8.72 & 3.08 & 1.35 & 4.23 & 6.41 & 15.4 \\
  \tMoGe~\citep{Wang2025:NeurIPS} & 7.72 & 3.64 & 2.12 & 4.67 & 4.74 & 15.4 \\
  \tMapAnything~\citep{Keetha2026} & 13.76 & \grayed{1.13} & 1.09 & 4.53 & 4.61 & 15.2 \\
  \tLotus~\citep{He2026:arxiv} & 7.90 & 3.13 & 1.07 & 4.30 & 5.98 & 14.4 \\
  \tMoGeThree~\citep{Kong2026:arxiv} & 7.28 & 3.15 & 2.71 & 4.58 & 4.54 & 14.4 \\
  \tUniKThreeD~\citep{Piccinelli2025:CVPR} & 7.28 & 4.79 & \grayed{1.20} & 4.01 & 4.28 & 13.2 \\
  \tChronoDepth~\citep{Shao2025:CVPR} & 4.34 & 3.08 & 0.98 & 6.66 & 5.31 & 13.2 \\
  \tUniDepthTwo~\citep{Piccinelli2025:PAMI} & 7.04 & 3.46 & \grayed{1.33} & 4.04 & 4.05 & 11.8 \\
  \tMetricAnything~\citep{Ma2026:ECCV} & 5.92 & 2.91 & \grayed{0.95} & 3.84 & 4.87 & 11.2 \\
  \tVGGTOmega~\citep{Wang2026:CVPR} & 9.99 & 0.95 & \grayed{0.54} & 3.61 & \cellsecond{3.45} & 9.0 \\
  \tDVD~\citep{Zhang2026:ICMLW} & 4.00 & 1.61 & \cellfirst{0.48} & 3.76 & 5.17 & 8.2 \\
  \tDepthAnythingThree~\citep{Lin2026:ICLR} & 5.18 & \grayed{1.16} & 0.72 & 3.39 & 4.67 & 7.8 \\
  \tDepthCrafter~\citep{Hu2025} & 3.85 & 0.98 & 0.59 & 3.72 & 4.14 & 6.4 \\
  \tDAGE~\citep{Ngo2026:CVPR} & \cellsecond{3.30} & \grayed{0.96} & \grayed{0.46} & 3.41 & 4.11 & 5.0 \\
  \tPiThree~\citep{Wang2026:ICLR} & \cellthird{3.71} & 1.04 & \grayed{0.39} & 3.49 & \cellthird{3.67} & 4.8 \\
  \tVGGT~\citep{Wang2025:CVPR} & 4.65 & 0.91 & \grayed{0.46} & \cellsecond{3.07} & 3.68 & 4.5 \\
  \tVideoDepthAnything~\citep{Chen2025:CVPR} & 3.98 & \cellsecond{0.77} & \cellthird{0.55} & \cellthird{3.27} & 3.84 & \cellsecond{3.6} \\
  \tVeloDepth~\citep{Piccinelli2026:3DV} & 6.83 & \cellfirst{0.61} & \grayed{0.43} & \cellfirst{2.93} & \cellfirst{3.30} & \cellfirst{3.5} \\
  \midrule
  \tOursH & \cellfirst{2.05} & \cellthird{0.90} & \cellsecond{0.51} & 3.61 & 4.04 & \cellthird{3.8} \\
  \bottomrule
\end{tabular}
}
\end{table}

\subsection{Runtime analysis}
\label{sec:appendix-runtime-baselines}

We measure the throughput and peak memory of \ours{} and baselines and report them in Table~\ref{tab:runtime-baselines}.

\mypara{Profiling protocol.}
We run every method on one B200 GPU.
Each method receives the same 32-frame \spring{} clip, resized beforehand to VGA ($640\times480$, 4:3) or to $1280\times720$, $1920\times1080$, or $3840\times2160$ (16:9).
Video methods process the whole clip at once.
Single-image methods process its first 8 frames, one call per frame. We time each call with the input already on the GPU and report the median of 11 timed calls after 3 warm-up calls. Throughput is the number of frames processed per second.
Peak memory is the most memory PyTorch reserves on the GPU during the timed calls, including weights and its allocator cache.
Every method runs its official inference code in eager PyTorch under bf16 autocast, with its default attention kernel. When the pinned PyTorch version is too old for the B200, we use PyTorch 2.8. \ours{} runs in PyTorch 2.14, whose default on the B200 is cuDNN attention. \depthcrafter{} keeps its whole pipeline on the GPU, without the CPU offload its release enables by default.
For fairness, no method in Table~\ref{tab:runtime-baselines} uses \texttt{torch.compile}.

\mypara{Native resolution.}
For runtime analysis, we have every method process the input at the native resolution, rounded to a multiple of its patch size, except \depthpro{} and \metricanything{}, which resize every input to $1536\times1536$. For \ours{} and the video baselines, we report throughput averaged over the frames of the whole clip.

\mypara{Baselines.}
The measurements differ in which stages they include. \ours{} estimates its focal length in every timed call, so its throughput includes its focal-length estimator, as \depthpro{}'s includes its focal-length head, and its parameter count includes the estimator. Only \metricanything{}, which has no estimator, receives the true focal length. \velodepth{}'s throughput includes its optical flow, which the released code computes on the CPU. \infinidepth{}'s throughput excludes the \mogetwo{} reference that it computes for every frame, so its reported throughput is an upper bound. Following its official inference code, \marigoldvtwo{} loads its diffusion transformer with 4-bit weights.

\begin{table}[!ht]
  \centering\ra{0.88}\scriptsize\setlength{\tabcolsep}{4pt}
  \caption{\textbf{Throughput and peak memory.} We profile all methods with the same protocol on the same Spring clip, on a B200 at bf16. All methods, ours included, run in PyTorch eager mode, so that differences in runtime reflect the models themselves. \ours{} and the video methods process 32 frames in one call. Single-image methods process the first 8 frames, one frame per call. Throughput is in frames per second and peak memory in GiB. VGA is $640\times480$ (4:3); the other input sizes are 16:9. Every method except those marked $^{\mathrm{f}}$ runs at native resolution: the input resolution rounded to a multiple of the method's patch size. \ours{} estimates the focal length. Bold marks the highest throughput in each column, excluding methods marked $^{\mathrm{f}}$. OOM:~out of memory. $^{\mathrm{s}}$:~single-image method; peak memory is for one frame. $^{\dagger}$:~measured end to end, including pre- and post-processing. $^{\mathrm{f}}$:~fixed $1536\times1536$ input, whatever the input size.}
  \label{tab:runtime-baselines}
  \fitwidth{\begin{tabular}{@{}lrrrrrrrrr@{}}
  \toprule
  & & \multicolumn{4}{c}{Throughput (FPS)~$\uparrow$} & \multicolumn{4}{c}{Peak reserved (GiB)~$\downarrow$} \\
  \cmidrule(lr){3-6}\cmidrule(l){7-10}
  Method & Params (M) & VGA & 720p & 1080p & 4K & VGA & 720p & 1080p & 4K \\
  \midrule
  \tInfiniDepth\smash{$^{\mathrm{s}}$} & 307 & 20.3 & 9.5 & 4.1 & 0.5 & 3.3 & 3.6 & 4.4 & 8.0 \\
  \tFoundationGeo\smash{$^{\mathrm{s}}$} & 313 & 29.1 & 13.6 & 5.2 & 1.0 & 2.5 & 3.5 & 5.9 & 18.3 \\
  \tMoGe\smash{$^{\mathrm{s}}$} & 326 & 48.2 & 29.3 & 12.4 & 1.5 & 2.4 & 3.5 & 5.4 & 16.0 \\
  \tFlashDepth & 338 & 61.7 & 39.2 & 15.2 & 1.6 & 3.0 & 5.1 & 8.9 & 29.4 \\
  \tUniDepthTwo\smash{$^{\mathrm{s}}$} & 354 & 52.3 & 30.0 & 12.4 & 1.4 & 2.9 & 4.7 & 7.9 & 24.8 \\
  \tUniKThreeD\smash{$^{\mathrm{s}}$} & 359 & 48.4 & 29.3 & 12.3 & 1.4 & 2.9 & 4.8 & 7.6 & 23.5 \\
  \tVideoDepthAnything & 383 & 95.5 & 26.9 & 8.6 & OOM & 8.6 & 21.6 & 44.0 & OOM \\
  \tVeloDepth & 390 & 50.0 & 31.2 & 16.8 & \textbf{3.9} & 3.4 & 6.9 & 13.3 & 48.6 \\
  Pixel-Perfect\,Depth\smash{$^{\mathrm{s}}$} & 804 & 7.2 & 2.7 & 0.8 & OOM & 4.4 & 6.6 & 16.7 & OOM \\
  Lapis\smash{$^{\mathrm{s}}$} & 828 & 26.6 & 19.0 & 9.2 & 1.6 & 3.7 & 4.9 & 6.9 & 18.0 \\
  \tMetricAnything\smash{$^{\mathrm{s},\mathrm{f}}$} & 877 & 8.4 & 8.4 & 8.4 & 8.4 & 9.9 & 9.6 & 9.9 & 9.3 \\
  \tDepthPro\smash{$^{\mathrm{s},\mathrm{f}}$} & 951 & 23.5 & 23.5 & 23.4 & 23.4 & 8.0 & 8.5 & 8.0 & 8.6 \\
  \tPiThree & 959 & 35.0 & 5.4 & 1.2 & 0.1 & 8.2 & 15.2 & 24.9 & 78.2 \\
  \tVGGTOmega & 1144 & 49.5 & 7.9 & 1.8 & 0.1 & 9.0 & 18.3 & 36.0 & 126.0 \\
  \tVGGT & 1191 & 26.6 & 4.1 & 0.9 & OOM & 13.9 & 29.7 & 68.0 & OOM \\
  \tMapAnything & 1228 & 48.1 & 7.7 & 1.7 & OOM & 13.7 & 27.0 & 51.9 & OOM \\
  \tMoGeThree\smash{$^{\mathrm{s}}$} & 1251 & 14.7 & 7.6 & 3.3 & 0.5 & 8.6 & 10.6 & 15.2 & 40.3 \\
  \tDAGE & 1285 & 111.5 & 38.8 & 12.1 & 1.4 & 16.0 & 34.4 & 38.4 & 91.8 \\
  \tDepthAnythingThree & 1540 & 24.6 & 4.2 & 1.0 & OOM & 21.7 & 52.0 & 114.6 & OOM \\
  \tDVD\smash{$^{\dagger}$} & 1546 & 31.6 & 9.7 & 3.6 & OOM & 11.5 & 25.9 & 52.5 & OOM \\
  \tChronoDepth\smash{$^{\dagger}$} & 2254 & 7.0 & 2.5 & 0.9 & OOM & 20.8 & 48.8 & 106.3 & OOM \\
  \tDepthCrafter\smash{$^{\dagger}$} & 2254 & 10.7 & 3.6 & 1.4 & OOM & 20.8 & 48.2 & 106.8 & OOM \\
  \tLotus\smash{$^{\mathrm{s},\dagger}$} & 17588 & 1.0 & 0.5 & 0.2 & $<$0.1 & 34.7 & 37.4 & 42.2 & 69.3 \\
  Marigold\,V2\smash{$^{\mathrm{s}}$} & 21407 & 2.8 & 2.5 & 1.1 & $<$0.1 & 15.3 & 18.3 & 23.6 & 45.8 \\
  \midrule
  \tOursS & 48 & \textbf{347.0} & \textbf{143.9} & \textbf{44.8} & 3.6 & 1.9 & 3.8 & 8.0 & 31.9 \\
  \tOursB & 120 & 269.7 & 82.5 & 23.6 & 1.8 & 2.6 & 6.6 & 14.0 & 64.8 \\
  \tOursL & 357 & 150.1 & 33.7 & 9.1 & 0.7 & 4.0 & 10.6 & 21.7 & 82.6 \\
  \tOursH & 896 & 95.9 & 20.2 & 5.3 & 0.4 & 8.3 & 16.3 & 29.5 & 114.6 \\
  \bottomrule
\end{tabular}
}
\end{table}

\clearpage

\subsection{Additional qualitative examples}
\label{sec:appendix-qualitative}

\begin{figure*}[tbhp]
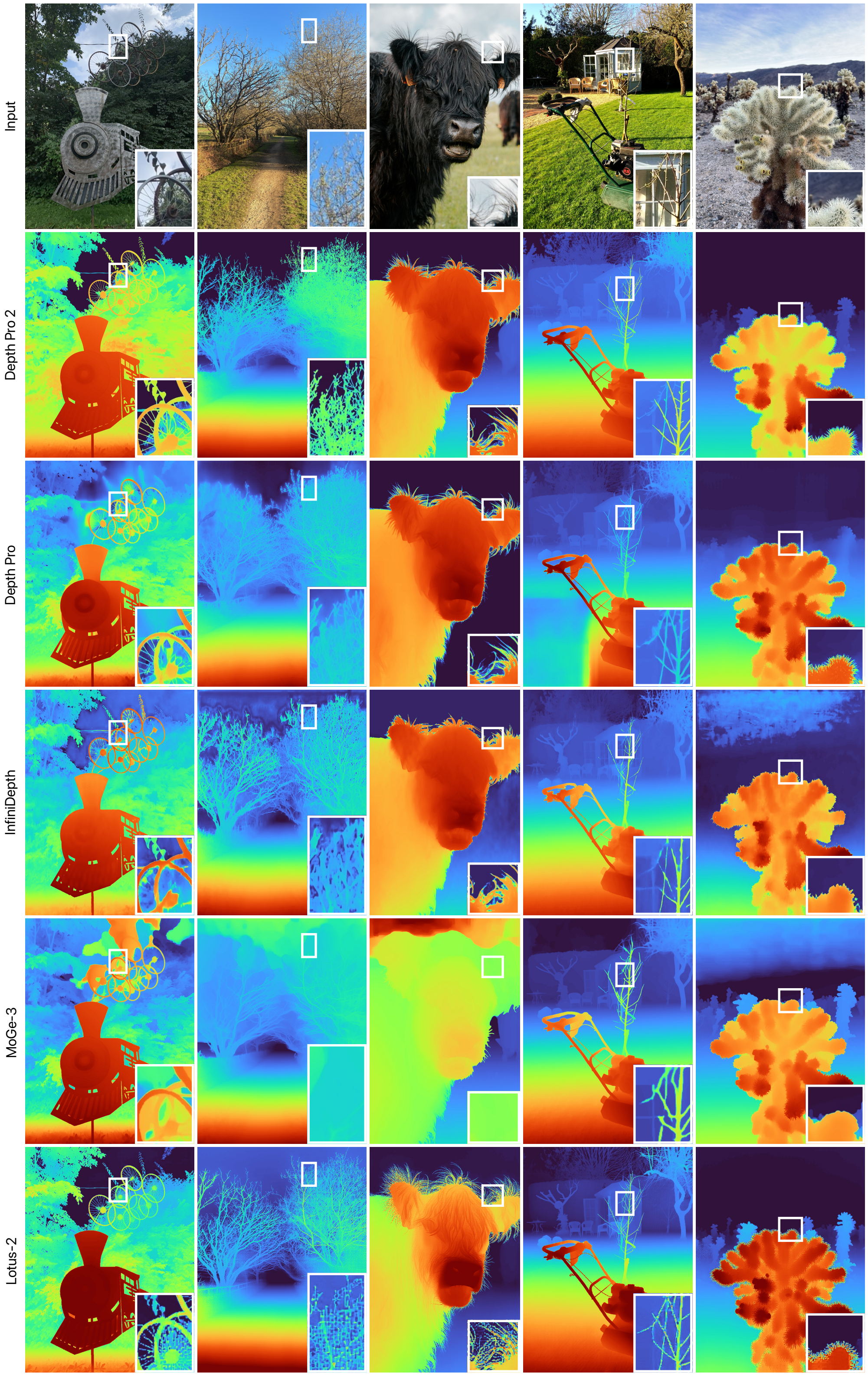

  \centering
  \qualfigure[rows=methods, zoom=bottom right, inset=0.35, qual accent,
                zoom border=true, zoom border methods={}, box width=1pt, box color=white]{spokes, branches, cow, mower, cholla}
  \caption{\textbf{Additional qualitative examples.} As in Figure~\ref{fig:qualitative-sharpness}, we compare \ours to a number of methods from Table~\ref{tab:paper-sharpness}. Insets zoom into the highlighted regions. Every method runs at its own shipped default resolution.}
  \label{fig:qualitative-sharpness-portrait}
\end{figure*}

Figures~\ref{fig:qualitative-sharpness-landscape} and~\ref{fig:qualitative-sharpness-portrait} show further qualitative comparisons to representative baselines.

\begin{figure*}[tbhp]
  \centering
  \qualfigure[rows=methods, zoom=bottom left, inset=0.4, qual accent,
                zoom border=true, zoom border methods={}, box width=1pt, box color=white]{balloon, fence, bonsai}
  \caption{\textbf{Additional qualitative examples.} As in Figure~\ref{fig:qualitative-sharpness}, we compare \ours to a number of methods from Table~\ref{tab:paper-sharpness}. Insets zoom into the highlighted regions. Every method runs at its own shipped default resolution.}
  \label{fig:qualitative-sharpness-landscape}
\end{figure*}

\clearpage

\section{Model family on individual datasets}
\label{sec:appendix-family-datasets}

For a comprehensive comparison of our model variants, we repeat the evaluation on metric accuracy, boundary sharpness, and temporal consistency.

\begin{table}[!ht]
  \centering\ra{1.05}\footnotesize\setlength{\tabcolsep}{2pt}
  \caption{\textbf{Zero-shot metric depth, single image.}
  $\delta_1$ (higher is better) on the datasets of Table~\ref{tab:paper-metric-image}.
  GT uses ground-truth intrinsics, and pred uses the focal length from our field-of-view estimator, shared by all four models. Shading: \textfirst{best}, \textsecond{second}, \textthird{third}.}
  \label{tab:family-metric-image}
  \fitwidth{\begin{tabular}{lcc@{\hspace{5pt}}cc@{\hspace{5pt}}cc@{\hspace{5pt}}cc@{\hspace{5pt}}cc@{\hspace{5pt}}cc@{\hspace{5pt}}cc@{\hspace{5pt}}cc@{\hspace{5pt}}cc@{\hspace{5pt}}cc@{\hspace{5pt}}cc}
  \toprule
  & \multicolumn{8}{c}{Real, indoor} & \multicolumn{4}{c}{Real, indoor+outdoor} & \multicolumn{4}{c}{Real, outdoor} & \multicolumn{6}{c}{Synthetic} \\
  \cmidrule(lr){2-9}\cmidrule(lr){10-13}\cmidrule(lr){14-17}\cmidrule(lr){18-23}
  Model & \multicolumn{2}{c}{\hdribims} & \multicolumn{2}{c}{\hdrmiddleburyf} & \multicolumn{2}{c}{\hdrbooster} & \multicolumn{2}{c}{\hdrhammer} & \multicolumn{2}{c}{\eththreed} & \multicolumn{2}{c}{\hdrdiode} & \multicolumn{2}{c}{\kitti} & \multicolumn{2}{c}{\hdrnuscenes} & \multicolumn{2}{c}{\sintel} & \multicolumn{2}{c}{\spring} & \multicolumn{2}{c}{\dydtof} \\
  \cmidrule(lr){2-3}\cmidrule(lr){4-5}\cmidrule(lr){6-7}\cmidrule(lr){8-9}\cmidrule(lr){10-11}\cmidrule(lr){12-13}\cmidrule(lr){14-15}\cmidrule(lr){16-17}\cmidrule(lr){18-19}\cmidrule(lr){20-21}\cmidrule(lr){22-23}
   & GT & pred & GT & pred & GT & pred & GT & pred & GT & pred & GT & pred & GT & pred & GT & pred & GT & pred & GT & pred & GT & pred \\
  \midrule
  \tOursS & 78.2 & 74.1 & 40.5 & 45.0 & \cellthird{42.1} & \cellthird{28.8} & 49.8 & 52.8 & 77.8 & 74.1 & 56.6 & 56.4 & 90.8 & 73.1 & 78.9 & 75.6 & 40.4 & 29.9 & \cellfirst{26.7} & \cellthird{31.5} & 59.3 & 49.8 \\
  \tOursB & \cellthird{85.6} & \cellthird{81.1} & \cellthird{49.9} & \cellthird{61.2} & 41.0 & 28.7 & \cellthird{62.0} & \cellthird{60.7} & \cellthird{83.7} & \cellthird{75.0} & \cellthird{62.7} & \cellthird{61.0} & \cellthird{95.5} & \cellthird{78.9} & \cellthird{85.9} & \cellthird{82.0} & \cellthird{45.3} & \cellthird{35.0} & 17.6 & 21.2 & \cellthird{70.3} & \cellthird{64.9} \\
  \tOursL & \cellsecond{94.2} & \cellfirst{88.5} & \cellsecond{67.4} & \cellsecond{74.9} & \cellsecond{55.8} & \cellsecond{34.8} & \cellsecond{78.3} & \cellsecond{75.9} & \cellsecond{91.5} & \cellsecond{83.5} & \cellsecond{71.2} & \cellsecond{66.3} & \cellsecond{97.8} & \cellsecond{82.1} & \cellsecond{88.6} & \cellsecond{85.4} & \cellsecond{53.8} & \cellsecond{40.1} & \cellsecond{21.5} & \cellsecond{33.2} & \cellsecond{78.2} & \cellsecond{65.8} \\
  \tOursH & \cellfirst{94.4} & \cellsecond{86.6} & \cellfirst{73.9} & \cellfirst{79.1} & \cellfirst{65.7} & \cellfirst{43.1} & \cellfirst{89.4} & \cellfirst{86.3} & \cellfirst{97.8} & \cellfirst{91.8} & \cellfirst{74.2} & \cellfirst{69.2} & \cellfirst{98.2} & \cellfirst{87.9} & \cellfirst{90.9} & \cellfirst{87.7} & \cellfirst{60.2} & \cellfirst{43.4} & \cellthird{20.5} & \cellfirst{37.0} & \cellfirst{80.5} & \cellfirst{78.2} \\
  \bottomrule
\end{tabular}
}
\end{table}

\begin{table}[!ht]
  \centering\ra{1.05}\footnotesize\setlength{\tabcolsep}{4pt}
  \caption{\textbf{Zero-shot metric depth, video.}
  $\delta_1$ (higher is better) on clips of up to 50 frames, as in Table~\ref{tab:paper-metric-video}.
  GT and pred intrinsics are as in Table~\ref{tab:family-metric-image}. Shading: \textfirst{best}, \textsecond{second}, \textthird{third}.}
  \label{tab:family-metric-video}
  \fitwidth{\begin{tabular}{lcc@{\hspace{5pt}}cc@{\hspace{5pt}}cc@{\hspace{5pt}}cc}
  \toprule
  & \multicolumn{2}{c}{Real, outdoor} & \multicolumn{6}{c}{Synthetic} \\
  \cmidrule(lr){2-3}\cmidrule(lr){4-9}
  Model & \multicolumn{2}{c}{\kitti} & \multicolumn{2}{c}{\sintel} & \multicolumn{2}{c}{\spring} & \multicolumn{2}{c}{\dydtof} \\
  \cmidrule(lr){2-3}\cmidrule(lr){4-5}\cmidrule(lr){6-7}\cmidrule(lr){8-9}
   & GT & pred & GT & pred & GT & pred & GT & pred \\
  \midrule
  \tOursS & 93.8 & 90.2 & 29.9 & 18.3 & \cellsecond{34.5} & \cellthird{24.3} & 64.0 & 59.5 \\
  \tOursB & \cellthird{97.1} & \cellthird{93.8} & \cellsecond{45.0} & \cellthird{29.4} & \cellthird{30.9} & 18.8 & \cellsecond{82.3} & \cellthird{77.2} \\
  \tOursL & \cellsecond{98.0} & \cellsecond{97.0} & \cellthird{41.6} & \cellsecond{29.6} & 27.7 & \cellsecond{27.7} & \cellthird{78.9} & \cellsecond{80.4} \\
  \tOursH & \cellfirst{98.1} & \cellfirst{97.2} & \cellfirst{51.2} & \cellfirst{41.4} & \cellfirst{51.7} & \cellfirst{45.0} & \cellfirst{86.2} & \cellfirst{94.8} \\
  \bottomrule
\end{tabular}
}
\end{table}

\begin{table}[!ht]
  \centering\ra{1.05}\footnotesize\setlength{\tabcolsep}{4pt}
  \caption{\textbf{Zero-shot boundary sharpness.}
  Boundary $F1$ (higher is better) on the datasets of Table~\ref{tab:paper-sharpness}. Shading: \textfirst{best}, \textsecond{second}, \textthird{third}.}
  \label{tab:family-sharpness}
  \begin{subtable}[t]{0.54\linewidth}
    \centering
    \caption{Single image}
    \label{tab:family-sharpness-image}
    \begin{tabular}{lccccc}
  \toprule
  & \multicolumn{2}{c}{Real, indoor} & \multicolumn{3}{c}{Synthetic} \\
  \cmidrule(lr){2-3}\cmidrule(lr){4-6}
  Model & \hdribims & \hdrmiddleburyf & \sintel & \spring & \dydtof \\
  \midrule
  \tOursS & 12.6 & 14.9 & 24.3 & 10.3 & 18.1 \\
  \tOursB & \cellthird{16.3} & \cellthird{19.8} & \cellthird{32.9} & \cellthird{13.2} & \cellthird{25.4} \\
  \tOursL & \cellsecond{19.5} & \cellsecond{21.5} & \cellsecond{38.5} & \cellsecond{14.2} & \cellsecond{30.7} \\
  \tOursH & \cellfirst{20.5} & \cellfirst{21.9} & \cellfirst{40.2} & \cellfirst{16.3} & \cellfirst{31.6} \\
  \bottomrule
\end{tabular}

  \end{subtable}\hfill
  \begin{subtable}[t]{0.42\linewidth}
    \centering
    \caption{Video}
    \label{tab:family-sharpness-video}
    \begin{tabular}{lccc}
  \toprule
  & \multicolumn{3}{c}{Synthetic} \\
  \cmidrule(lr){2-4}
  Model & \sintel & \spring & \dydtof \\
  \midrule
  \tOursS & 25.7 & 10.7 & 19.3 \\
  \tOursB & \cellthird{35.8} & \cellthird{13.1} & \cellthird{26.9} \\
  \tOursL & \cellsecond{42.2} & \cellfirst{17.1} & \cellsecond{32.9} \\
  \tOursH & \cellfirst{43.4} & \cellsecond{16.8} & \cellfirst{34.4} \\
  \bottomrule
\end{tabular}

  \end{subtable}
\end{table}

\begin{table}[!ht]
  \centering\ra{1.05}\footnotesize\setlength{\tabcolsep}{4pt}
  \caption{\textbf{Zero-shot temporal consistency.}
  (a) Jitter on the datasets of Table~\ref{tab:paper-temporal}, scaled by $10^2$. (b) TAE on the datasets of Table~\ref{tab:paper-temporal-tae}, scaled by $10^2$. Lower is better. Shading: \textfirst{best}, \textsecond{second}, \textthird{third}.}
  \begin{subtable}[t]{0.4\linewidth}
  \centering
    \caption{Jitter.}
    \label{tab:family-temporal}
    \begin{tabular}{lcc}
  \toprule
  Model & \sintel & \spring \\
  \midrule
  \tOursS & 2.43 & 1.13 \\
  \tOursB & \cellthird{1.92} & \cellthird{0.98} \\
  \tOursL & \cellsecond{1.55} & \cellfirst{0.85} \\
  \tOursH & \cellfirst{1.39} & \cellsecond{0.86} \\
  \bottomrule
\end{tabular}

    \end{subtable}\hfill
  \begin{subtable}[t]{0.6\linewidth}
    \centering
    \caption{Temporal alignment error.}
    \label{tab:family-temporal-tae}
    \begin{tabular}{lccccc}
  \toprule
  Model & \sintel & \spring & \scannet & \bonn & \kitti \\
  \midrule
  \tOursS & 3.18 & \cellthird{1.13} & 0.90 & 3.77 & 4.46 \\
  \tOursB & \cellthird{2.68} & \cellsecond{1.05} & \cellthird{0.69} & \cellsecond{3.62} & \cellthird{4.22} \\
  \tOursL & \cellsecond{2.16} & \cellfirst{0.90} & \cellsecond{0.56} & \cellthird{3.67} & \cellsecond{4.10} \\
  \tOursH & \cellfirst{2.05} & \cellfirst{0.90} & \cellfirst{0.51} & \cellfirst{3.61} & \cellfirst{4.04} \\
  \bottomrule
\end{tabular}

  \end{subtable}
\end{table}

\clearpage

\section{Applications}
\label{sec:appendix-applications}

\subsection{Bokeh}
\label{sec:appendix-bokeh}

\begin{table}[tbhp]
  \centering\ra{1.05}\footnotesize\setlength{\tabcolsep}{5pt}
  \caption{\textbf{Depth-guided bokeh rendering on \realbokeh.} Metric depth estimates from each method are provided to the same frozen \bokehme renderer without per-image alignment or fitting. Results are averaged over 220 test scenes and ordered by Boundary LPIPS. $\sim$ denotes an unreported inference resolution inferred from the method's configuration constraints. $^{\dagger}$ Run at the native input resolution.}
  \label{tab:bokeh-realbokeh}
  \fitwidth{  \begin{tabular}{lw{r}{4em}@{$\times$}w{l}{4em}ccccc}
    \toprule
    Depth source & \multicolumn{2}{c}{Inference resolution} & PSNR $\uparrow$ & SSIM $\uparrow$ & LPIPS $\downarrow$ & Boundary LPIPS $\downarrow$ & DISTS $\downarrow$ \\
    \midrule
    \dathreeng$^{\dagger}$ & 2002 & 1498 & 22.47 & 0.831 & 0.341 & 0.630 & 0.290 \\
    \unikthreedl & $\sim\!$896 & 672 & 23.01 & 0.841 & 0.320 & 0.557 & 0.248 \\
    \mogetwol & 966 & 728 & 23.27 & 0.838 & 0.319 & 0.537 & 0.226 \\
    \unidepthtwol & $\sim\!$896 & 672 & 23.22 & \cellthird{0.845} & \cellthird{0.310} & 0.536 & 0.236 \\
    \mogethreel & 966 & 728 & \cellthird{23.28} & 0.840 & 0.317 & 0.535 & \cellsecond{0.222} \\
    \depthpro & 1536 & 1536 & 23.20 & 0.838 & 0.319 & 0.530 & 0.236 \\
    \mogethreeg & 966 & 728 & \cellthird{23.28} & 0.841 & 0.313 & \cellthird{0.525} & \cellthird{0.223} \\
    \dathreeng & 504 & 378 & \cellsecond{23.44} & \cellsecond{0.848} & \cellsecond{0.306} & \cellsecond{0.505} & 0.230 \\
    \midrule
    \tOursH & 2000 & 1500 & \cellfirst{23.95} & \cellfirst{0.858} & \cellfirst{0.291} & \cellfirst{0.476} & \cellfirst{0.215} \\
    \bottomrule
  \end{tabular}
}
\end{table}

\begin{figure}[tbhp]
  \centering
  \setlength{\tabcolsep}{1pt}
  \renewcommand{\arraystretch}{0}
  \begin{tabular}{@{}cccc@{}}
    \footnotesize\sffamily Input & \footnotesize\sffamily \dathreeng & \footnotesize\sffamily \ours (ours) & \footnotesize\sffamily Real $f/2$ \\[1pt]
    \includegraphics[width=0.244\linewidth]{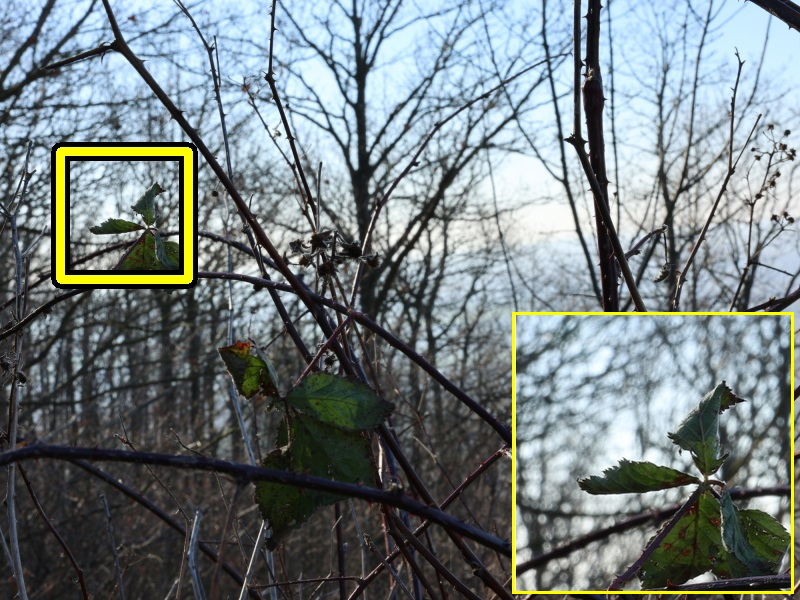} & \includegraphics[width=0.244\linewidth]{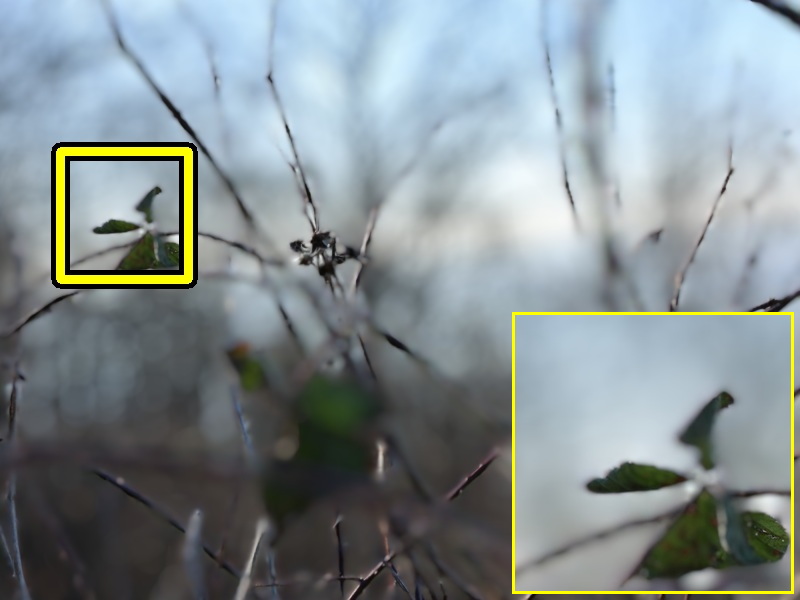} & \includegraphics[width=0.244\linewidth]{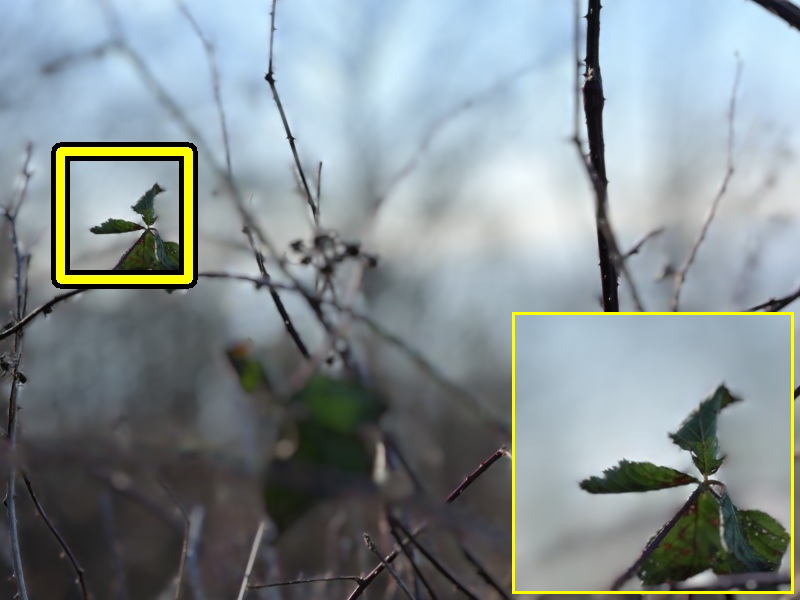} & \includegraphics[width=0.244\linewidth]{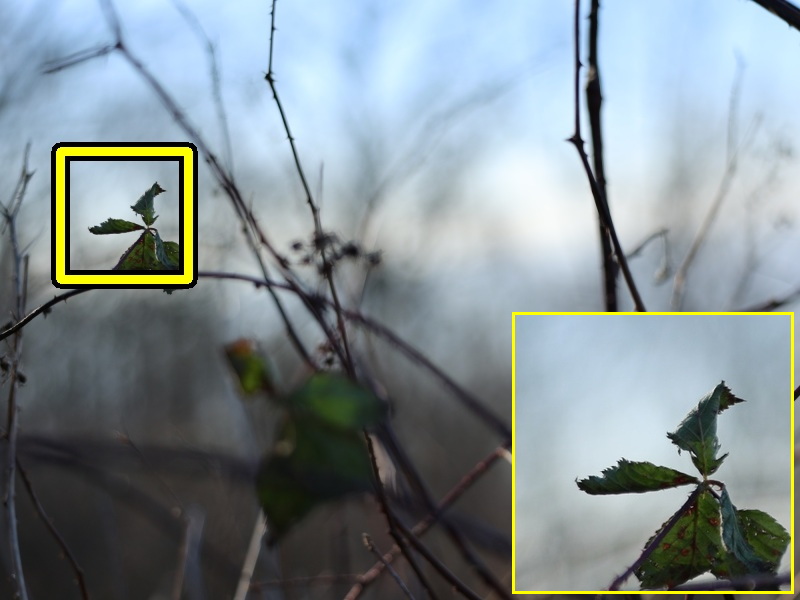} \\
     & \includegraphics[width=0.244\linewidth]{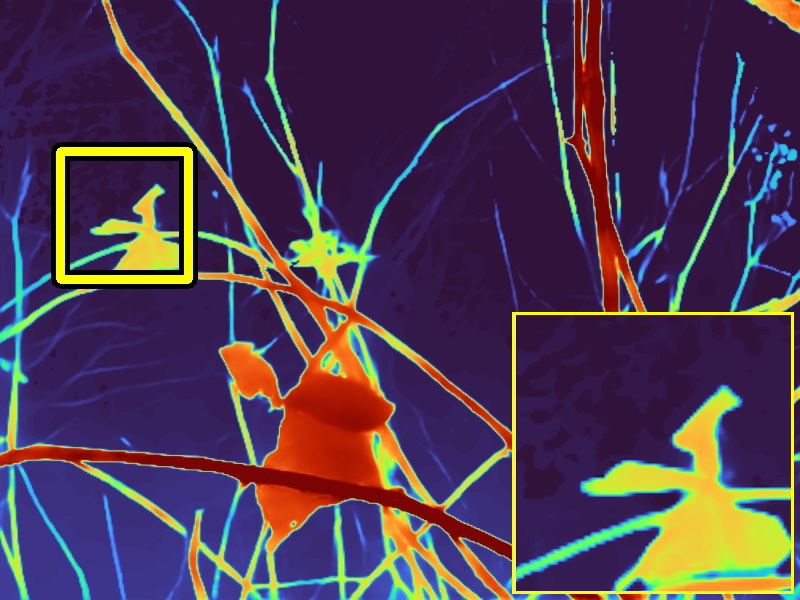} & \includegraphics[width=0.244\linewidth]{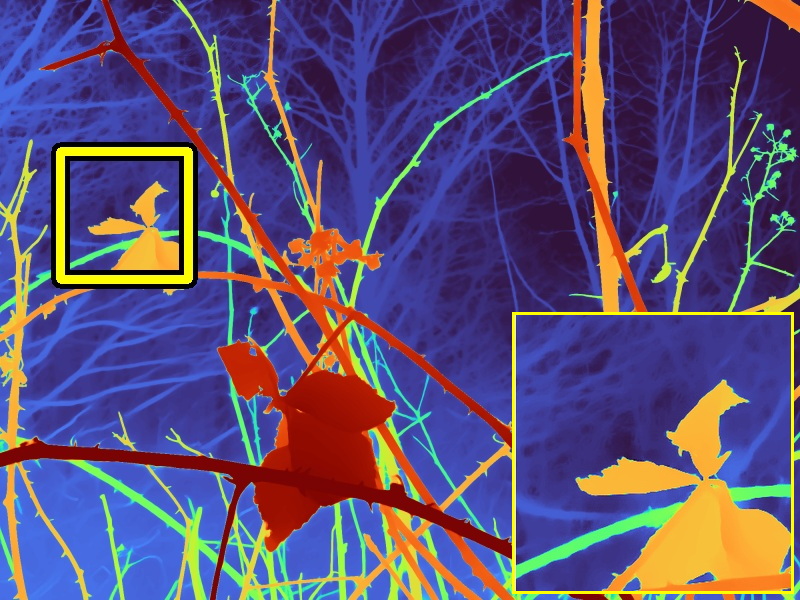} &  \\
    \includegraphics[width=0.244\linewidth]{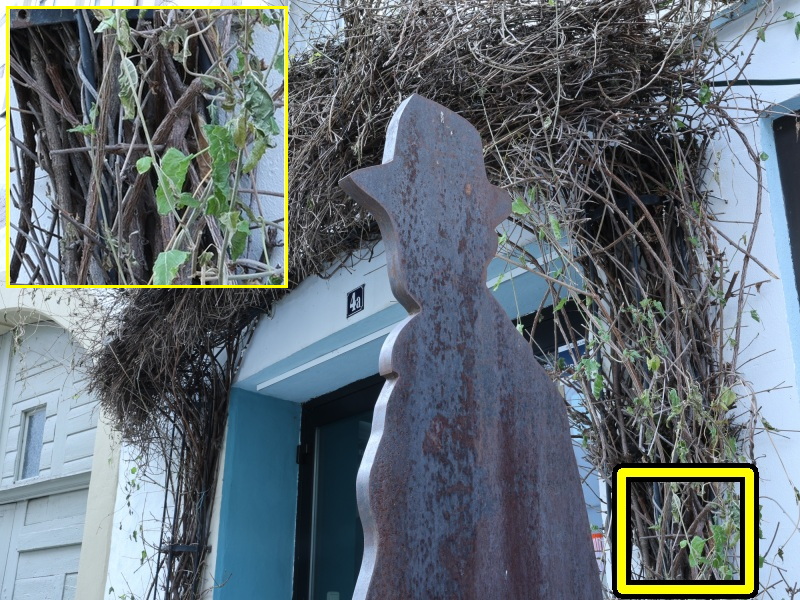} & \includegraphics[width=0.244\linewidth]{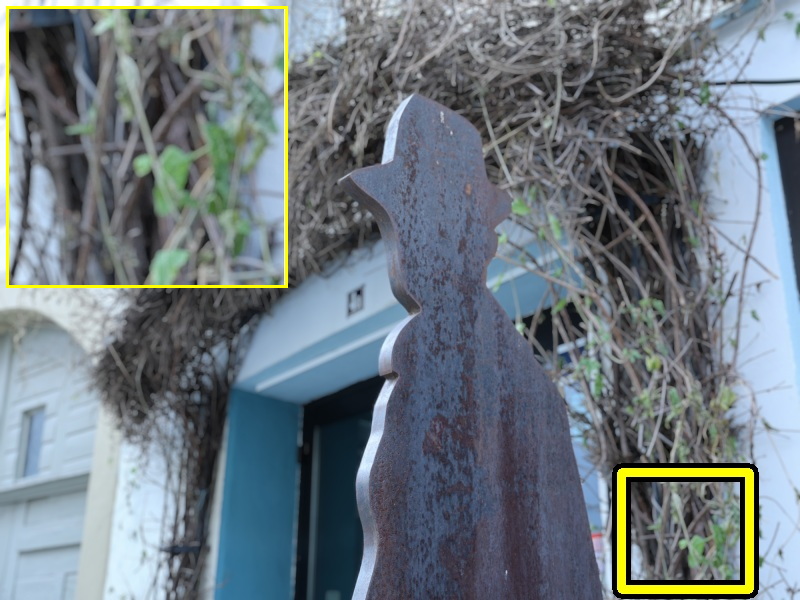} & \includegraphics[width=0.244\linewidth]{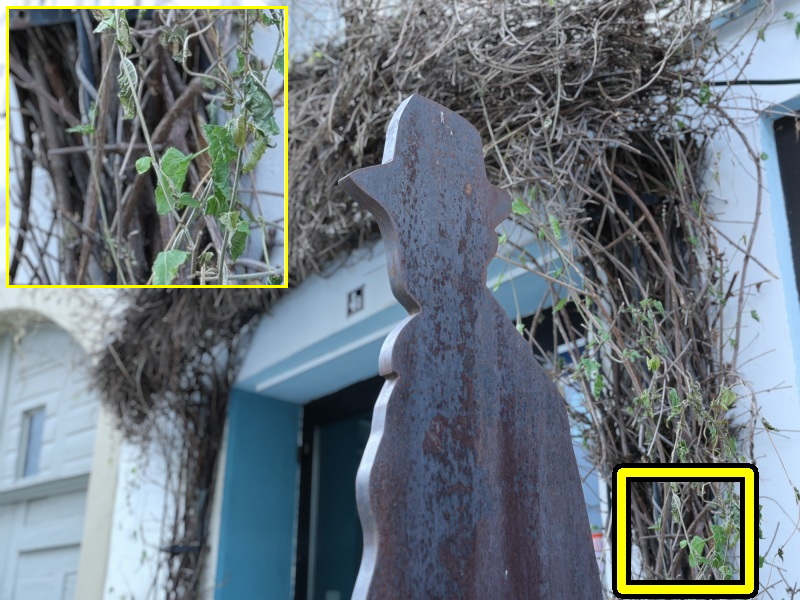} & \includegraphics[width=0.244\linewidth]{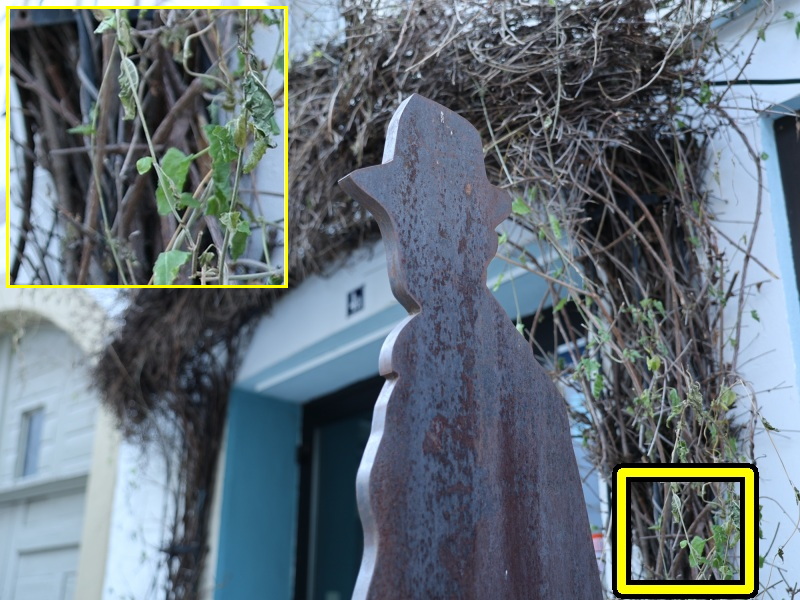} \\
     & \includegraphics[width=0.244\linewidth]{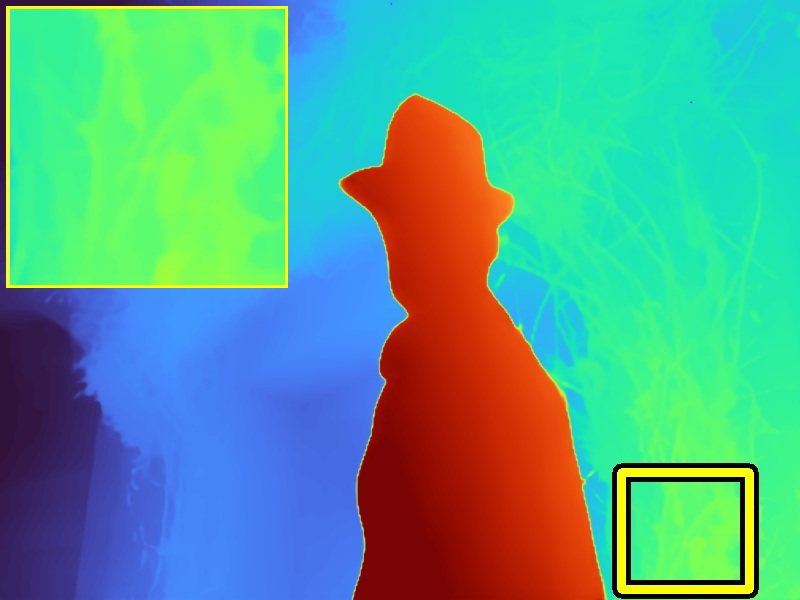} & \includegraphics[width=0.244\linewidth]{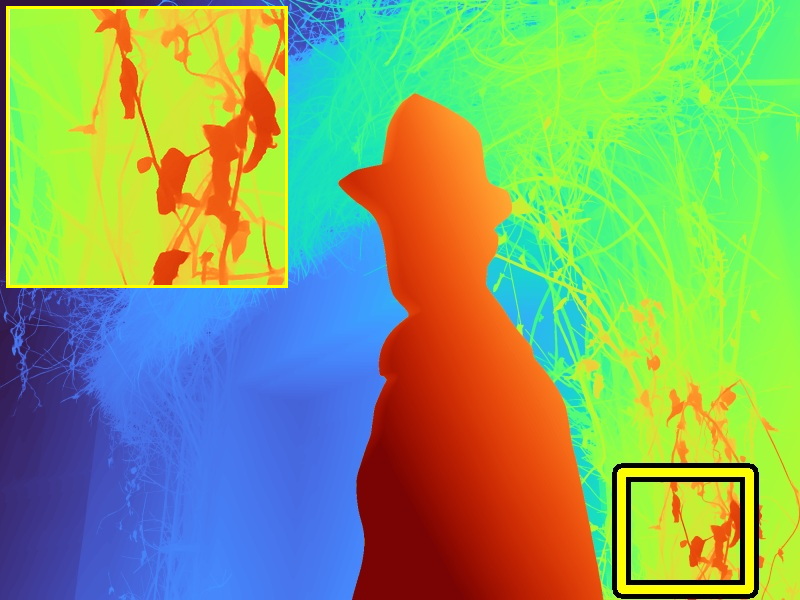} &  \\
    \includegraphics[width=0.244\linewidth]{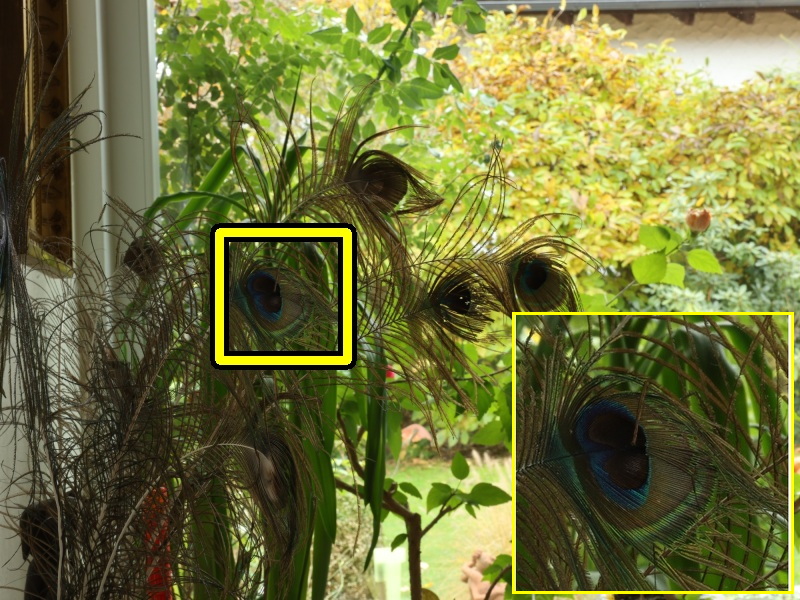} & \includegraphics[width=0.244\linewidth]{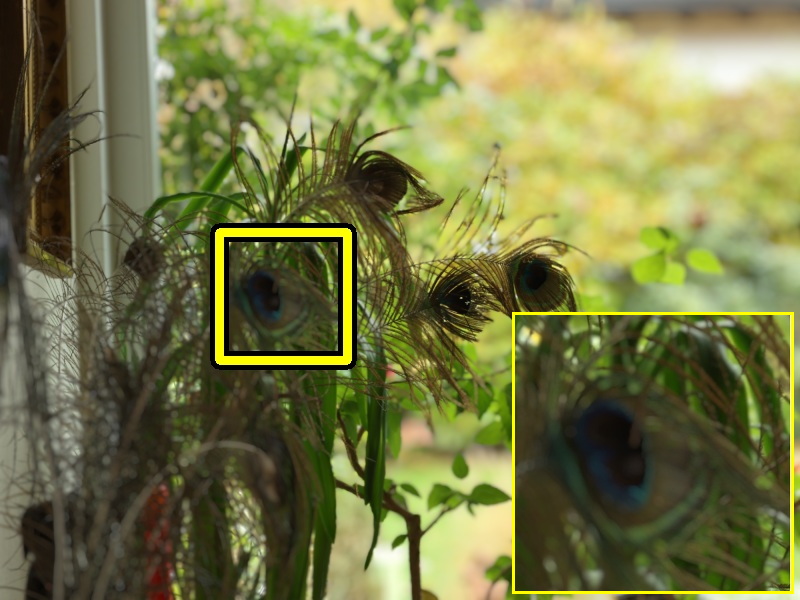} & \includegraphics[width=0.244\linewidth]{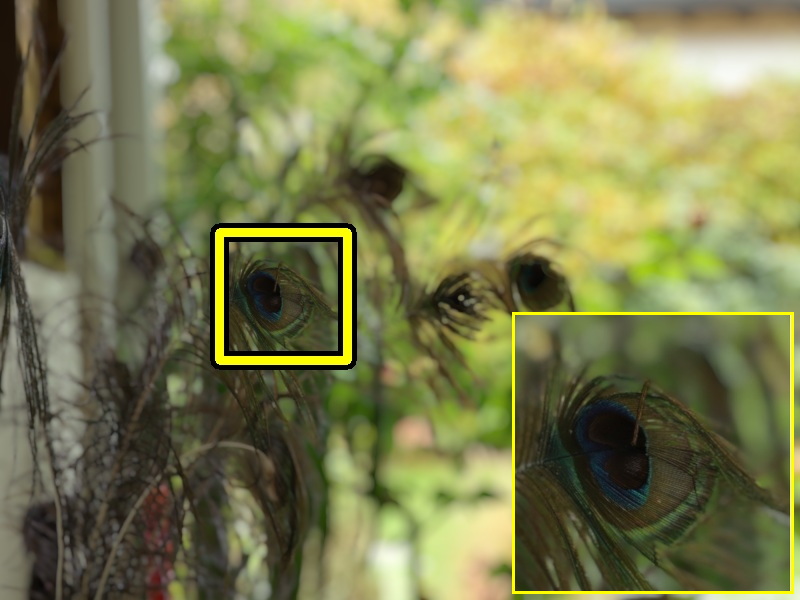} & \includegraphics[width=0.244\linewidth]{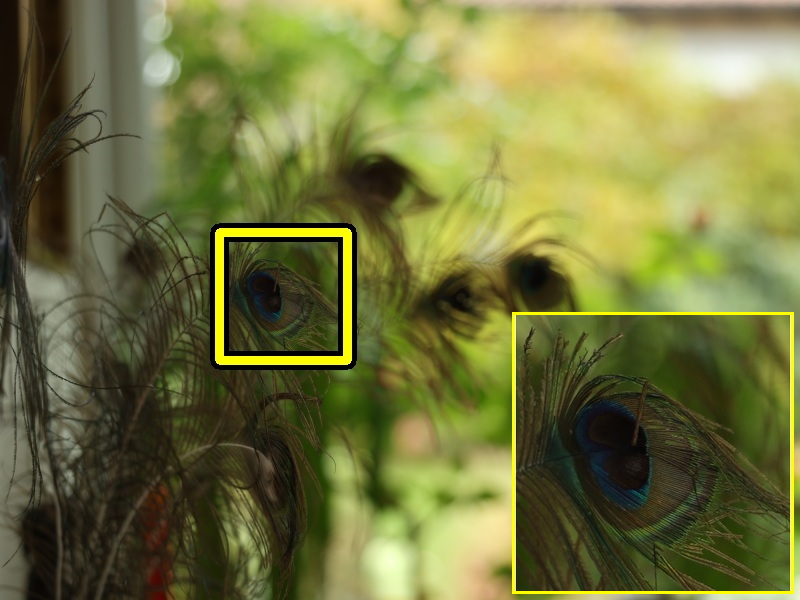} \\
     & \includegraphics[width=0.244\linewidth]{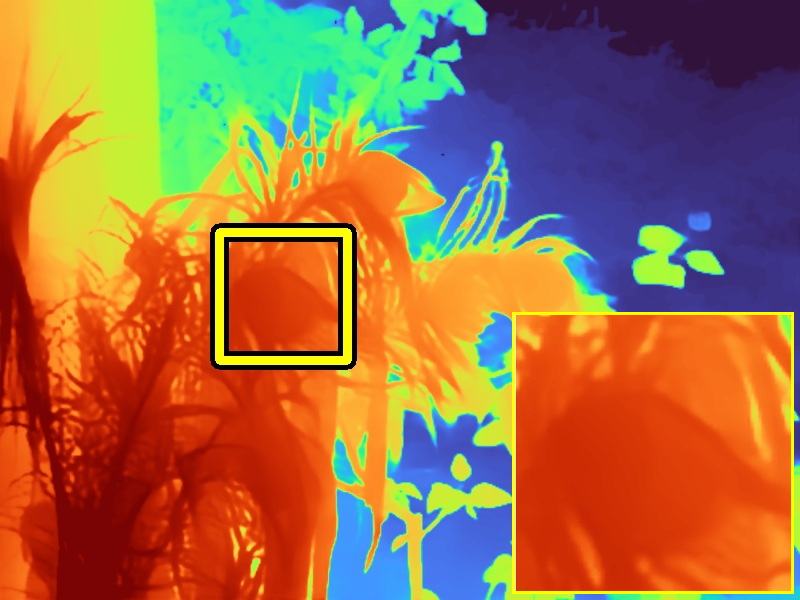} & \includegraphics[width=0.244\linewidth]{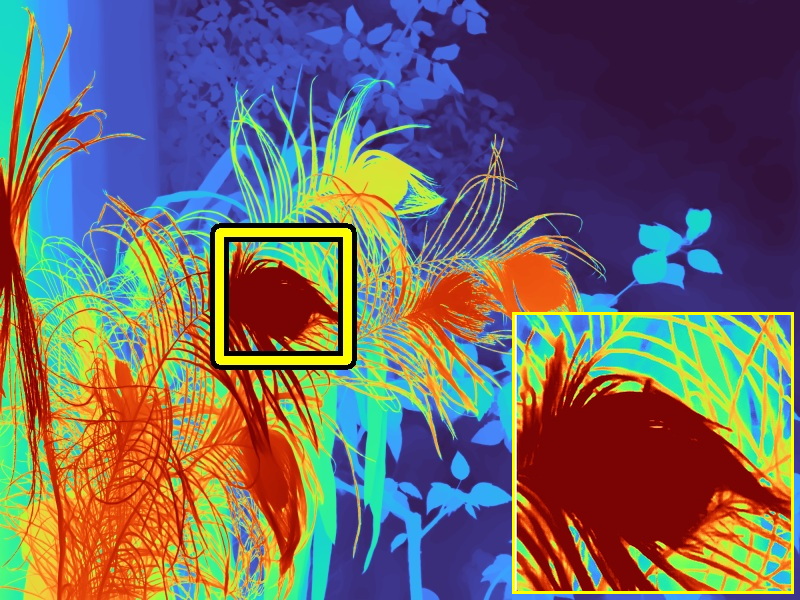} &  \\
  \end{tabular}

  \caption{\textbf{Depth-guided bokeh rendering on \realbokeh.} Each scene shows the rendered result followed by predicted disparity. Every panel contains a $4\times$ enlargement of its boxed region.}
  \label{fig:bokeh-rendering}
\end{figure}

\begin{figure}[tbhp]
  \centering
  \setlength{\tabcolsep}{1pt}
  \renewcommand{\arraystretch}{0}
  \begin{tabular}{@{}ccccc@{}}
    & \footnotesize\sffamily $t=2.20$\,s & \footnotesize\sffamily $t=5.03$\,s & \footnotesize\sffamily $t=5.43$\,s & \footnotesize\sffamily $t=6.23$\,s \\[4pt]
    \parbox[b][0.4089\linewidth][c]{11pt}{\centering\rotatebox[origin=c]{90}{\scriptsize\sffamily Input}} & \includegraphics[width=0.23\linewidth]{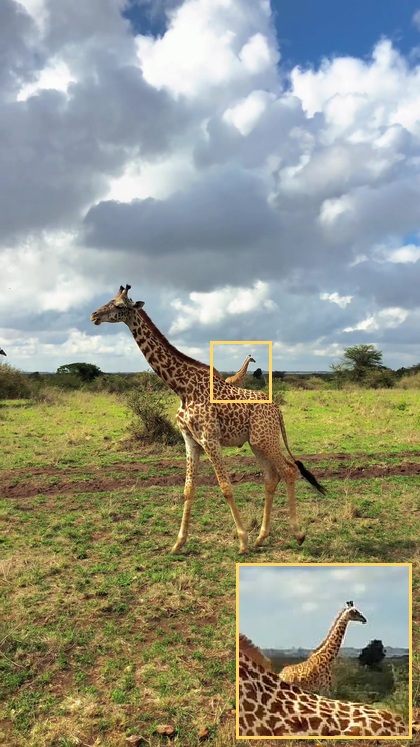} & \includegraphics[width=0.23\linewidth]{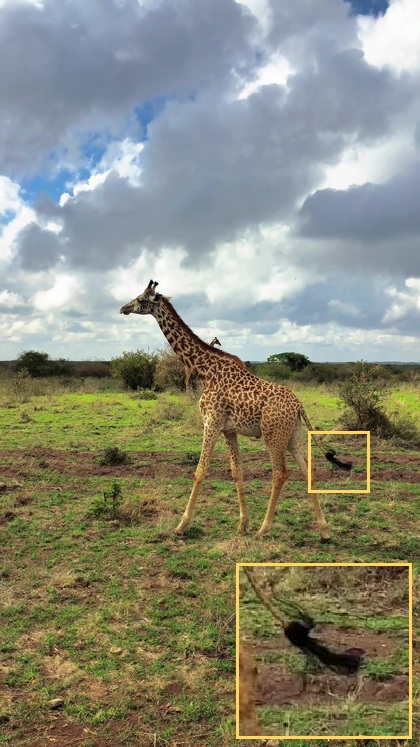} & \includegraphics[width=0.23\linewidth]{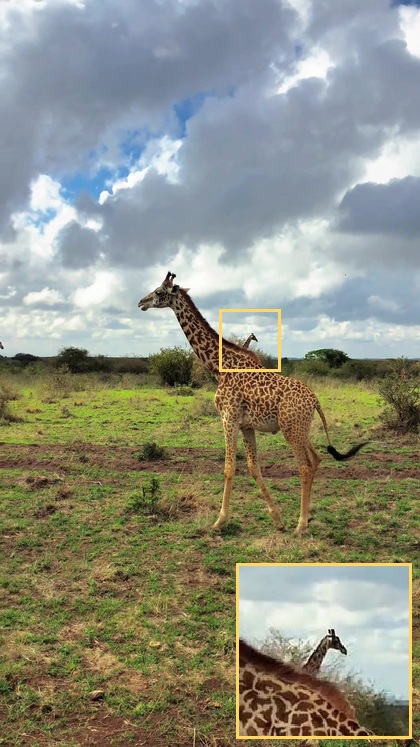} & \includegraphics[width=0.23\linewidth]{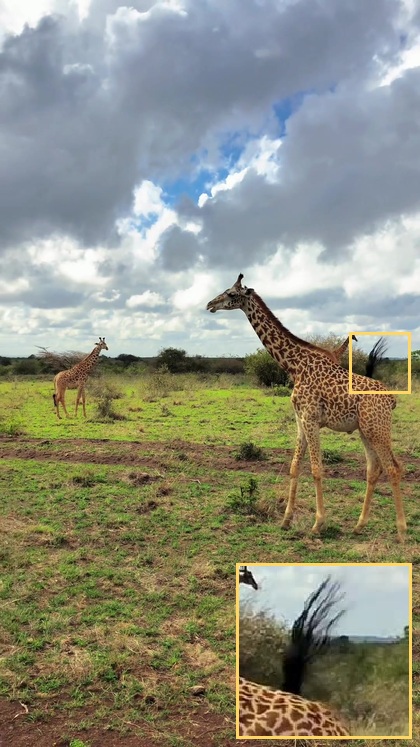} \\
    \parbox[b][0.4089\linewidth][c]{11pt}{\centering\rotatebox[origin=c]{90}{\scriptsize\sffamily \dathreeng}} & \includegraphics[width=0.23\linewidth]{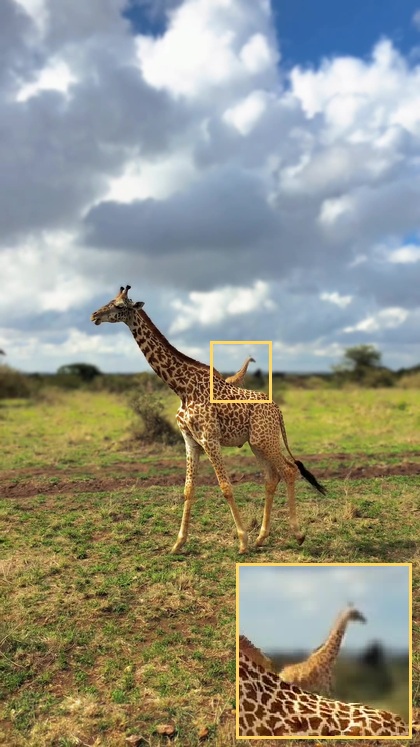} & \includegraphics[width=0.23\linewidth]{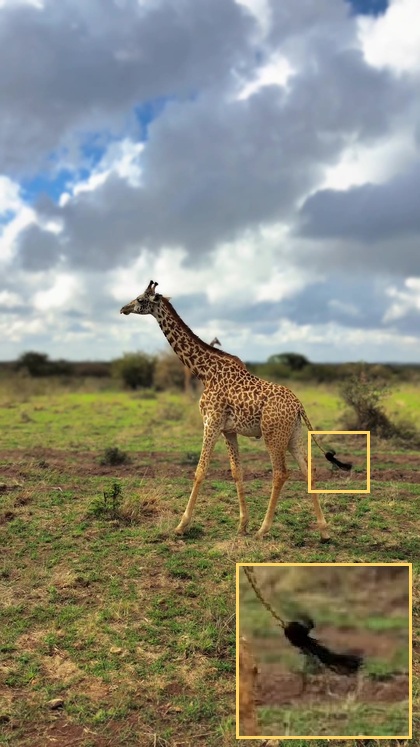} & \includegraphics[width=0.23\linewidth]{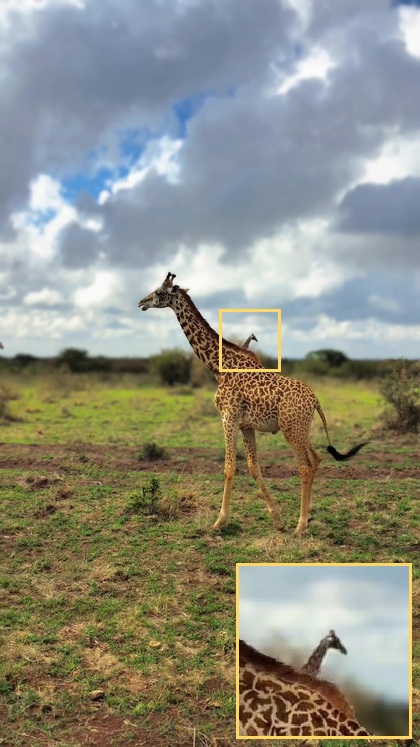} & \includegraphics[width=0.23\linewidth]{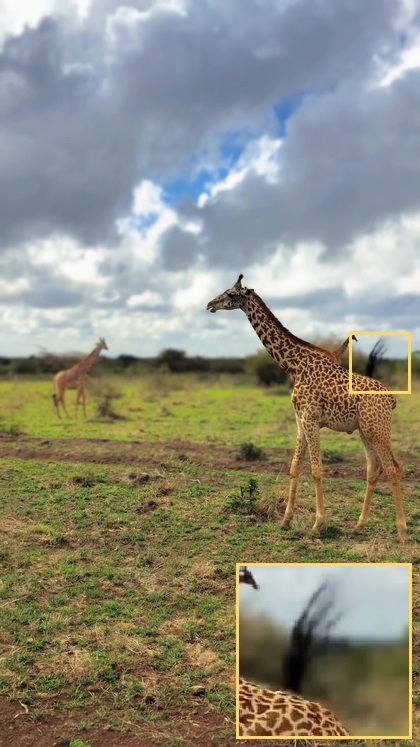} \\
    \parbox[b][0.4089\linewidth][c]{11pt}{\centering\rotatebox[origin=c]{90}{\scriptsize\sffamily \ours}} & \includegraphics[width=0.23\linewidth]{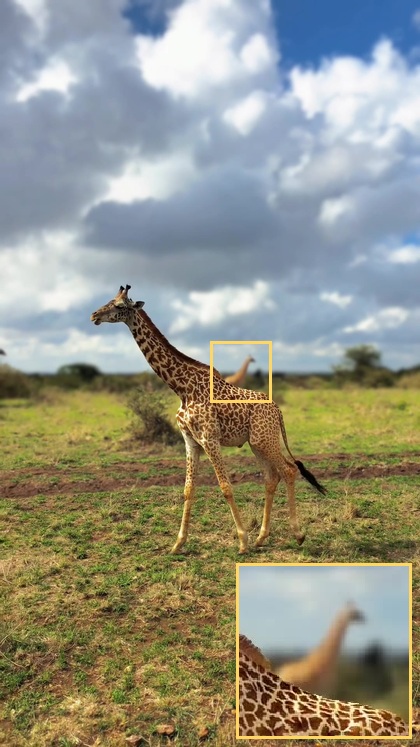} & \includegraphics[width=0.23\linewidth]{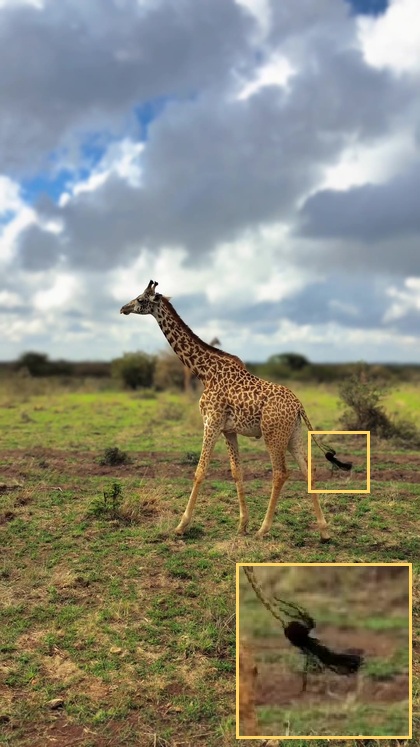} & \includegraphics[width=0.23\linewidth]{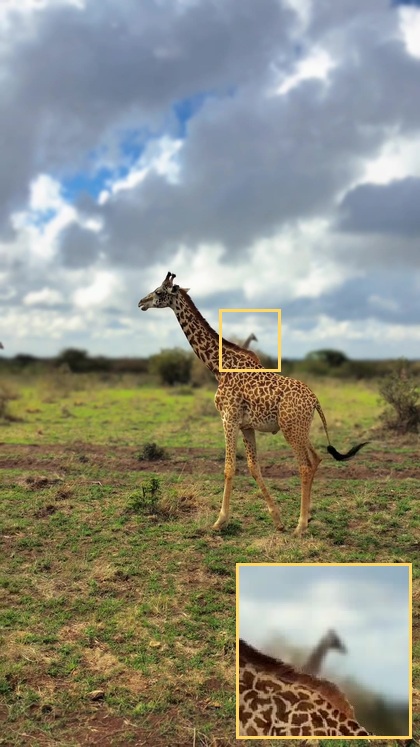} & \includegraphics[width=0.23\linewidth]{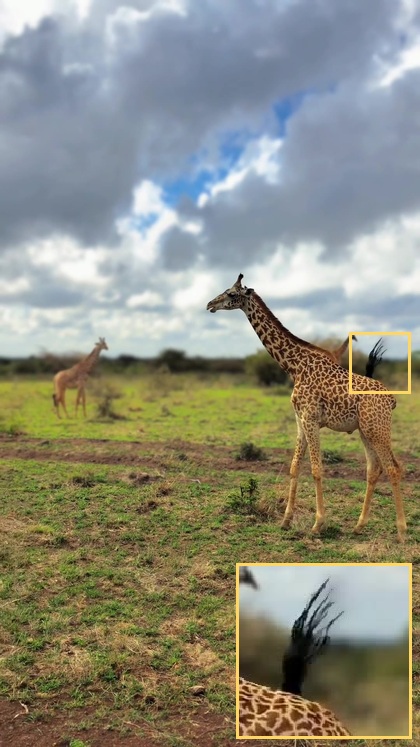} \\
  \end{tabular}

    \caption{\textbf{Depth-guided bokeh on video.} Rows show the input and the outputs rendered from each depth estimator; columns show four time steps of one handheld clip. Each panel includes a $4\times$ nearest-neighbor close-up of the boxed region.}
  \label{fig:bokeh-video-giraffe}
\end{figure}

\mypara{Evaluation protocol.} We evaluate depth-guided bokeh rendering, where depth errors near object boundaries produce visible foreground--background blending artifacts. We hold the renderer fixed and vary only its input depth, thereby evaluating the downstream utility of each estimator rather than comparing end-to-end bokeh-generation systems.

We use the 220 test scenes from \realbokeh~\citep{seizinger2025bokehlicious}, each containing an $f/22$ source image and a corresponding real $f/2$ photograph. For each source image, we estimate metric depth and provide inverse metric depth to a frozen \bokehme renderer~\citep{Peng2022BokehMe}. For \ours{}, we convert focal-normalized disparity using its predicted focal length; baselines are converted according to their native metric representations. We apply no depth renormalization or per-image alignment. Non-metric methods are excluded because their missing scale and shift cannot be recovered without scene-specific fitting. We also exclude \metricanything{}, which requires a supplied focal length to produce metric depth, whereas no camera intrinsics are provided to any method in this study.

All images are rendered and evaluated at $2000\timess1500$ resolution. The renderer and all its parameters are fixed across methods, including one global disparity multiplier selected on a disjoint validation set of 60 scenes. No parameters are fitted on the test set.

\mypara{Boundary LPIPS.} In addition to full-image PSNR, SSIM, LPIPS~\citep{Zhang2018:CVPR}, and DISTS~\citep{Ding2022:TPAMI}, we measure LPIPS within a 16-pixel band around the capture's defocus boundaries. This band is derived solely from the $f/22$ source and $f/2$ target images and is therefore identical across methods. We first exposure-align the target luma to the source using a global affine least-squares fit. For each image, we compute local high-frequency energy as $E=\mathrm{box}_{15}((\nabla^2\ell)^2)$. Pixels whose source energy exceeds its 60th percentile are classified as in focus when $\rho=E_{\mathrm{tgt}}/(E_{\mathrm{tgt}}+E_{\mathrm{src}})\geq0.25$. We close the resulting mask with a 9-pixel elliptical element, retain its largest 8-connected component, and fill its holes. Boundary LPIPS is the mean spatial LPIPS within eight pixels of the resulting boundary on either side, obtained using an exact Euclidean distance transform.

\mypara{Results.}
\oursh{} performs best across all five metrics. Relative to \dathreeng{} at $504\times378$, the strongest baseline on four of the five metrics, it improves PSNR by $0.50$\,dB and reduces global and Boundary LPIPS by $4.9\%$ and $5.6\%$, respectively. Relative to \depthpro{}, it improves PSNR by $0.75$\,dB and reduces the two LPIPS measures by $8.8\%$ and $10.1\%$. These results indicate that the metric depth and camera calibration produced by \ours{} provide more effective guidance near defocus boundaries. This single-image benchmark does not assess temporal consistency.

\mypara{Video.} We additionally apply the frozen renderer frame by frame to a 197-frame handheld clip from \citet{Pexels}. Figure~\ref{fig:bokeh-video-giraffe} compares \ours{} with \dathreeng{} (run at $1540$\,px resolution), qualitatively showing that \ours{} preserves sharp structures consistently over time.

\subsection{Novel-view synthesis}
\label{sec:appendix-nvs}

We qualitatively compare \ours to \dathreeg~\citep{Lin2026:ICLR}, \mogetwo~\citep{Wang2025:NeurIPS}, and   \unikthreed~\citep{Piccinelli2025:CVPR} on novel view synthesis.
As the novel view renderer, we employ Tiled Multiplane Image renderer~\citep{Khan2023:ICCV} with its released weights and the 4-plane setup.
As the renderer requires a disparity map with each image, we convert each method's prediction accordingly.
We clamp the disparities to $[0.01, 100]$, then min-max normalize it to $[0,1]$ per image.
To demonstrate each model's utility for high-resolution novel view synthesis, we selected images from DIS5K~\citep{Qin2022:ECCV}, a dataset of high-resolution photographs with thin structures that are sensitive to depth-boundary errors.
When applicable, we downscale each input image with bilinear interpolation to 4096px on the long side.
The camera uses fixed intrinsics ($f=0.58W$, with principal point at the image center).
We keep the input image, rendering parameters, and camera fixed across methods and only change the disparity predictor.

\begin{figure}[t]
  \centering
  \setlength{\tabcolsep}{2pt}
  \renewcommand{\arraystretch}{0.6}
  \resizebox{\linewidth}{!}{%
  \begin{tabular}{lcccc}
     & \footnotesize\sffamily Airplane & \footnotesize\sffamily Seal & \footnotesize\sffamily Balcony & \footnotesize\sffamily Swing \\
    \parbox[b][2.6cm][c]{0.9cm}{\centering\rotatebox{90}{\footnotesize\sffamily Input}} & \includegraphics[height=2.6cm]{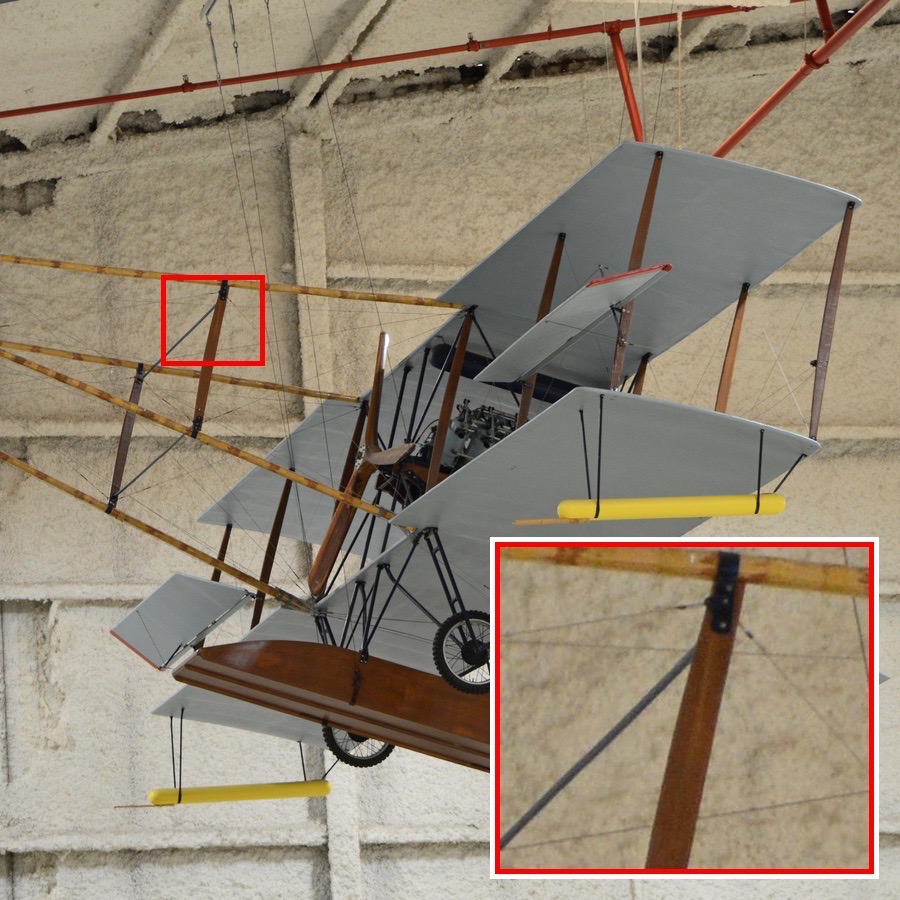} & \includegraphics[height=2.6cm]{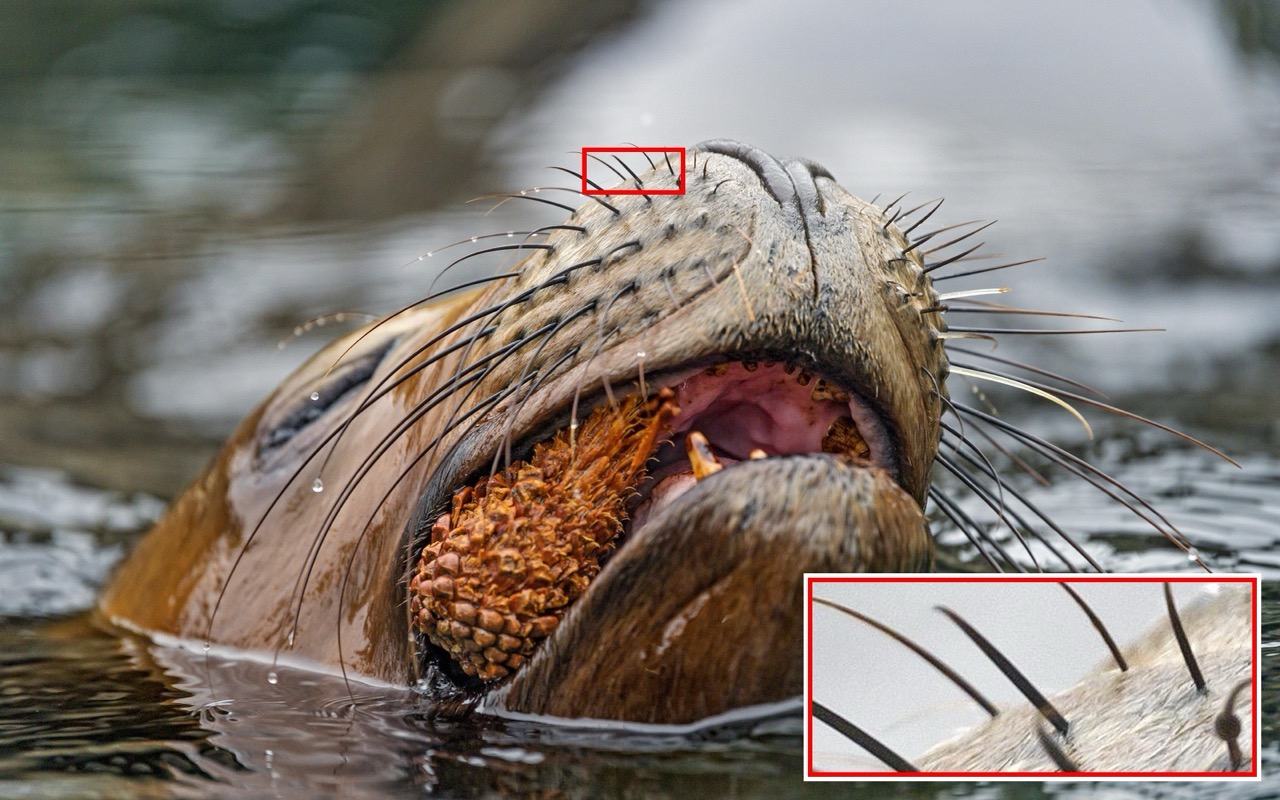} & \includegraphics[height=2.6cm]{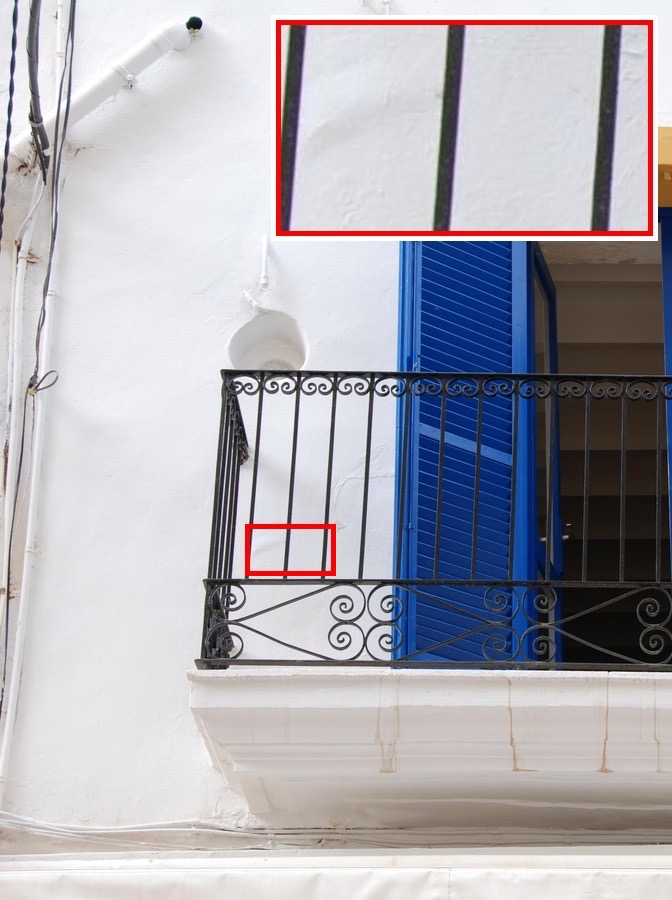} & \includegraphics[height=2.6cm]{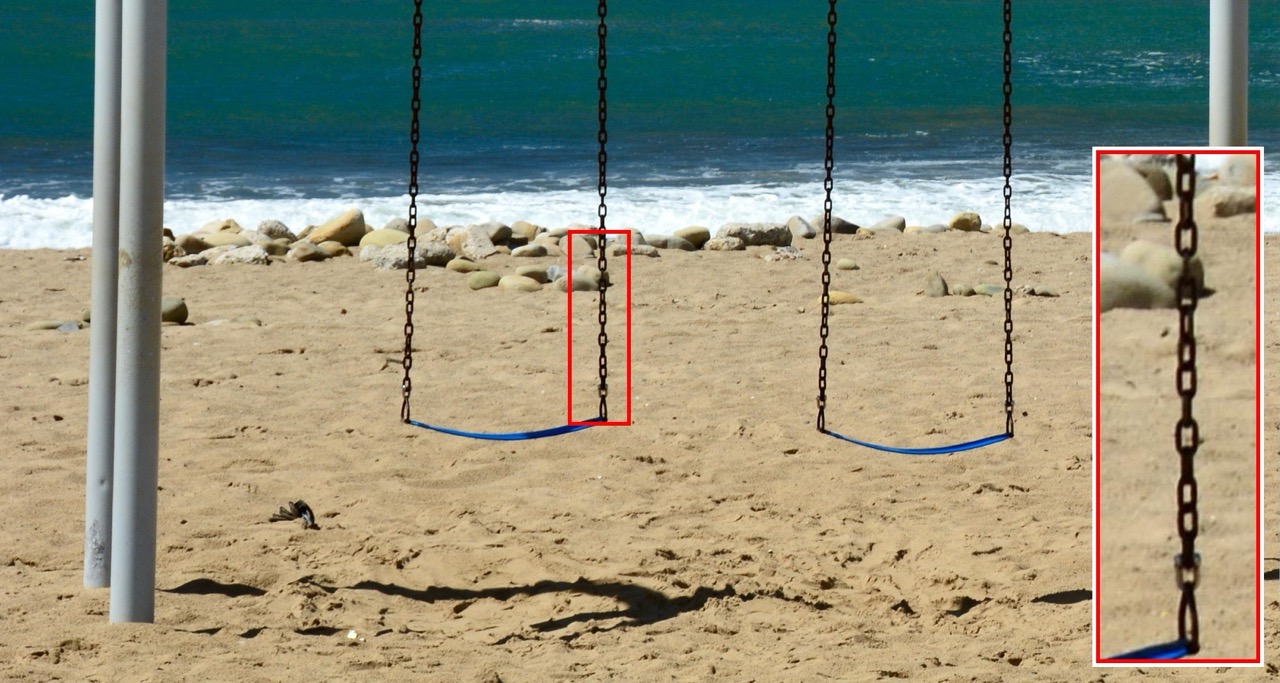} \\
    \parbox[b][2.6cm][c]{0.9cm}{\centering\rotatebox{90}{\footnotesize\sffamily \oursh}} & \includegraphics[height=2.6cm]{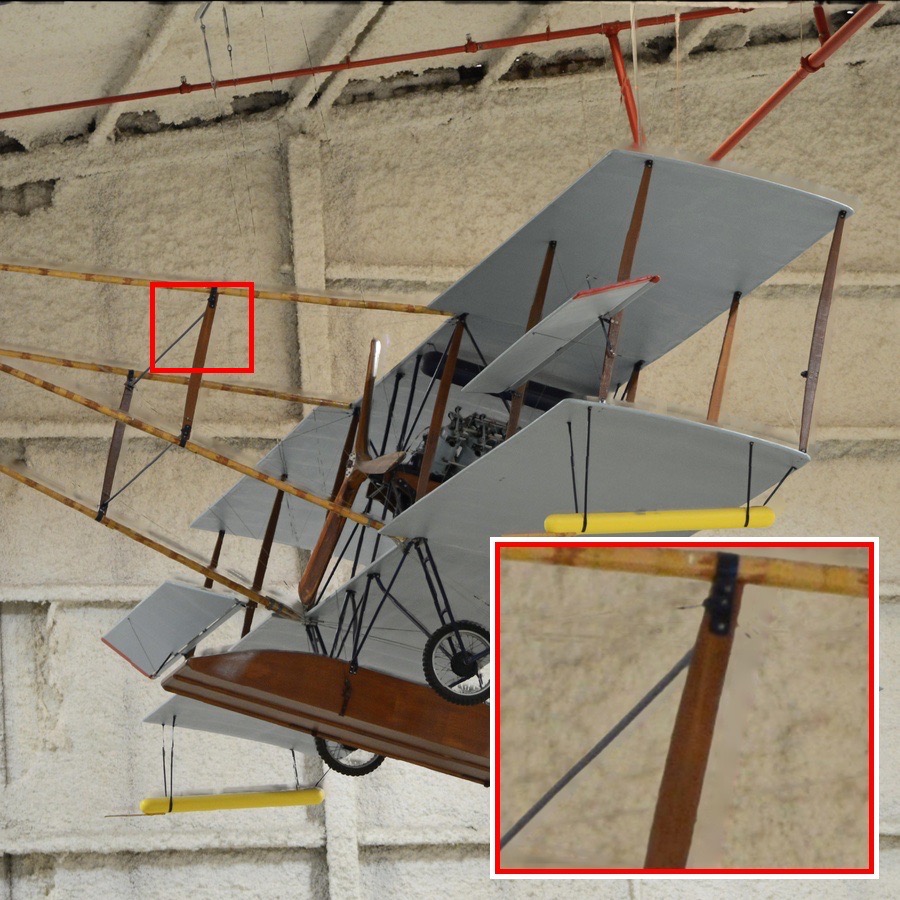} & \includegraphics[height=2.6cm]{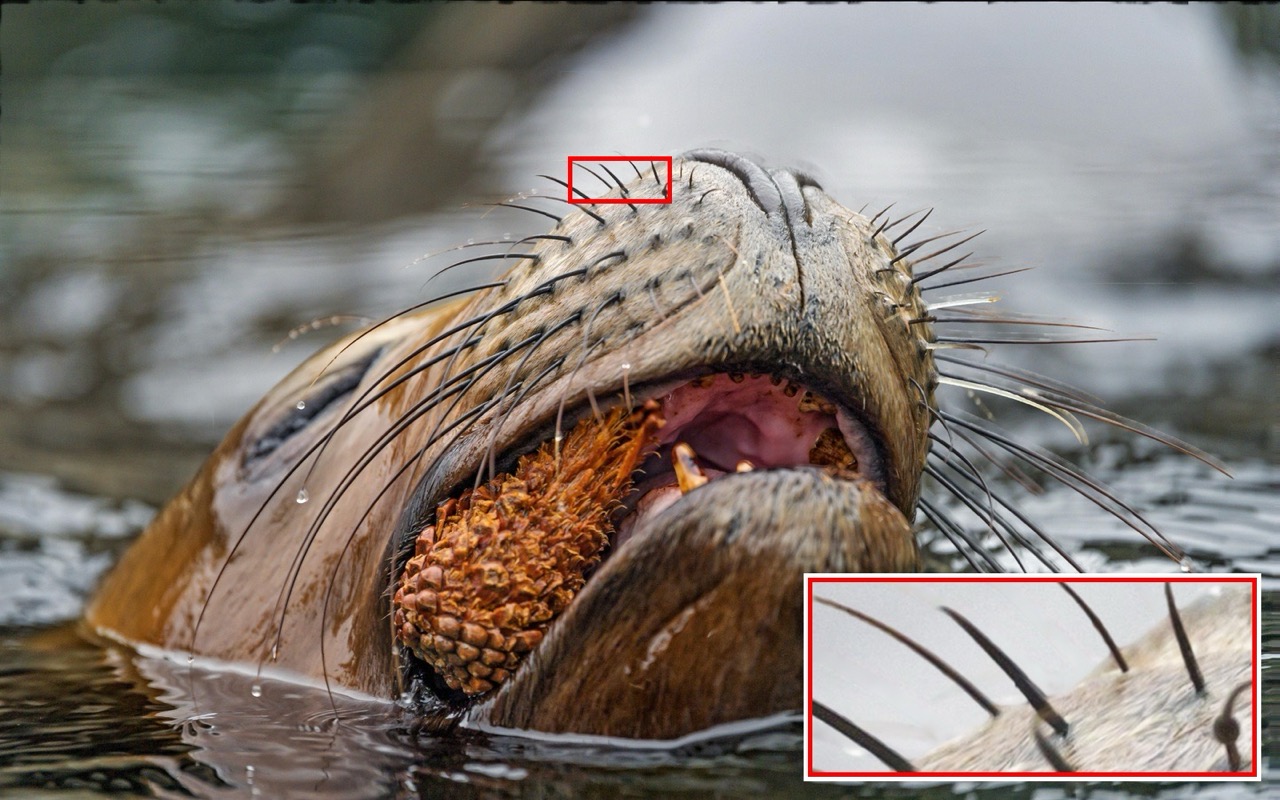} & \includegraphics[height=2.6cm]{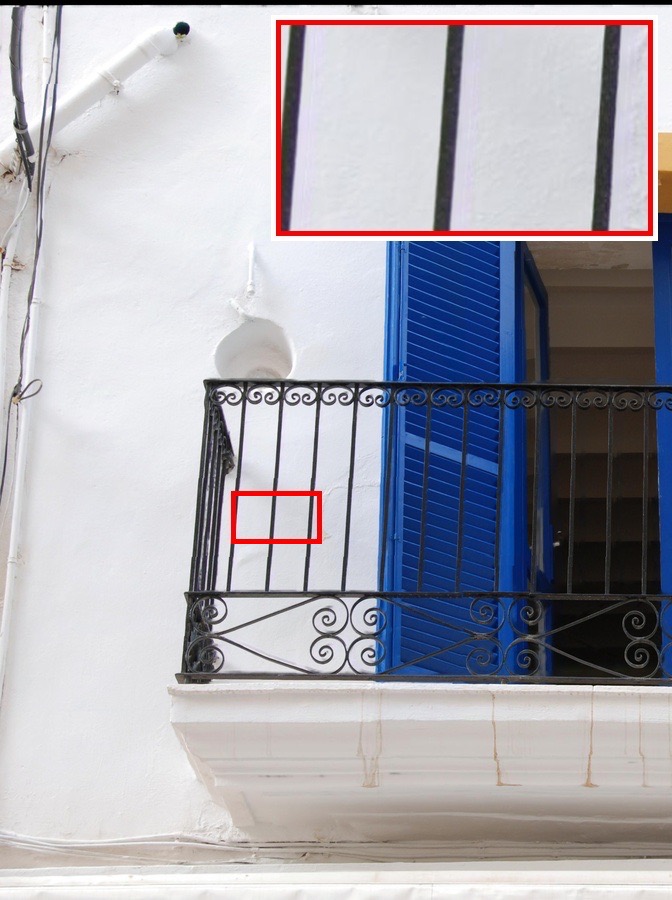} & \includegraphics[height=2.6cm]{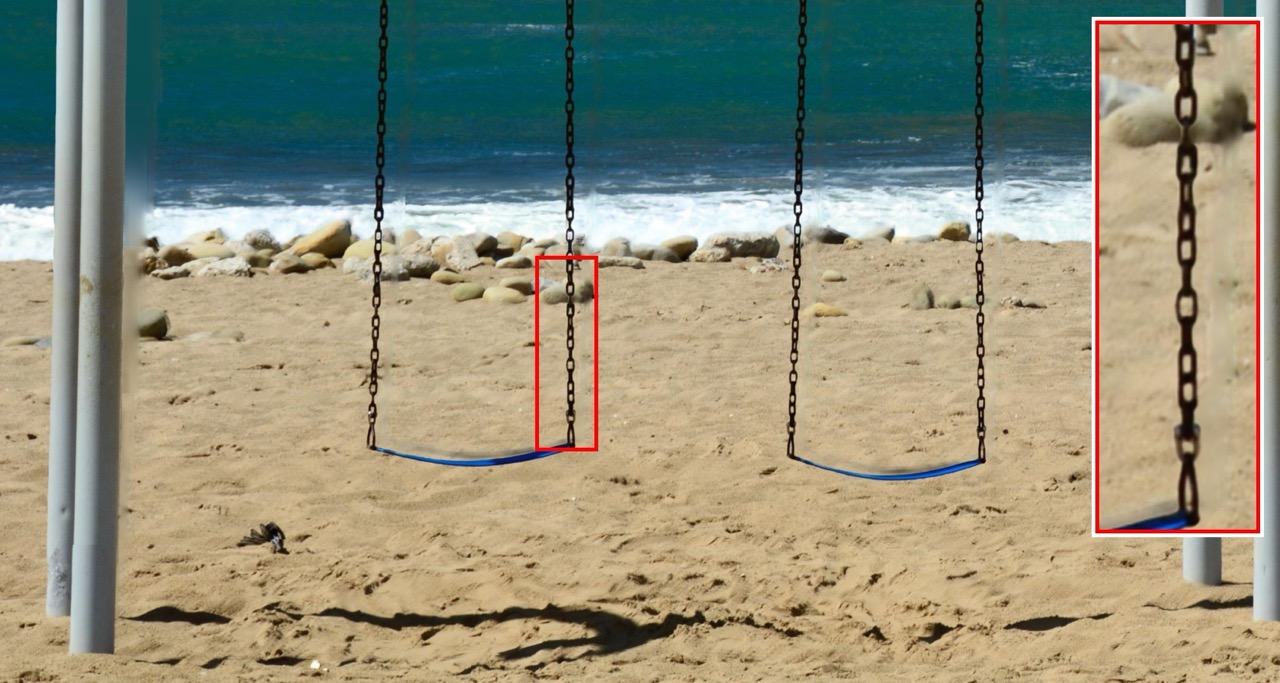} \\
    \parbox[b][2.6cm][c]{0.9cm}{\centering\rotatebox{90}{\footnotesize\sffamily \dathreeg}} & \includegraphics[height=2.6cm]{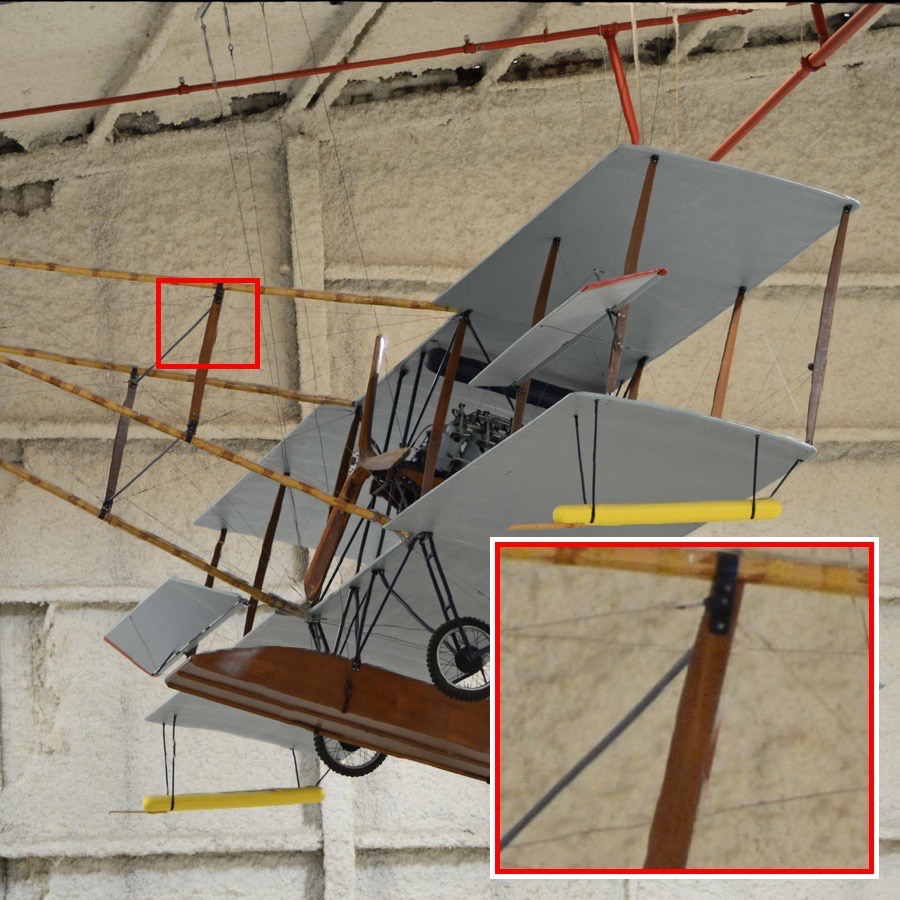} & \includegraphics[height=2.6cm]{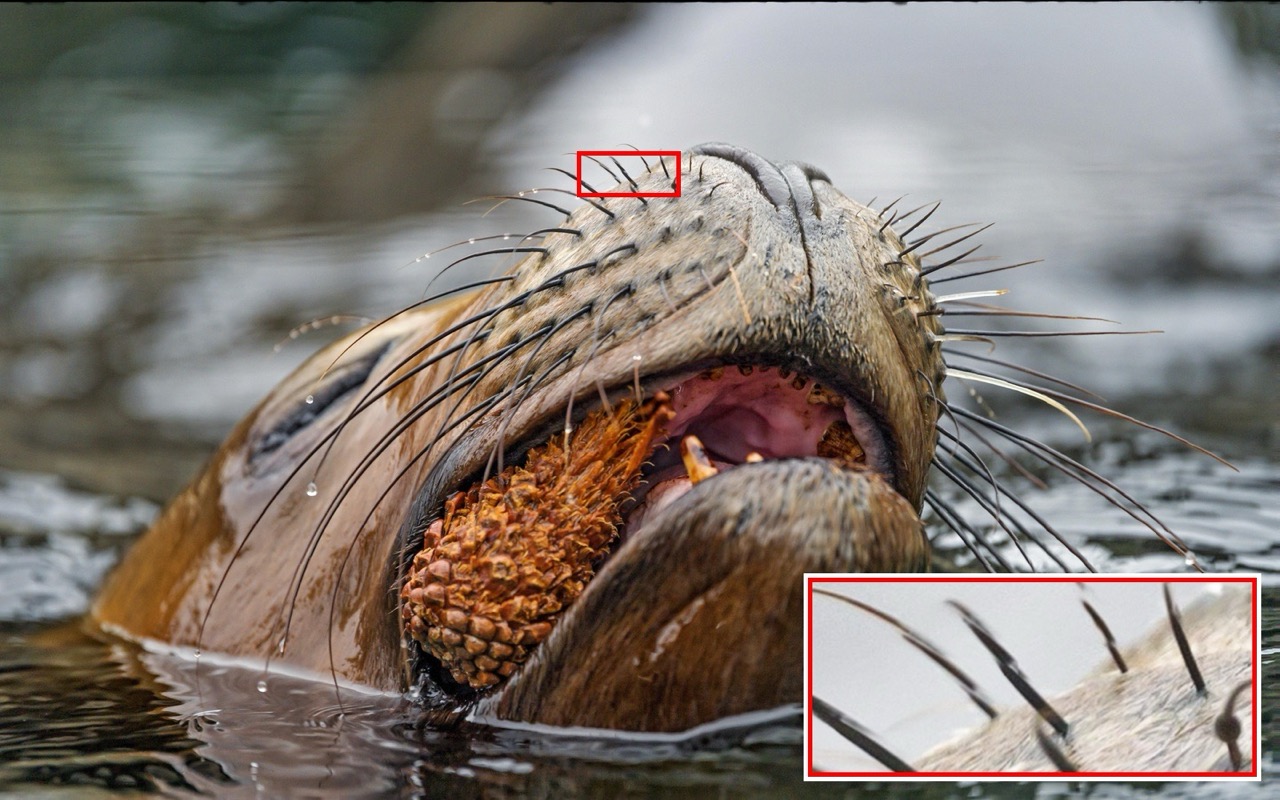} & \includegraphics[height=2.6cm]{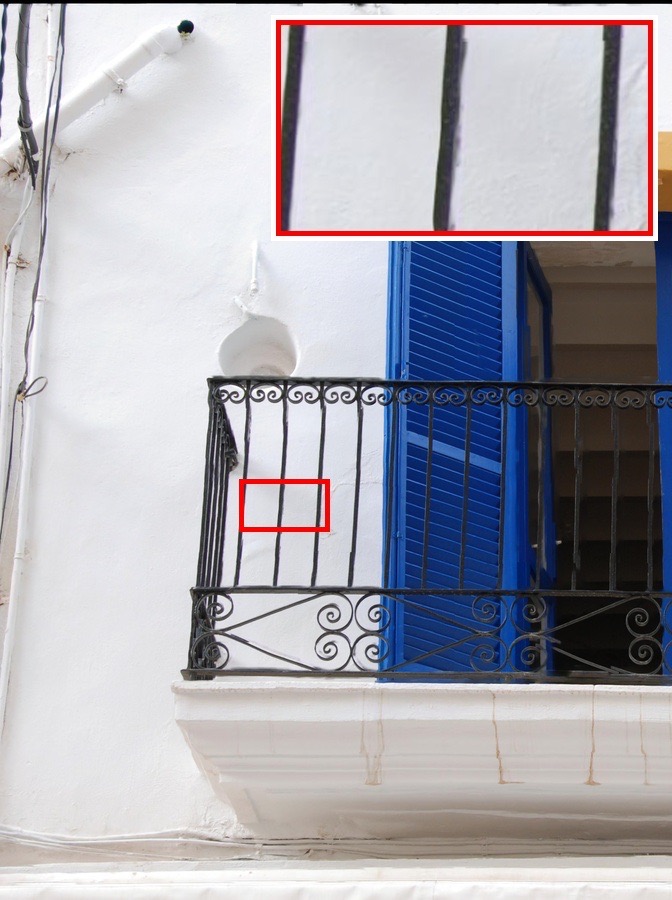} & \includegraphics[height=2.6cm]{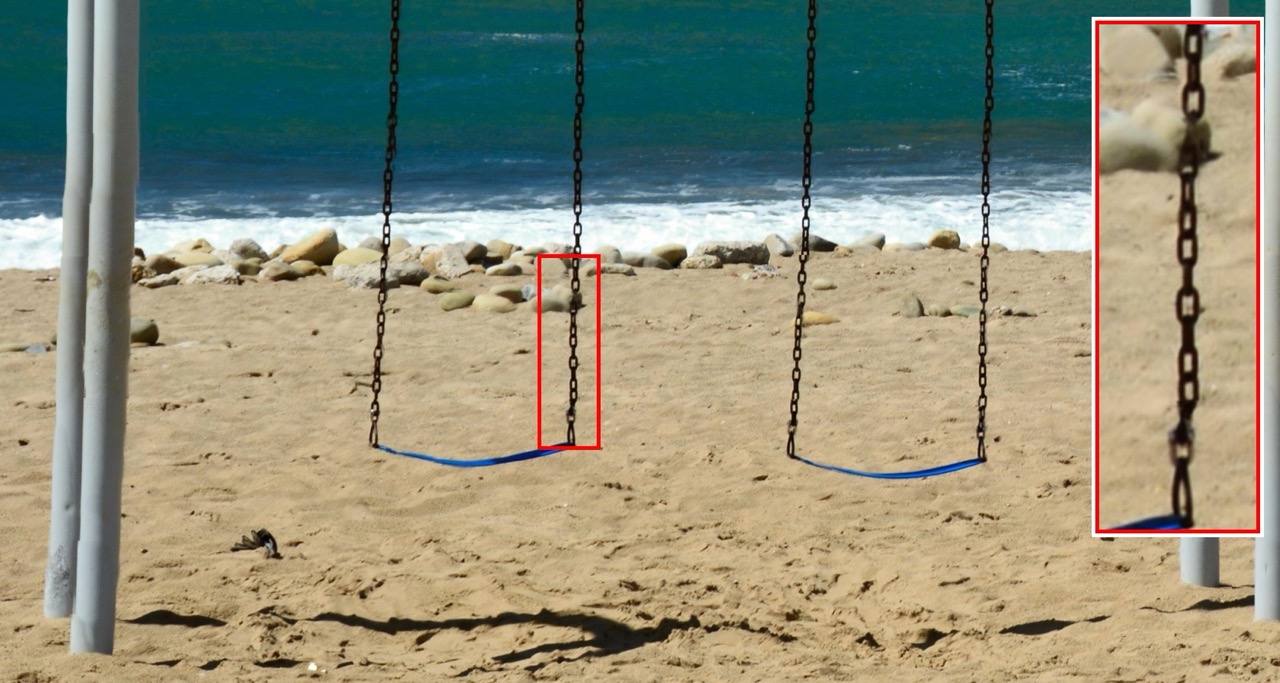} \\
    \parbox[b][2.6cm][c]{0.9cm}{\centering\rotatebox{90}{\footnotesize\sffamily \mogetwo}} & \includegraphics[height=2.6cm]{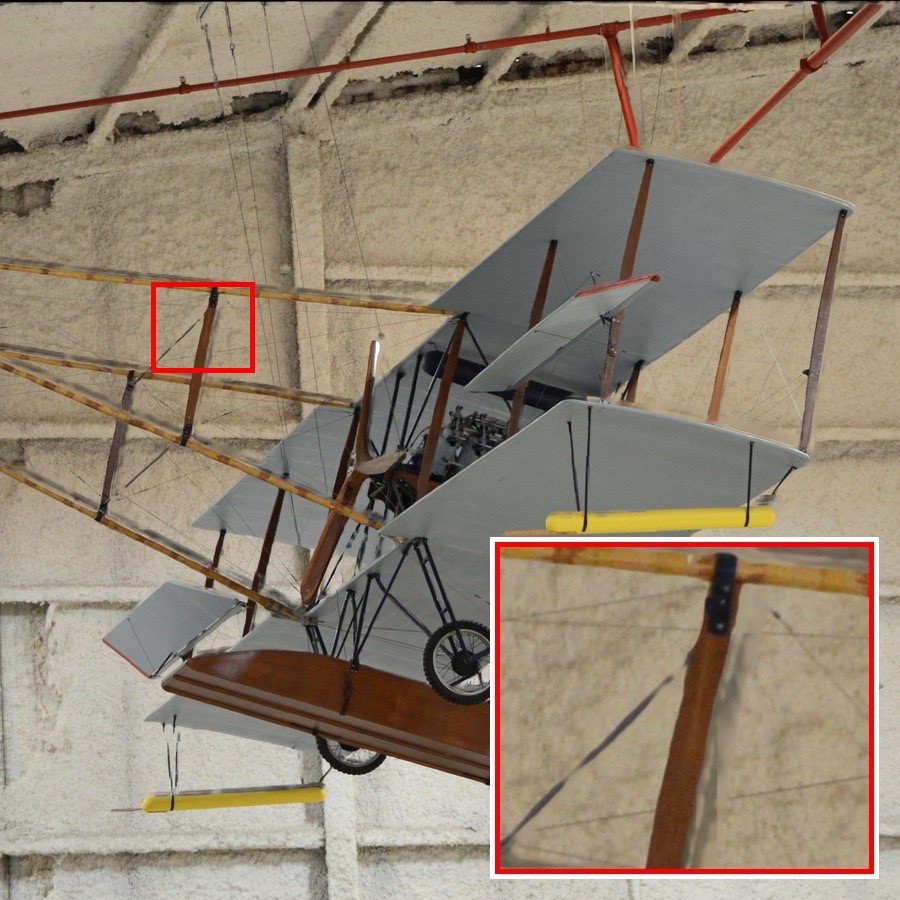} & \includegraphics[height=2.6cm]{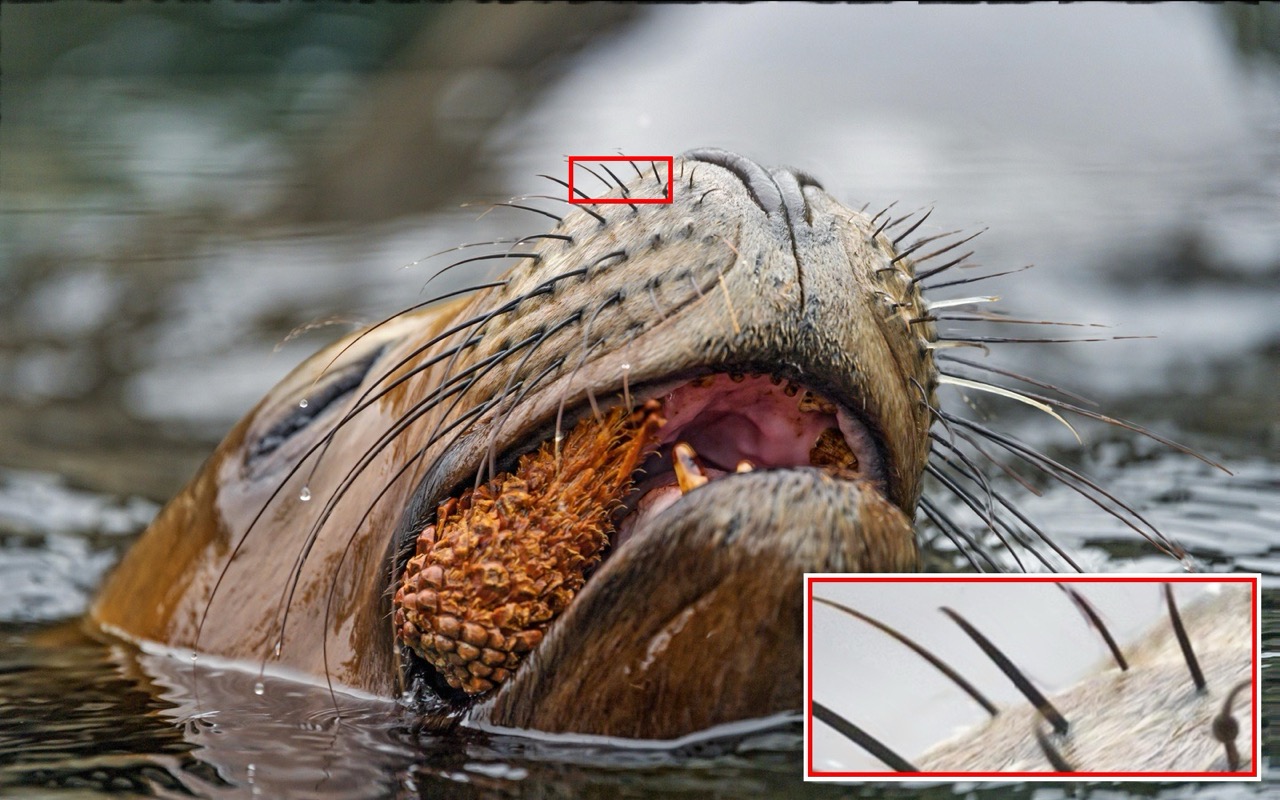} & \includegraphics[height=2.6cm]{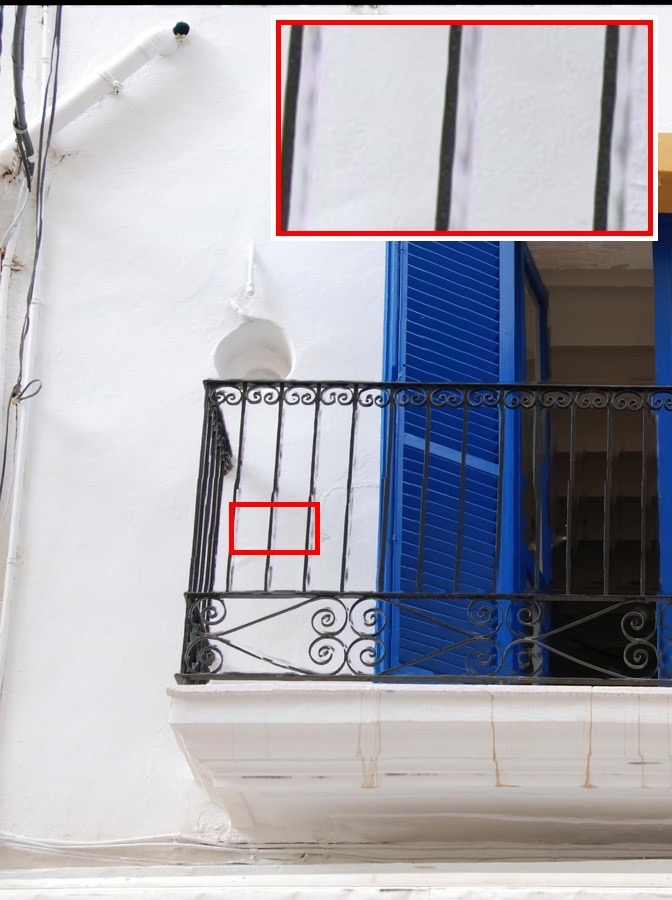} & \includegraphics[height=2.6cm]{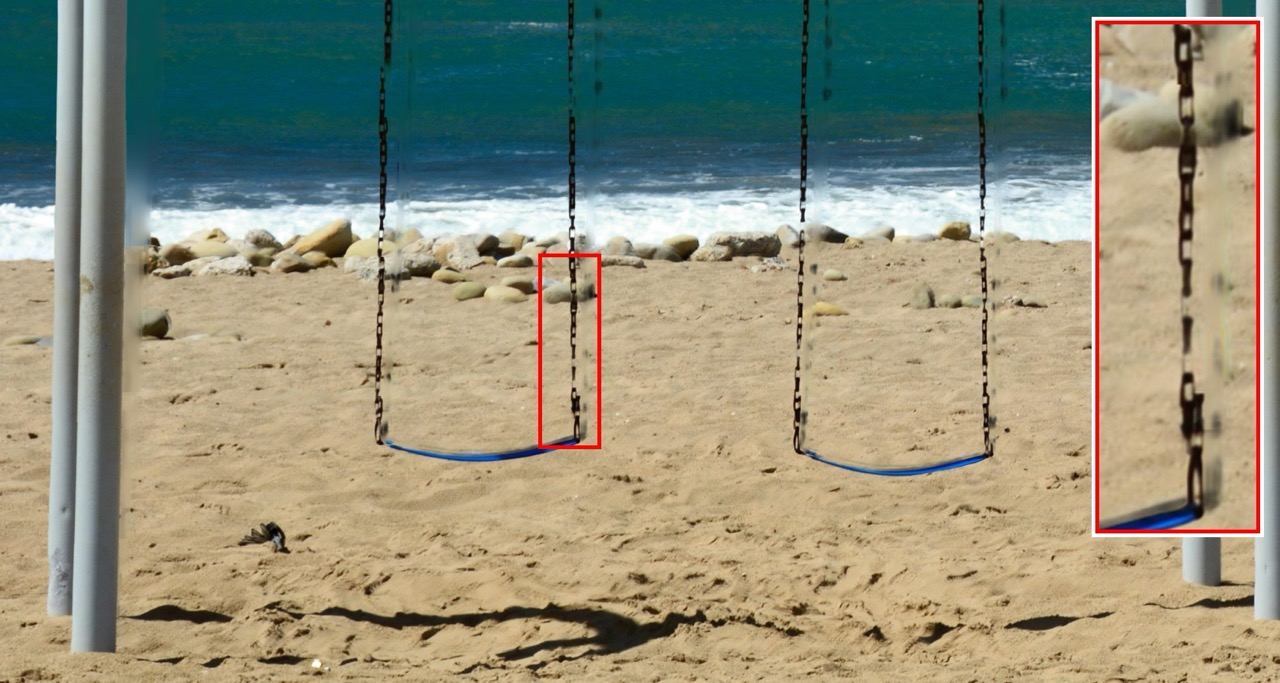} \\
    \parbox[b][2.6cm][c]{0.9cm}{\centering\rotatebox{90}{\footnotesize\sffamily \unikthreed}} & \includegraphics[height=2.6cm]{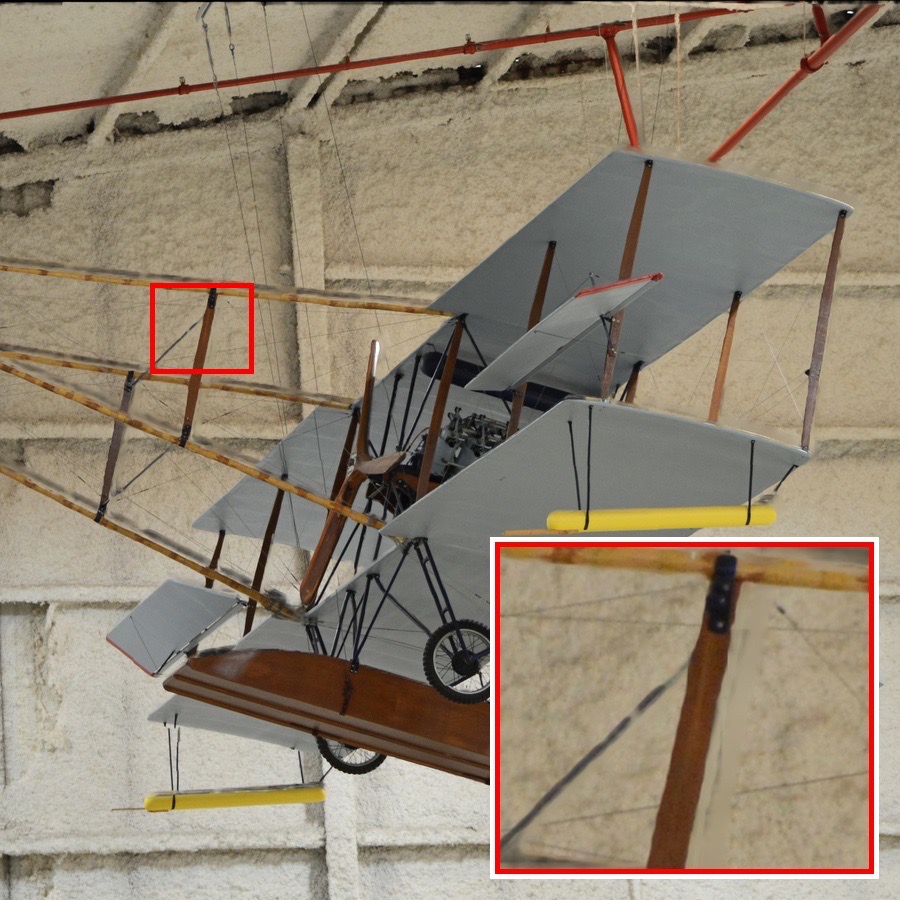} & \includegraphics[height=2.6cm]{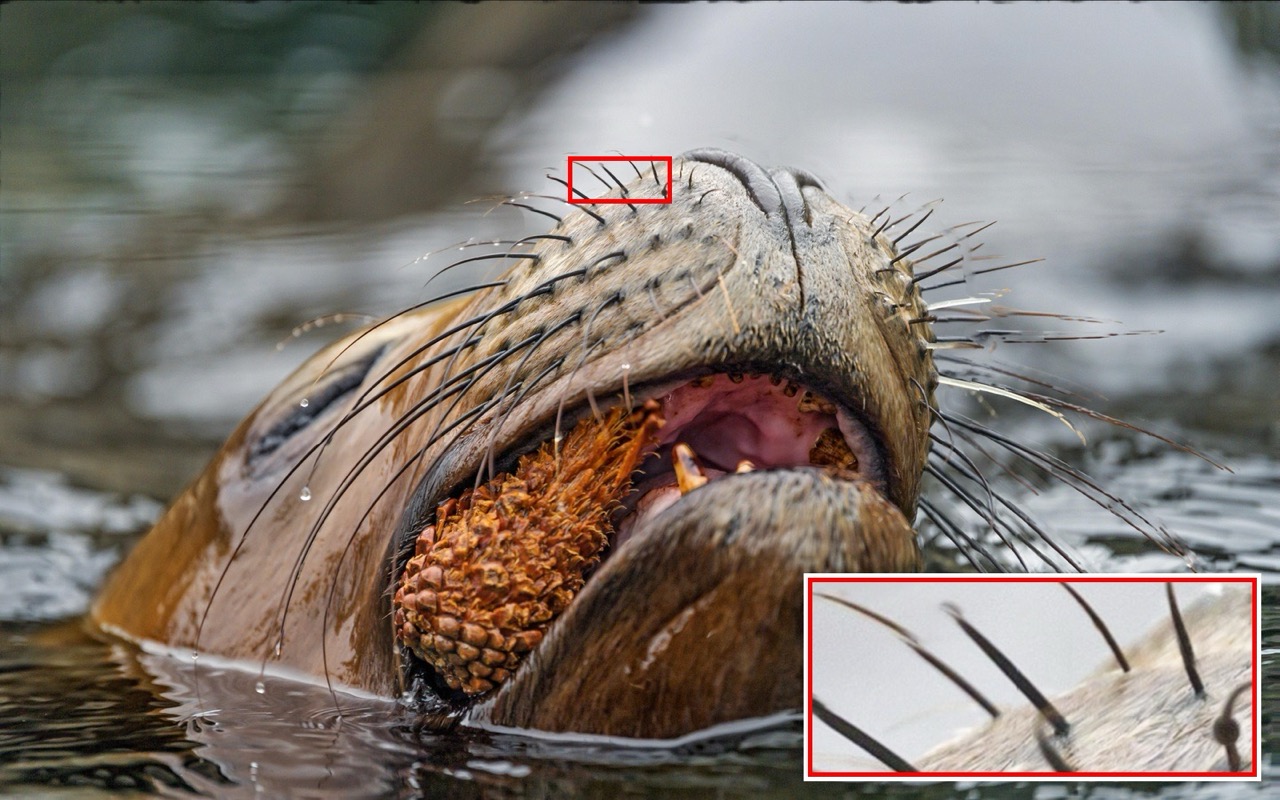} & \includegraphics[height=2.6cm]{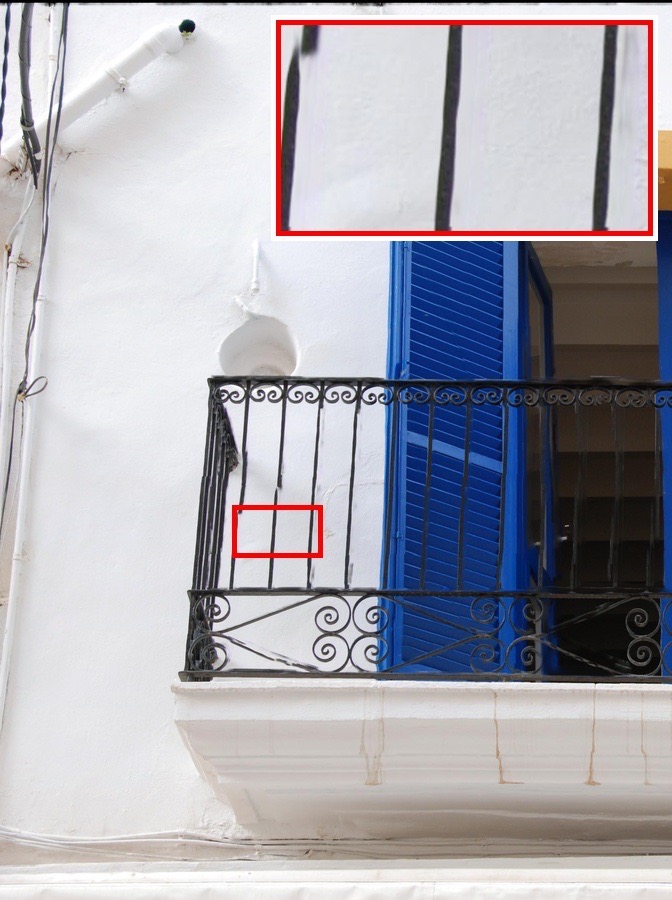} & \includegraphics[height=2.6cm]{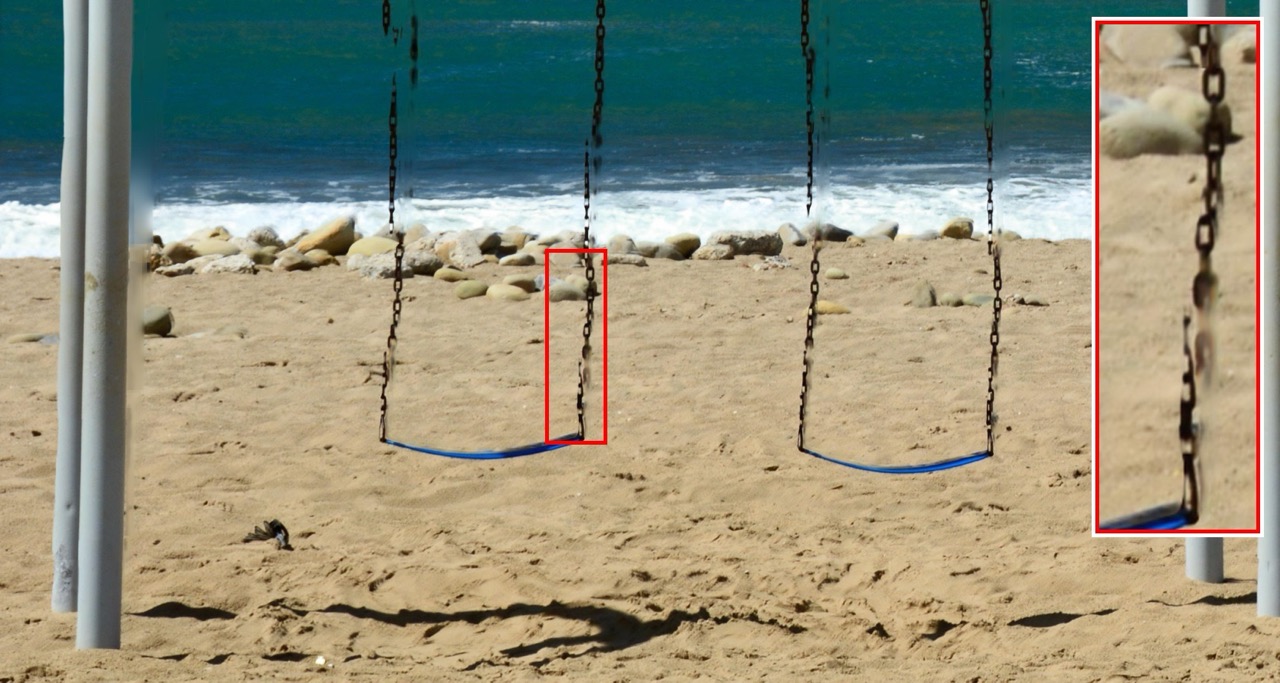} \\
  \end{tabular}%
  }
  \caption{%
    \textbf{High-resolution novel-view synthesis.} Top to bottom: input images, results from
    \oursh (2nd row), and results from representative baselines:
    \dathreeg~\citep{Lin2026:ICLR}, \mogetwo~\citep{Wang2025:NeurIPS}, and
    \unikthreed~\citep{Piccinelli2025:CVPR}.
    Depth is estimated for four high-resolution scenes from DIS5K~\citep{Qin2022:ECCV}, and every method's novel view is rendered with the same Tiled Multiplane Image renderer~\citep{Khan2023:ICCV}, at up to $4096$\,px on the long side.
    \dathree, \mogetwo, and \unikthreed break or distort thin structures. \oursh renders sharp details.
  }
  \label{fig:nvs-comparison}
\end{figure}

\mypara{Qualitative comparison.}
As shown in Figure~\ref{fig:nvs-comparison}, the baselines render airplane struts, whiskers, railings, and chains as jagged, disconnected, or missing. \oursh{} preserves these structures.%

\clearpage

\section{Controlled experiments}
\label{sec:appendix-controlled}

\subsection{Design choices}
\label{sec:appendix-hyperparameters}

\mypara{Setup.} Each condition changes the baseline training recipe along one axis. We evaluate every condition on the held-out validation sequences of Appendix~\ref{sec:appendix-validation}: Waymo and ARKitScenes for depth accuracy, and DynamicReplica for boundary sharpness and temporal consistency. Only condition D1 trains on ARKitScenes; for every other condition, ARKitScenes measures generalization to an unseen dataset. The baseline (A1) is a ViT-L model trained for 100K steps on 8 GPUs. We train A1 with 3 random seeds and report the mean and standard deviation; every other condition is trained once. Table~\ref{tab:paper-ablations} lists all results.

\begin{table}[tbh]
    \centering\ra{1.05}\footnotesize\setlength{\tabcolsep}{3.5pt}
    \caption{\textbf{Controlled experiments.}
    Results on held-out validation sequences from Waymo~\citep{Sun2020:CVPR}, ARKitScenes~\citep{Dehghan2021:NeurIPS}, and DynamicReplica~\citep{Karaev2023:CVPR}.
    The baseline A1 is trained for 100K steps; its entries are the mean $\pm$ standard deviation $\sigma$ over 3 random seeds. Differences from A1 larger than $1\sigma$ are tinted green (better) or red (worse).
    All models run at native resolution, except E3, whose input we resize to a short side of 1536 pixels to match its training.
    $\delta_1$ and F1 are in \%, and jitter is scaled by $10^2$.
    }
    \label{tab:paper-ablations}
    \fitwidth{\begin{tabular}{@{}clcccccc@{}}
  \toprule
  & & \multicolumn{2}{c}{Metric $\delta_1$ (\%) $\uparrow$} & \multicolumn{2}{c}{Relative $\delta_1$ (\%) $\uparrow$} & \multicolumn{1}{c}{Boundary F1 (\%) $\uparrow$} & \multicolumn{1}{c}{Jitter $\downarrow$} \\
  \cmidrule(lr){3-4}\cmidrule(lr){5-6}
  & & \multicolumn{1}{c}{Waymo} & \multicolumn{1}{c}{ARKitScenes} & \multicolumn{1}{c}{Waymo} & \multicolumn{1}{c}{ARKitScenes} & \multicolumn{2}{c}{DynamicReplica} \\
  \midrule
  \multicolumn{8}{@{}l}{\textbf{100k steps}} \\
  A1 & Baseline: \oursl, 8 GPUs & 91.1\,$\scriptstyle\pm0.6$ & 77.0\,$\scriptstyle\pm3.8$ & 92.8\,$\scriptstyle\pm0.1$ & 97.9\,$\scriptstyle\pm0.1$ & 50.2\,$\scriptstyle\pm0.1$ & 0.36\,$\scriptstyle\pm0.01$ \\
  \addlinespace[2pt]
   & \multicolumn{7}{@{}l}{\textit{Model size and compute}} \\
  B1 & \quad \ourss, 128 GPUs & \worseC{$-$7.4} & \worseC{$-$11.9} & \worseC{$-$3.0} & \worseC{$-$3.9} & \worseC{$-$11.7} & \worseC{$+$0.09} \\
  B2 & \quad \oursb, 128 GPUs & \betterA{$+$0.8} & \worseB{$-$8.1} & \worseB{$-$0.1} & \worseB{$-$1.5} & \worseB{$-$0.1} & \betterA{$-$0.04} \\
  B3 & \quad \oursl, 128 GPUs & \betterB{$+$2.8} & $-$0.4 & \betterB{$+$1.7} & \betterB{$+$0.8} & \betterB{$+$5.8} & \betterB{$-$0.11} \\
  B4 & \quad \oursh, 128 GPUs & \betterC{$+$4.3} & \betterC{$+$15.4} & \betterC{$+$3.0} & \betterC{$+$1.3} & \betterC{$+$7.5} & \betterC{$-$0.12} \\
  \addlinespace[2pt]
   & \multicolumn{7}{@{}l}{\textit{Training data}} \\
  D1 & \quad additional real-world data & $+$0.2 & \betterC{$+$14.3} & \betterB{$+$0.5} & \betterC{$+$0.7} & \worseB{$-$1.2} & \worseB{$+$0.03} \\
  D2 & \quad synthetic data only & \worseB{$-$5.5} & \worseB{$-$12.3} & \worseB{$-$3.0} & \worseB{$-$0.5} & \betterB{$+$0.4} & $-$0.01 \\
  D3 & \quad internal data only & \worseC{$-$16.8} & \worseA{$-$7.2} & \worseC{$-$7.0} & \worseA{$-$0.4} & \worseC{$-$4.5} & \worseC{$+$0.07} \\
  D4 & \quad external data only & \betterC{$+$3.2} & \worseC{$-$45.2} & \betterC{$+$2.0} & \worseC{$-$0.8} & \betterC{$+$4.7} & \betterC{$-$0.07} \\
  \addlinespace[2pt]
   & \multicolumn{7}{@{}l}{\textit{Training crops}} \\
  E1 & \quad any aspect ratio, stride 1 & $+$0.4 & $+$0.6 & \betterC{$+$0.2} & 0.0 & \worseA{$-$0.4} & $+$0.01 \\
  E2 & \quad square, native scale & \worseB{$-$46.6} & \worseC{$-$16.2} & \worseB{$-$1.1} & \worseC{$-$0.2} & \worseB{$-$2.1} & \worseC{$+$0.06} \\
  E3 & \quad square, resized to $1536^2$ & \worseC{$-$48.5} & \worseB{$-$6.3} & \worseC{$-$1.4} & $+$0.1 & \worseC{$-$5.7} & \worseB{$+$0.01} \\
  \addlinespace[2pt]
   & \multicolumn{7}{@{}l}{\textit{Backbone}} \\
  H1 & \quad \dinovtwo{} & $-$0.2 & \worseB{$-$12.8} & $+$0.1 & \worseB{$-$0.4} & \worseB{$-$0.5} & \betterC{$-$0.01} \\
  H2 & \quad frozen \dinovthree{} & \worseC{$-$13.1} & \worseC{$-$29.4} & \worseC{$-$4.4} & \worseC{$-$6.7} & \worseC{$-$7.0} & \worseC{$+$0.33} \\
  \addlinespace[2pt]
   & \multicolumn{7}{@{}l}{\textit{Training losses}} \\
  I1 & \quad multi-view consistency loss off & \betterB{$+$0.9} & $+$1.2 & \betterB{$+$0.4} & $+$0.1 & $+$0.1 & \worseB{$+$0.05} \\
  I2 & \quad \depthpro's 1st-order boundary loss & \betterA{$+$0.8} & $+$1.2 & $-$0.1 & \betterC{$+$0.2} & \worseA{$-$3.5} & \worseA{$+$0.03} \\
  I3 & \quad \depthpro's full boundary losses & \worseC{$-$3.3} & \worseC{$-$11.2} & \worseC{$-$1.5} & $-$0.1 & \worseB{$-$3.8} & \worseC{$+$0.09} \\
  I4 & \quad confidence-weighted losses & \betterC{$+$1.4} & $+$1.1 & \betterC{$+$0.4} & 0.0 & \worseC{$-$6.8} & \worseC{$+$0.07} \\
  \bottomrule
\end{tabular}
}
\end{table}

\mypara{Model and batch size.}
Increasing the number of GPUs from 8 (A1) to 128 (B3) multiplies the batch size, and thus the number of samples seen in 100K steps, by 16 and improves every metric except metric $\delta_1$ on ARKitScenes. At 128 GPUs, every metric improves monotonically with model size, from \ourss{} (B1) to \oursh{} (B4).

\mypara{Training data.} Our training data combine external datasets and internal data, and both contain synthetic and real-world parts. We vary this mix in four ways: adding real-world data (D1), removing all real-world data (D2), removing all external data (D3), and removing all internal data (D4). D1 adds Objectron~\citep{Ahmadyan2021:CVPR},  Argoverse2~\citep{Wilson2021}, MapFree~\citep{Arnold2022:ECCV}, the training sequences of ARKitScenes~\citep{Dehghan2021:NeurIPS}, WayveScenes101~\citep{Zuern2024}, TartanRGBT~\citep{Maheshwari2026}, the Aria captures (ADT~\citep{Pan2023:ICCV}, HOT3D~\citep{Banerjee2025:CVPR}, Nymeria~\citep{Ma2024:ECCV}, and Reading in the Wild~\citep{Yang2025:NeurIPS}), and additional internal real-world captures.

Adding real-world data (D1) improves depth accuracy but smooths boundaries. We attribute the smoothing to the ground truth of real-world datasets, which comes from depth sensors and is often inaccurate at occlusion boundaries. Consistent with this, removing all real-world data (D2) sharpens boundaries beyond the baseline but degrades depth accuracy. Training on internal data only (D3) degrades every metric. Removing all internal data (D4) improves accuracy on Waymo, sharpens boundaries, and lowers jitter, but lowers metric $\delta_1$ on ARKitScenes by 45.2 percentage points. Internal data thus improves generalization to ARKitScenes, the one validation dataset outside the training mix, at some cost in boundary sharpness and temporal consistency.

\mypara{Training resolution.} The baseline (A1) trains on crops of any aspect ratio at native scale and subsamples some crops with a stride of 2, which adds lower-resolution views without interpolating the ground truth. Removing the strided subsampling (E1) slightly improves accuracy but degrades boundary sharpness. Training only on square crops at native scale
(E2) preserves most of the sharpness but lowers metric $\delta_1$ on Waymo by 46.6 percentage points. Waymo frames have aspect ratios of 3:2 and about 2.2:1, which suggests that the model needs varied aspect ratios and fields of view in training to recover metric scale. Resizing the square crops to $1536\times1536$, as \depthpro{}~\citep{Bochkovskii2025:ICLR} does (E3), lowers boundary sharpness further, consistent with the benefit of training at native resolution.

\mypara{Backbone.} Replacing the pretrained DINOv3 backbone with DINOv2 ViT-L/14 (H1) slightly lowers jitter but lowers metric $\delta_1$ on ARKitScenes by 12.8 points. Freezing DINOv3 (H2) degrades every metric.

\mypara{Training losses.} Removing the multi-view consistency loss (I1) raises jitter by 0.05. Replacing our gradient loss with the first-order boundary loss of \depthpro{} (I2), which computes the same gradient error at 6 scales instead of 3, lowers boundary F1 by 3.5 percentage points. The full set of \depthpro{} boundary losses (I3), which adds second-order and squared-gradient terms to I2, lowers boundary F1 by 3.8 percentage points and lowers metric $\delta_1$ on Waymo by 3.3 points. Weighting our inverse-depth losses by predicted confidence (I4), as feed-forward reconstruction methods do, lowers boundary F1 by 6.8 points. We hypothesize that the model is least certain at boundaries, so confidence weighting lets it discount the loss exactly where it must learn to resolve them.

\subsection{Inference-time analysis}
\label{sec:appendix-diagnostic}
We study how the trained \oursh{} behaves at inference time, without retraining. Both studies use the held-out validation windows of Appendix~\ref{sec:appendix-validation}. The cross-frame attention study (Table~\ref{tab:xattn-drop}) uses DynamicReplica, which provides the ground-truth optical flow that jitter requires. The resolution study   (Table~\ref{tab:window-resolution}) uses all three datasets.

\mypara{Cross-frame attention vs. temporal consistency.}
To probe the causal relationship between the cross-frame attention mechanism and temporal consistency, we randomly drop connections between frames with a certain probability.
The dropout pattern is consistent across cross-frame attention blocks throughout a forward pass.
We repeat three times with different dropout patterns and report the mean and standard deviation.
We report metric depth accuracy, relative depth accuracy, boundary F1, and jitter in Table~\ref{tab:xattn-drop}.
As we remove more connections, jitter rises monotonically from 0.24 to 0.84.
This suggests the cross-frame attention mechanism is key to our model for temporally consistent depth estimation.
In addition, sharpness also degrades as more connections are dropped, suggesting
the model uses information from other frames to predict sharp boundaries as well.
\begin{table}[h]
  \centering\ra{1.05}\footnotesize
  \caption{\textbf{Cross-frame attention vs temporal consistency.} We evaluate \oursh{} on 32-frame clips from DynamicReplica, while randomly dropping cross-frame attention between pairs of frames. All blocks follow the same dropout pattern throughout the forward pass. We report metrics over three draws. Jitter rises steadily as more connections are removed. Accuracy holds until cross-frame attention is fully disabled. Boundary F1 declines slowly as connections are removed and drops most when cross-frame attention is fully disabled. Jitter is scaled by $10^2$. Bold: best.}
  \label{tab:xattn-drop}
  \fitwidth{\begin{tabular}{@{}lcccc@{}}
  \toprule
  Connections dropped & \multicolumn{1}{c}{Metric $\delta_1$ (\%) $\uparrow$} & \multicolumn{1}{c}{Relative $\delta_1$ (\%) $\uparrow$} & \multicolumn{1}{c}{Boundary F1 (\%) $\uparrow$} & \multicolumn{1}{c}{Jitter $\downarrow$} \\
  \midrule
  0\,\% & \textbf{98.9} & 98.9 & \textbf{57.7} & \textbf{0.24} \\
  20\,\% & \textbf{98.9\,$\scriptstyle\pm0.1$} & 98.9\,$\scriptstyle\pm0.1$ & \textbf{57.7\,$\scriptstyle\pm0.1$} & 0.28\,$\scriptstyle\pm0.01$ \\
  40\,\% & \textbf{98.9\,$\scriptstyle\pm0.1$} & 98.9\,$\scriptstyle\pm0.1$ & 57.5\,$\scriptstyle\pm0.1$ & 0.33\,$\scriptstyle\pm0.02$ \\
  60\,\% & \textbf{98.9\,$\scriptstyle\pm0.1$} & 98.9\,$\scriptstyle\pm0.1$ & 57.3\,$\scriptstyle\pm0.1$ & 0.47\,$\scriptstyle\pm0.01$ \\
  80\,\% & \textbf{98.9\,$\scriptstyle\pm0.1$} & 98.9\,$\scriptstyle\pm0.1$ & 56.7\,$\scriptstyle\pm0.1$ & 0.76\,$\scriptstyle\pm0.02$ \\
  100\,\% & 98.1 & \textbf{99.0} & 53.4 & 0.84 \\
  \bottomrule
\end{tabular}
}
\end{table}

\mypara{Input resolution vs. sharpness.}
To quantify the correlation between resolution and sharpness,
we vary the input resolution to see how that impacts metrics.
We resize the shorter side of every input image to fixed sizes, i.e.,
360px, 540px, 720px, or keep their native resolution.
We report metric depth accuracy, relative depth accuracy, boundary sharpness,
and temporal consistency, as shown in Table~\ref{tab:window-resolution}.
We find that from 360px to native,
the boundary F1 score more than doubles, from 24.1 to 57.7.
Jitter also improves as resolution increases from 0.33 down to 0.24.
Metric and relative depth accuracy also improve slightly through scaling resolution.
\begin{table}[h]
  \centering\ra{1.05}\footnotesize
  \caption{\textbf{Input resolution vs. sharpness.} We evaluate \oursh{} at different input resolutions, on clips of up to 32 frames from Waymo, ARKitScenes, and DynamicReplica. Native resolution means 720px for DynamicReplica, 886px or 1280px for Waymo, and 1440px for ARKitScenes. Boundary F1 score more than doubles as input resolution increases. Jitter is scaled by $10^2$. Bold: best.}
  \label{tab:window-resolution}
  \fitwidth{\begin{tabular}{@{}lcccccc@{}}
  \toprule
  & \multicolumn{2}{c}{Metric $\delta_1$ (\%) $\uparrow$} & \multicolumn{2}{c}{Relative $\delta_1$ (\%) $\uparrow$} & \multicolumn{1}{c}{Boundary F1 (\%) $\uparrow$} & \multicolumn{1}{c}{Jitter $\downarrow$} \\
  \cmidrule(lr){2-3}\cmidrule(lr){4-5}
  Input short side & \multicolumn{1}{c}{Waymo} & \multicolumn{1}{c}{ARKitScenes} & \multicolumn{1}{c}{Waymo} & \multicolumn{1}{c}{ARKitScenes} & \multicolumn{2}{c}{DynamicReplica} \\
  \midrule
  360 & 94.3 & 90.4 & 94.8 & 99.0 & 24.1 & 0.33 \\
  540 & 95.0 & 91.3 & 95.5 & \textbf{99.2} & 38.0 & 0.28 \\
  720 & 95.0 & \textbf{92.5} & 95.6 & \textbf{99.2} & \textbf{57.7} & \textbf{0.24} \\
  Native & \textbf{95.4} & 92.4 & \textbf{95.8} & \textbf{99.2} & \textbf{57.7} & \textbf{0.24} \\
  \bottomrule
\end{tabular}
}
\end{table}

\subsection{Long video sequences}
\label{sec:appendix-long-video}
\begin{figure}[tbh]
  \centering
  \includegraphics[width=\linewidth]{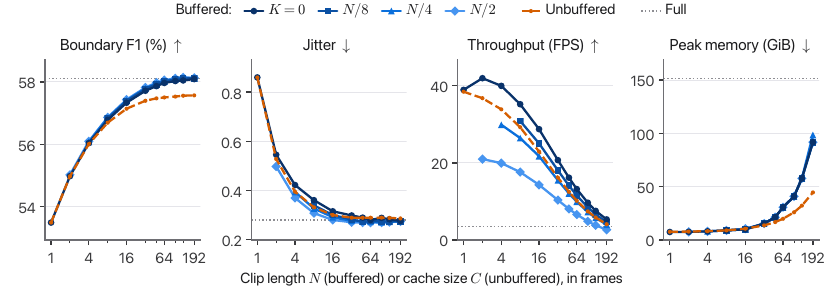}
  \caption{\textbf{Temporal context on long videos.} \oursh{} on 300-frame DynamicReplica sequences against the clip length $N$ in buffered streaming (solid blue) and the cache size $C$ in unbuffered streaming (dashed orange). Buffered streaming is shown for overlaps $K$ of 0, $N/8$, $N/4$ and $N/2$ frames where these are whole numbers, each with its own marker and a lighter blue for more overlap. The overlaps nearly coincide in boundary F1 and peak memory, with F1 changing by up to about 0.1 points. The dotted line is a single pass over all 300 frames (Full). Throughput and peak memory are measured in PyTorch eager mode at 720p as in Figure~\ref{fig:runtime}. Jitter is scaled by $10^2$. Jitter levels off at $N=64$ and at $C=32$. Overlap reduces jitter slightly but lowers throughput. Metric and relative $\delta_1$ barely change and are not plotted. They stay between 98.8\% and 99.3\% in buffered streaming and Full, and between 98.8\% and 99.2\% in unbuffered streaming.}
  \label{fig:long-video}
\end{figure}

We evaluate both inference modes (Section~\ref{sec:long-video-inference}) on 300-frame DynamicReplica sequences. In both modes, temporal context mainly improves stability: jitter decreases and boundary F1 increases with clip length $N$ and cache size $C$, while depth accuracy is largely unaffected. In buffered streaming, these gains level off at $N=64$, where buffered streaming matches full-sequence inference at 3.8 times the throughput and a fifth of the peak memory (Figure~\ref{fig:long-video}).
Overlapping clips reduce jitter further but lower throughput.
In unbuffered streaming, jitter levels off at $C=32$, the length of the longest training clip (Figure~\ref{fig:long-video}), while each cached frame adds a roughly constant cost in latency and peak memory.

\mypara{Evaluation setup.}
We use the first 300 frames of each of the held-out DynamicReplica sequences (Appendix~\ref{sec:appendix-validation}) at 720p. We provide  ground-truth intrinsics, so metric accuracy reflects depth estimation alone and excludes errors from focal length estimation. In buffered streaming, we blend overlapping clips as described in Section~\ref{sec:long-video-inference}. We measure throughput, latency, and peak memory with the protocol of Appendix~\ref{sec:appendix-runtime-protocol}.

\subsubsection{Buffered streaming}
Buffered streaming processes a video that exceeds the memory of a single forward pass as a sequence of clips. Figure~\ref{fig:long-video} (solid blue) shows the effect of the clip length $N$ and the overlap $K$ between consecutive clips.

\mypara{Clip length.} Longer clips improve temporal consistency and boundary accuracy. Without overlap, increasing $N$ from 1 to 64 frames reduces jitter from 0.860 to 0.280 and raises boundary F1 from 53.5\% to 58.0\%. Most of the gain comes early: 8-frame clips account for more than 85\% of the reduction in jitter. Jitter keeps decreasing up to $N=64$, twice the longest training clip, and a single pass over all 300 frames matches this level (0.279) at a peak memory of 151.5\,GiB and 3.5\,FPS. Peak memory grows with $N$, and throughput decreases with $N$ from 2-frame clips onward.

\mypara{Overlap.} Overlap reduces jitter while leaving depth and boundary accuracies largely unchanged. Since overlapping frames are processed twice, throughput scales with the fraction of new frames per clip, $(N-K)/N$. Peak memory is the same for every overlap of a clip length, except for $N=16$ with $K=8$ and $N=192$ with $K=48$.

\mypara{Recommended settings.} We recommend 16-frame clips with an overlap of 4 frames as the default. They reach a jitter of 0.291 at 21.7\,FPS and 10.0\,GiB, with a metric $\delta_1$ of 99.2\%. Longer clips and overlap lower jitter by at most a further 0.022 (0.269 at $N/K = 48/24$, $64/32$ and $96/48$), at less than half of the throughput.

\subsubsection{Unbuffered streaming}
Unbuffered streaming immediately processes each input frame upon arrival.
During the processing, the cross-attention layers attend over a window of $C$ frames: the current frame and the $C-1$ frames before it, whose keys and values are cached. Figure~\ref{fig:long-video} (dashed orange) shows how changing $C$ affects performance.
We find that increasing the cache size improves temporal consistency: jitter drops from 0.860 with a single frame to 0.300 at 16 frames and levels off at 0.289 from 32, the length of the longest training clip. Beyond $C=16$, jitter improves by at most 5\%.
Boundary F1 follows the same trend, while metric and relative $\delta_1$ stay between 98.8\% and 99.2\% for all cache sizes. Each cached frame adds a roughly constant cost of about 1.14\,ms in latency and 0.19\,GiB in peak memory, from  44\,ms and 10.6\,GiB at $C=16$ to 62\,ms and 13.7\,GiB at $C=32$. We therefore recommend $C=16$ as the default.

\subsubsection{Buffered vs.\ unbuffered streaming}
Table~\ref{tab:streaming} compares the two modes with their default settings, buffered streaming with $N=16$ and $K=4$ and unbuffered streaming with $C=16$, on the same sequences. Figure~\ref{fig:runtime} shows the throughput and peak memory of unbuffered streaming. Unbuffered streaming matches buffered streaming in quality for all four models.
Relative $\delta_1$ and boundary F1 differ by at most 0.2 and 0.3 points, respectively, and jitter is 3\% higher for \oursl{} and \oursh{} and less than 1\% lower for \ourss{} and \oursb{}. Buffered streaming attains higher metric $\delta_1$ for all four models, by up to 0.6 points for \oursb{}.
\begin{table}[h]
  \centering\ra{1.05}\small
  \caption{\textbf{Buffered vs.\ unbuffered streaming.} Buffered streaming uses clips of $N=16$ frames that overlap by $K=4$. Unbuffered streaming uses a cache of $C=16$ frames. 300-frame
  DynamicReplica sequences at 720p. The two modes are on par for all models. Jitter is scaled by $10^2$. Bold: best per model.}
  \label{tab:streaming}
  \fitwidth{\begin{tabular}{@{}llcccc@{}}
  \toprule
  Model & Mode & \multicolumn{1}{c}{Metric $\delta_1$ (\%) $\uparrow$} & \multicolumn{1}{c}{Relative $\delta_1$ (\%) $\uparrow$} & \multicolumn{1}{c}{Boundary F1 (\%) $\uparrow$} & \multicolumn{1}{c}{Jitter $\downarrow$} \\
  \midrule
  \tOursS & Buffered & \textbf{96.7} & \textbf{97.3} & \textbf{38.2} & 0.567 \\
   & Unbuffered & 96.3 & 97.2 & 38.1 & \textbf{0.563} \\
  \midrule
  \tOursB & Buffered & \textbf{98.7} & \textbf{98.5} & \textbf{49.9} & 0.399 \\
   & Unbuffered & 98.1 & 98.3 & 49.7 & \textbf{0.397} \\
  \midrule
  \tOursL & Buffered & \textbf{99.3} & \textbf{99.1} & \textbf{55.8} & \textbf{0.311} \\
   & Unbuffered & 99.0 & 99.0 & 55.5 & 0.321 \\
  \midrule
  \tOursH & Buffered & \textbf{99.2} & 99.1 & \textbf{57.4} & \textbf{0.291} \\
   & Unbuffered & 99.1 & 99.1 & 57.1 & 0.300 \\
  \bottomrule
\end{tabular}
}
\end{table}

\subsection{Focal-length estimator}
\label{sec:appendix-focal-design}

\mypara{Separate estimator vs.\ bespoke head per depth model.}
We explored several ways to estimate focal length for our family of Depth Pro 2 models.
One attractive option is to attach a small, trainable FoV head to each depth model's frozen encoder, which adds negligible compute by reusing depth encoder features during a forward pass.
This works well on \oursh: a 1.9M-parameter transformer head reaches a mean $\delta_{25}$ of 86.5\% (Table~\ref{tab:focal-design}).
On the smaller models, however, this approach does not reach satisfactory accuracy.
The same head on \ourss reaches only 74.5\%, and a head eight times larger still falls short (78.8\%).
This motivates us to use a single separate estimator for all four models (88.6\%).
A separate estimator costs one extra forward pass, at a capped 512-pixel short side, and 23.5M extra parameters.
In return, we can ensure every Depth Pro 2 model receives the same focal length for a given image, and the estimator does not need to be retrained when the depth models change.

\begin{table}[h]
  \centering\ra{1.05}\footnotesize
  \caption{
  {\bf FoV heads on frozen depth encoders vs.\ our separate estimator.}
  Mean $\delta_{25}$ over the six photograph benchmarks of Table~\ref{tab:focal-photo}.
  All variants are trained on Flickr. Finetuning a DINOv3 encoder outperforms any variant using a frozen \ours{} encoder.
  }
  \label{tab:focal-design}
  \fitwidth{
\begin{tabular}{lrcrc}
    \toprule
    Encoder & FoV head & Input & Trainable params & FoV $\delta_{25}$ \\
    \midrule
    frozen \ourss & 1.9M & native, $\le$768 & 1.9M & 74.5 \\
    frozen \ourss & 15.1M & native, $\le$768 & 15.1M & 78.8 \\
    frozen \oursh & 1.9M & native, $\le$768 & 1.9M & 86.5 \\
    \midrule
    finetuned DINOv3 ViT-S (ours) & 1.9M & native, $\le$512 & 23.5M & \textbf{88.6} \\
    \bottomrule
\end{tabular}
}
\end{table}

\begin{figure}[h]
  \centering
  \includegraphics[width=0.7\linewidth]{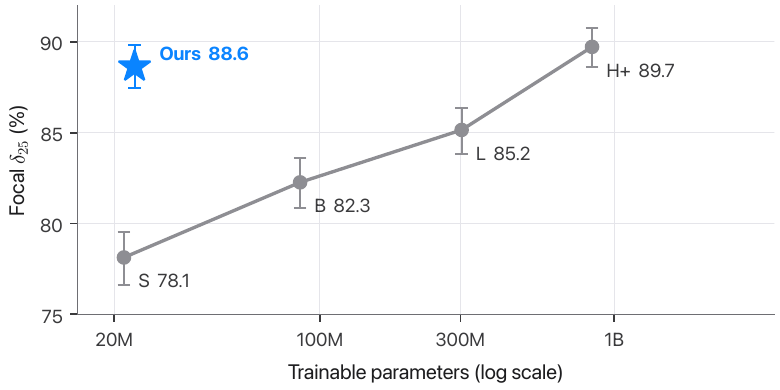}
  \caption{
  {\bf FoV accuracy vs.\ estimator size.}
  Mean $\delta_{25}$ over the six photograph benchmarks; bars are 95\% bootstrap intervals over images.
  Gray: the $256 \times 256$ baseline recipe on four DINOv3 sizes.
  }
  \label{fig:focal-params}
\end{figure}

\mypara{Model size.} Our focal-length estimator uses ViT-S, the smallest DINOv3 backbone. Under a simple recipe ($256\times256$ input and a linear head on the mean patch token), FoV $\delta_{25}$ grows steadily with backbone size (Figure~\ref{fig:focal-params}). Our training pipeline lets ViT-S outperform ViT-L (paired 95\% bootstrap interval $[+2.2, +4.9]$) and come within 1.1 points of ViT-H\textsuperscript{+}. It does so with slightly less compute than ViT-L and a third of the compute of ViT-H\textsuperscript{+}, and with $13\times$ and  $36\times$ fewer parameters. Most of the gain comes from the native aspect ratio, the 512-pixel short side, and the optimization described in Appendix~\ref{sec:appendix-focal-setup}; the attention head contributes 1.6 points.

\mypara{Input resolution and aspect ratio.} The estimator processes each image at its native aspect ratio. We train with a single 512-pixel short side, the resolution at which an earlier multi-resolution model was most accurate. At test time, the estimator is stable for short sides between 384 and 768 pixels and degrades outside that range: mean $\delta_{25}$ is 80.1\% at 256, 86.9\% at 384, 88.6\% at 512, 87.3\% at 768, and 84.9\% at 1024. We therefore resize test images with a short side above 512 pixels to 512, round the short side down to a multiple of 32, and never upsample smaller images, because upsampling them has mixed effects ($+5.6$ $\delta_{25}$ on PPR10K, $-6.0$ on iBims).

\section{Evaluation protocol}
\label{sec:appendix-evaluation-protocols}

\subsection{Temporal consistency}
\label{sec:appendix-jitter}
We evaluate temporal stability along point tracks computed on synthetic data.
Let track $k$ be one such point, seen at pixel $p_k(i)$ in frame $i$. We measure the residual $e_k(i)$ between predicted and true log-depth, and define its change from one frame to the next as $\mathrm{jitter}_k$:
\begin{equation}
 e_k(i) = \log\hat{\mathbf Z}_i(p_k(i)) - \log\mathbf Z_i(p_k(i)),
 \qquad
 \mathrm{jitter}_k = \operatorname{mean}_i\bigl|e_k(i{+}1) - e_k(i)\bigr|,
 \label{eq:tc}
\end{equation}
$\mathrm{jitter}_k$ is identical to TEPE~\citep{Karaev2023:CVPR, Zhu2025:CVPR},
but computed in log-space so values are comparable across varying scene scales.

\subsection{Zero-shot evaluation details}
\label{sec:appendix-eval-details}

\mypara{Zero-shot status.} We show a result in gray when the evaluated checkpoint was trained on the evaluation dataset. Gray entries are excluded from the per-dataset ranks and from the average ranks across datasets, so the average rank of a method with gray entries covers only the datasets on which it is zero-shot.

\mypara{Single-image and video settings.} \ours{} accepts both single frames and multi-frame clips, as do several baselines. Most other baselines accept only single images. \depthcrafter{}, \velodepth{} and \vda{} accept only clips, and \velodepth{} reduces to \unikthreed{} on a single frame. To compare both kinds of method fairly, we evaluate metric depth and boundary sharpness in two settings. In the single-image setting, every method receives one frame at a time. On video and multi-view datasets, we predict and score each of the first 50 frames of a sequence as a separate image, and methods estimate the focal length for every frame independently. In the video setting, every method receives the first 50 frames of a sequence as one clip. Video  methods process the clip jointly, and single-image methods process its frames independently. We evaluate temporal consistency in the video setting only. The boundary metric of \citet{Bochkovskii2025:ICLR} and our temporal-consistency metrics are scale-invariant, so we also evaluate methods that predict only relative depth on them.

\mypara{Baseline evaluation.} We evaluate all baselines from their released checkpoints with the protocols in this section, unless noted otherwise. Our scores may differ from published results, because studies differ in camera intrinsics, alignment procedures, sampled frames or sequences, and evaluation resolution.

\mypara{Evaluation splits.} \citet{Lin2026:ICLR} and \citet{Piccinelli2025:CVPR} evaluate on the indoor split of \diode{}~\citep{Vasiljevic2019:arxiv} only. We evaluate on the full validation split, indoor and outdoor. The outdoor scenes are harder, so our \diode{} scores are lower than those reported on the indoor split. We use the official \diode{} validity mask for every metric. For nuScenes, we use the evaluation split of \citet{Bochkovskii2025:ICLR}, and for HAMMER, the split of \citet{Kong2026:arxiv}. Both splits consist of individual frames, so we evaluate them in the single-image setting only.

\subsection{Baseline configurations}
\label{sec:appendix-baselines}
\label{sec:appendix-alignment}

\mypara{Common input.} We evaluate every method on the same frames: one contiguous clip of the first 50 frames per video sequence, sampled at stride one. Prior work samples frames, sets clip lengths, and aligns predictions in different ways, so published video-depth results are hard to compare directly. For example, \vggtomega{} evaluates 10 randomly sampled frames per sequence~\citep[\S4]{Wang2026:CVPR}. \vda{} evaluates up to 500 frames per video, reports 90- and 110-frame settings for comparison with \depthcrafter{}, and aligns predictions in inverse-depth space~\citep{Chen2025:CVPR}. The one exception to our common input is the temporal alignment evaluation on ScanNet (Table~\ref{tab:paper-temporal-tae}), where we use the 170-frame clips of the \vda{} evaluation~\citep{Chen2025:CVPR}.

\mypara{Clip processing.} Each method processes the clip with its official inference procedure. Methods that accept 50 frames receive the whole clip in one pass; the others use their released windowing, \eg, \vda{} processes 32-frame windows. On ScanNet, \oursh{} runs in buffered streaming with clips of $N=64$ frames and an overlap of $K=32$.

\mypara{Input resolution.}
We retain each method’s official inference resolution because changing it can affect metric accuracy, boundary sharpness, temporal consistency, and runtime. For example, \depthpro{} resizes every frame to $1536\times1536$. Some methods crop the input in their official preprocessing (\eg, \vggtomega{}). For these methods, we pad each frame so that the crop removes only padding, and then remove the padding from the prediction. Every method is thus evaluated on the same image region.

\mypara{Output resizing.} Because methods differ in their internal preprocessing, their predictions may not match the spatial resolution of the ground truth. In such cases, we resize the prediction to the ground-truth resolution using nearest-neighbor interpolation, thereby avoiding additional smoothing at depth discontinuities.

\mypara{Affine alignment.} Before computing relative-depth accuracy, boundary F1, and temporal consistency, we align each prediction to the ground truth with a scale and a shift. We align every method this way, including metric-depth models and \ours{}, so that these metrics measure the structure of a prediction independently of its metric scale. For relative-depth accuracy (Tables~\ref{tab:paper-relative-single} and~\ref{tab:paper-relative-video}) and boundary F1 (Table~\ref{tab:paper-sharpness}), we fit the scale and shift for each frame separately, also when a method predicts the frames of a clip jointly. For jitter  (Table~\ref{tab:paper-temporal}) and TAE (Table~\ref{tab:paper-temporal-tae}), we fit one scale and shift per video, which preserves the variation in scale across frames that these metrics measure.

By default, we fit the scale and shift in inverse-depth space by trimmed least squares: we keep the 80\% of sampled valid pixels with the smallest squared residuals. Three methods output log depth (\lapis{}, \pixelperfectdepth{}, and \marigoldvtwo{}). Following their evaluation protocols, we fit these methods in log-depth space with the same trimming.
Tables~\ref{tab:paper-sharpness-single}, \ref{tab:paper-sharpness-video}, and~\ref{tab:paper-temporal} mark them with ``(log)''.

\subsection{Held-out validation}
\label{sec:appendix-validation}

The controlled experiments (Appendix~\ref{sec:appendix-controlled}), the inference-time analysis studies (Appendix~\ref{sec:appendix-diagnostic}),  and the long-video evaluation (Appendix~\ref{sec:appendix-long-video}) use held-out validation sequences from three datasets that do not appear in the zero-shot evaluation: Waymo~\citep{Sun2020:CVPR} (real, outdoor; 42 sequences at $1280\times1920$ or $886\times1920$), ARKitScenes~\citep{Dehghan2021:NeurIPS} (real, indoor; 92 sequences at $1440\times1920$), and DynamicReplica~\citep{Karaev2023:CVPR} (synthetic; 44 sequences at $720\times1280$). We train on the training sequences of Waymo and DynamicReplica, and on those of ARKitScenes only in condition D1 of the controlled experiments. Conditions D2 and D3 leave out Waymo and D3 also DynamicReplica. We hold out 5\% of the data shards of each dataset from training, selected by a fixed hash. Of the 46 held-out DynamicReplica sequences, we discard two in which the camera passes through scene geometry: over part of each sequence, the ground truth contains a surface a few centimeters from the camera that the image does not show.

From each sequence, we take one window of up to 32 consecutive frames, starting at a seeded random frame, and process it in a single forward pass, at native resolution unless a study varies the input size. The long-video evaluation instead uses the first 300 frames of each sequence.

We report four metrics: metric $\delta_1$ with the ground-truth focal length; relative $\delta_1$ after a per-frame affine fit in inverse-depth space; the boundary F1 of \citet{Bochkovskii2025:ICLR}; and, on DynamicReplica, jitter (Eq.~\ref{eq:tc}) along point tracks obtained by chaining the ground-truth optical flow.

\subsection{Runtime measurement}
\label{sec:appendix-runtime-protocol}

\begin{figure}[thb]
  \centering
  \includegraphics[width=0.9\linewidth]{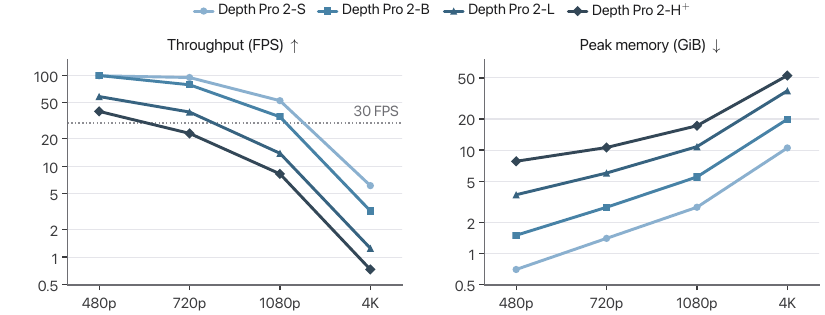}
  \caption{
  \textbf{Throughput and peak memory of \ours{} in unbuffered streaming.} The four \ours{} models on one B200 GPU with a cache of $C=16$ frames, for input
  sizes from 480p to 4K, with the focal length given (Appendix~\ref{sec:appendix-runtime-protocol}). The Y axes are in log scale. Darker lines denote larger models.
  The dotted line marks the 30\,fps frame rate of a typical camera. Unbuffered streaming runs in real time at 480p for all models, at 720p for
  \ourss{}, \oursb{}, and \oursl{}, and at 1080p for \ourss{} and \oursb{}. All measurements are in PyTorch eager mode.
  }
  \label{fig:runtime}
\end{figure}

\mypara{Setup.} We measure the runtime of \ours{} in Figures~\ref{fig:long-video} and~\ref{fig:runtime} with our code release on one B200 GPU, in eager PyTorch without \texttt{torch.compile}, with cuDNN attention (the PyTorch default on this GPU). The backbone and decoder run under bf16 autocast, and the depth head runs in fp32. The inputs are synthetic 16:9 frames. The focal length is given, so the field-of-view network is excluded from the timings. Each configuration, defined by the model, the input size, and $N$ and $K$ or $C$, runs in its own process. We run each configuration 3 times and report the median across runs.

\mypara{Timing.} We time two operations: a buffered pass over a clip of $N$ frames, and an unbuffered step with a full cache of $C$ frames. Each time is the median of 20 calls after 2 warm-up calls. Latency and throughput follow from these times.

\mypara{Latency.} Latency is the time from the arrival of a frame from a 30\,fps camera to the availability of its depth map. In unbuffered streaming, each frame is processed on arrival, so its latency is the time of one step. This assumes that the GPU is idle when the frame arrives, which holds when throughput is at least 30\,fps; at lower throughput, the frames of a live camera queue up or must be dropped.

\mypara{Throughput.} Throughput is the number of frames finalized per second on a long video.
In buffered streaming, each pass finalizes $N-K$ frames, so throughput is $N-K$ divided by the time of one pass. In unbuffered streaming, it is one divided by the time of one step.

\mypara{Peak memory.} Peak memory is the maximum memory reserved by PyTorch's caching allocator while processing a video, over one warm-up pass and three timed passes. In buffered streaming, the video spans at least two clips, as a long video does: it has 128 frames for clips of up to 64 frames and $2N$ frames for $N \in \{96, 128, 192\}$. The \textit{Full} variant in Figure~\ref{fig:long-video} uses 300 frames. In unbuffered streaming, the video has 128 frames for every cache size, since the cache is allocated in full on the first frame.

\end{document}